\documentclass[twocolumn, switch, 10pt]{article}

\usepackage{preprint}

\usepackage{amsmath, amsthm, amssymb, amsfonts}

\usepackage[sort&compress]{natbib}
\AtBeginDocument{\def\NAT@space{\relax}}%
\usepackage[utf8]{inputenc}	%
\usepackage[T1]{fontenc}	%
\usepackage{xcolor}		%
\usepackage[colorlinks = true,
            linkcolor = purple,
            urlcolor  = blue,
            citecolor = cyan,
            anchorcolor = black]{hyperref}	%
\usepackage{booktabs} 		%
\usepackage[shortlabels,inline]{enumitem}
\setlist{leftmargin=*}

\usepackage{xurl}            %
\usepackage{nicefrac}       %
\usepackage{microtype}      %
\usepackage{cleveref}       %
\usepackage{listings}
\usepackage[listings,theorems,breakable,skins]{tcolorbox}
\usepackage{xltabular}
\usepackage{makecell}
\usepackage{blindtext}
\usepackage{wasysym} %
\usepackage{tikz}
\usepackage{soul}

\usepackage{subcaption}

\usepackage{threeparttable}

\definecolor{darkColor}{HTML}{D4E4E0} %
\definecolor{lightColor}{HTML}{CC505C}
\definecolor{mediumGray}{HTML}{6D6E70}
\definecolor{lightGray}{HTML}{e8e9eb}

\definecolor{darkgreen}{rgb}{0.0, 0.5, 0.0}
\definecolor{blue}{rgb}{0.0, 0.47, 0.75}
\definecolor{dartmouthgreen}{rgb}{0.05, 0.5, 0.06}
\definecolor{drab}{rgb}{0.59, 0.44, 0.09}
\definecolor{navyblue}{rgb}{0.0, 0.0, 0.5}

\definecolor{codegreen}{rgb}{0,0.6,0}
\definecolor{codegray}{rgb}{0.5,0.5,0.5}
\definecolor{codepurple}{rgb}{0.58,0,0.82}
\definecolor{backcolour}{rgb}{0.95,0.95,0.92}

\newcolumntype{Y}{>{\centering\arraybackslash}X}

\global

\tcbset{aibox/.style={
    breakable,
    break at=\maxdimen,
    height fixed for=first and middle,
}}

\newtcolorbox{AIbox}[2][]{aibox,title=#2,#1}

\definecolor{aiyellow}{HTML}{FBF1D1}
\definecolor{aigold}{RGB}{244,210, 1} 
\definecolor{aigreen}{RGB}{210,244,211} 

\sethlcolor{aigreen}

\definecolor{aired}{RGB}{255,180,181}

\newcommand{\hle}[2][aiyellow]{{%
    \colorlet{foo}{#1}%
    \sethlcolor{foo}\hl{#2}}%
}

\newtcbox{\mybox}[1][green]{on line,
arc=0pt,outer arc=0pt,colback=#1!10!white,colframe=#1!50!black,
boxsep=0pt,left=0pt,right=0pt,top=0pt,bottom=0pt,
boxrule=0pt,bottomrule=0pt,toprule=0pt}

\newcommand{\vx}{\mathbf{x}}

\newtcolorbox[auto counter,number within=subsection]{contextbox}[2][]{%
enhanced,
colback=white,
colframe=darkColor,
fonttitle=\bfseries\fontfamily{phv}\selectfont\color{black},
size=small, boxsep=0mm, top=1mm, boxrule=0.5mm, left=0mm, right=0mm,
title=,#1}

\NewTColorBox[]{promptbox}{O{}mO{GPT-4V Answer:}}{%
enhanced,frame hidden,%
colback=white,
coltitle=black,
skin=bicolor, size=small,
colbacklower=lightGray, boxrule=0mm,%
before lower={\parbox[][10pt][t]{\linewidth}{\textit{\textbf{#3}}}},%
  fontupper=\fontsize{6}{7.2}\selectfont\fontfamily{phv}\selectfont, %
  #1,%
  fontlower=\fontsize{6}{7.2}\selectfont\fontfamily{phv}\selectfont
}

\usepackage{titlesec}
\titlespacing\section{0pt}{12pt plus 3pt minus 3pt}{1pt plus 1pt minus 1pt}
\titlespacing\subsection{0pt}{10pt plus 3pt minus 3pt}{1pt plus 1pt minus 1pt}
\titlespacing\subsubsection{0pt}{8pt plus 3pt minus 3pt}{1pt plus 1pt minus 1pt}

\title{From Concept to Manufacturing: Evaluating Vision-Language Models for Engineering Design}

\usepackage{authblk}

\newbox{\orcid}\sbox{\orcid}{\includegraphics[scale=0.06]{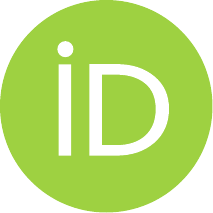}}

\author[1*]{%
	\href{https://orcid.org/0000-0002-3434-0383}{\usebox{\orcid}\hspace{1mm}Cyril Picard}%
}
\author[1*]{%
	\href{https://orcid.org/0009-0000-4316-8119}{\usebox{\orcid}\hspace{1mm}Kristen M.~Edwards}} %
\author[1]{%
\href{https://orcid.org/0009-0003-0125-0909}{\usebox{\orcid}\hspace{1mm}Anna C.~Doris}%
}
\author[1]{%
	\href{https://orcid.org/0009-0002-9892-430X}{\usebox{\orcid}\hspace{1mm}Brandon Man}%
}
\author[1,2]{\\%
	\href{https://orcid.org/0000-0002-5810-4414}{\usebox{\orcid}\hspace{1mm}Giorgio Giannone}%
}
\author[1]{%
	\href{https://orcid.org/0000-0002-8469-7591}{\usebox{\orcid}\hspace{1mm}Md Ferdous Alam}%
}
\author[1]{%
	\href{https://orcid.org/0000-0002-5227-2628}{\usebox{\orcid}\hspace{1mm}Faez Ahmed}%
}

\affil[1]{Department of Mechanical Engineering, Massachusetts Institute of Technology, Cambridge, MA 02139}
\affil[2]{Department of Applied Mathematics and Computer Science, Technical University of Denmark, Lyngby, DK 2800}
\affil[*]{\footnotesize Equal Contribution. Corresponding authors: \texttt{cyrilp@mit.edu} and \texttt{kme@mit.edu}}

\date{}

\begin{document}

\twocolumn[ %
  \begin{@twocolumnfalse} %

  \maketitle
\begin{abstract} %
  Engineering design is undergoing a transformative shift with the advent of AI, marking a new era in how we approach product, system, and service planning. Large language models have demonstrated impressive capabilities in enabling this shift. Yet, with text as their only input modality, they cannot leverage the large body of visual artifacts that engineers have used for centuries and are accustomed to. This gap is addressed with the release of multimodal vision-language models (VLMs), such as GPT-4V, enabling AI to impact many more types of tasks. 
  Our work presents a comprehensive evaluation of VLMs across a spectrum of engineering design tasks, categorized into four main areas: Conceptual Design, System-Level and Detailed Design, Manufacturing and Inspection, and Engineering Education Tasks. Specifically in this paper, we assess the capabilities of two VLMs, GPT-4V and LLaVA 1.6 34B, in design tasks such as sketch similarity analysis, CAD generation, topology optimization, manufacturability assessment, and engineering textbook problems. Through this structured evaluation, we not only explore VLMs' proficiency in handling complex design challenges but also identify their limitations in complex engineering design applications. 
  Our research establishes a foundation for future assessments of vision language models. It also contributes a set of benchmark testing datasets, with more than 1000 queries, for ongoing advancements and applications in this field.
\end{abstract}
\vspace{0.35cm}

  \end{@twocolumnfalse} %
] %

\section{Introduction}
Large language models (LLMs) have shown promising performance in domains ranging from medicine~\citep{arkoudas2023gpt}, to law~\citep{katz2023gpt}, to mathematics and coding~\citep{bubeck_sparks_2023}. The chat-like interfaces offered by tools such as Google's Bard~\citep{bard2023} or OpenAI's ChatGPT~\citep{gpt4v} have enabled millions of users to query and leverage their immense implicit knowledge to assist in tasks ranging from diagnosing diseases to creating manufacturing drawings~\citep{makatura_how_2023}, to supporting the conceptual design stage for a robotic arm~\citep{stella_how_2023}. Their use of natural language as the input modality offers an intuitive interface for humans to express a variety of problems, often in their mother tongue and without the need for technical jargon~\citep{bubeck_sparks_2023}. The diversity of possibilities is immense and a review identified many tasks within engineering design that could be automated using natural language processing tools~\citep{siddharth_natural_2022}. Examples include automated design knowledge retrieval, discovery of user needs, and documentation generation.

However, expressing some tasks in text alone is prohibitively complex. In certain domains, especially those involving spatial information, a picture is truly worth one thousand words. Imagine if one had to write assembly instructions for IKEA furniture using text instructions alone. One would have to describe each part, define its orientation, and instruct where to put screws repeatedly using text similar to ``Insert screw A in the third hole starting from the top, where the top of the board is the shortest edge that has a slot through its length.'' Unsurprisingly, engineers have relied on visual artifacts to communicate and support their work for centuries~\citep{henderson_line_1999}. From sketches that highlight the key working principles of mechanisms to manufacturing drawings that elicit all the information needed to manufacture, assemble, and inspect parts, visual representations are ubiquitous in engineering.

Recently, powerful multimodal language models have been proposed. In particular, text and vision models like GPT-4V~\citep{gpt4v}, LLaVA 1.6-34B~\citep{liu2024llavanext,Liu2023}, and Fuyu-8B~\citep{fuyu-8b} have shown immense promise since their public release. These vision-language models (VLMs) can take images and text as input and generate text as output. Specifically, most VLMs are based on a LLM and extend its capabilities by adding the embedding from a vision encoder as token the LLM's input context. For example, GPT-4V builds upon the leading LLM, GPT-4V. Researchers have begun exploring the capabilities of VLMs in several application domains. Examples include image understanding and reasoning~\citep{yang2023dawn}, image and language association~\citep{liu2023hallusionbench}, and optical character recognition~\citep{shi2023exploring}. 

To better understand LLMs' capabilities, researchers have investigated the performance of LLMs on standardized tests in specific fields such as law~\citep{katz2023gpt}, or medicine~\citep{nori2023capabilities,Rosol2023,Takagi2023}, or on specific skills, such as reasoning or mathematics~\citep{arkoudas2023gpt,bubeck_sparks_2023}. 
However, there is interest in evaluating LLMs and VLMs on their performance on tasks outside of existing standardized exams. Doing this requires creating assessment questions and answers that accurately reflect the task of interest. Ref. \citep{feng2024evaluate} evaluate GPT-4 on an upper-level undergraduate computer science course. The authors compiled and classified a dataset of past assessment questions, and released their benchmark dataset. Other researchers have focused on evaluating the applicability of LLMs through their chat interface for specific topics, including design for manufacturing~\citep{makatura_how_2023}, design for additive manufacturing~\citep{BADINI2023278}, and the design of cementitious composites~\citep{CAI2024135965}.

One challenge for comparing different LLMs or VLMs accurately is that most evaluation research is looking at a select-few models on a specific task with a specific dataset and metric. If models' performance are not compared on the same tasks with the same datasets using the same metrics, it is difficult to draw accurate conclusions about how the two models compare. To address this, \citet{chang2024survey} survey current LLM research to understand how and where LLMs are being evaluated. They compile 46 popular benchmarks ranging from general language tasks to domain-specific question and answering tasks. Further, Chang et al. outline the major areas where LLMs have been evaluated thus far. These are: natural language processing tasks; robustness, ethics, biases and trustworthiness; social sciences; natural science and engineering; and medical applications. Chang et al. identify nearly 100 publications in these categories. They point out, however, that there remains a lack of standardization even within each category, but this can be improved by the development of standardized benchmarks and automatic evaluation.

The challenging tradeoff between automatic and human evaluation is quite present in this field~\citep{chang2024survey, llm_as_judge_2023, zhu2024promptbench}. For certain tasks around human preferences, automatic evaluation techniques may be insufficient, however human ratings are expensive and hard to scale. Researchers are exploring the use of LLMs as judges, focusing on their agreement with human evaluations and identifying their biases~\citep{llm_as_judge_2023}. New standardized and automatic benchmarks are also being developed to address this. PromptBench, for example, is a unified Python library for evaluating LLMs with modular options for different LLMs, tasks, datasets, benchmarks, prompts, and prompt engineering techniques~\citep{zhu2024promptbench}. Still, many of these benchmarks and tasks are developed for language-only models.

Looking to evaluate \textit{vision}-language models rather than language-only models, MMBench~\citep{MMBench} is a work that specifically addresses benchmarking VLMs via ``an objective evaluation pipeline of 2,974 multiple-choice questions covering 20 ability
dimensions.'' This work highlights the importance of developing VLM benchmarks that are relevant to a given field. However, the 20 ability dimensions in~\citep{MMBench} are quite general, and not specialized to engineering design tasks.

In this work, we aim to create an evaluation methodology including curated tasks, datasets, and benchmarks to answer the question ``Can VLMs effectively perform tasks or assist engineers in engineering design?''

\begin{figure}
    \centering
    \includegraphics[width=0.7\linewidth]{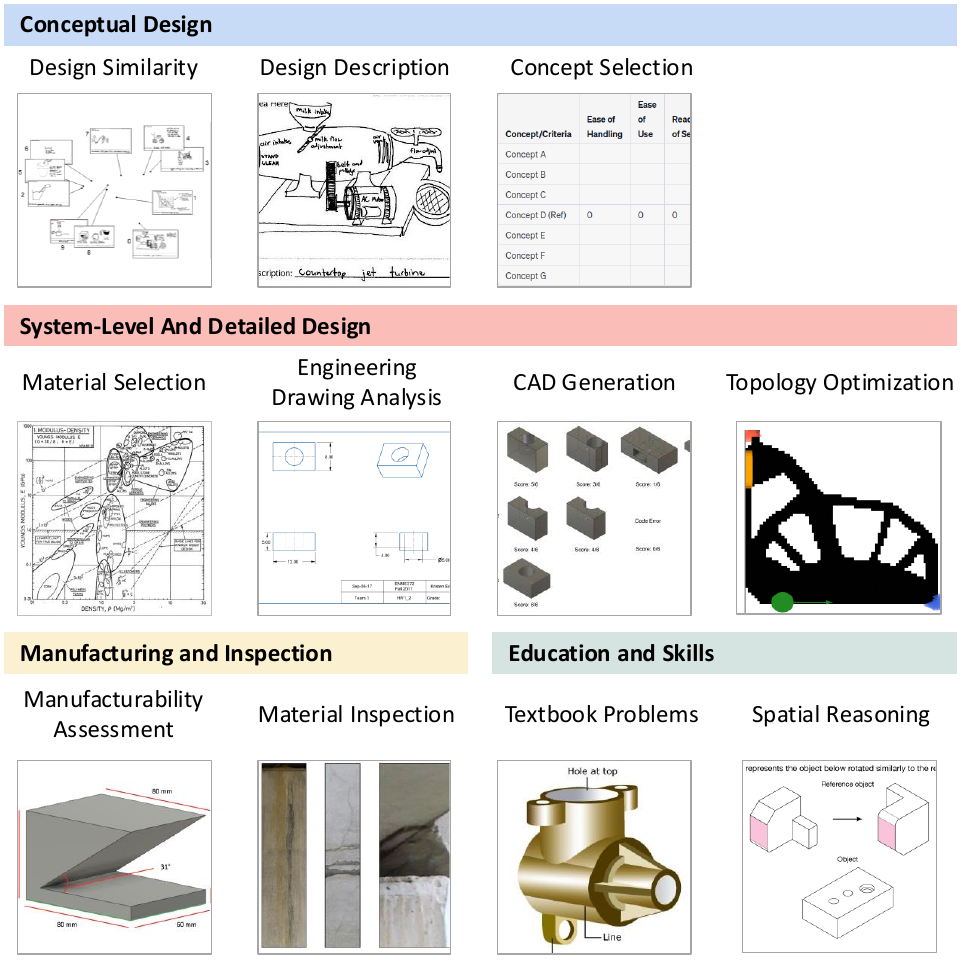}
    \caption{We explored GPT-4V and LLaVA~1.6~34B's ability to perform numerous engineering design tasks that utilize both visual and textual information. Panel ``Textbook Problems'' adapted from~\citep{OCW_2007} under CC BY-NC-SA 4.0. Panel ``Material Inspection'' adapted from~\citep{mundt2019meta} under its specific license.}
    \label{fig:contributions}
\end{figure}

Engineering design encompasses a broad range of tasks within the product design process, as shown in Fig.~\ref{fig:contributions}. These include:
\begin{enumerate*}[(1)]
    \item generating and selecting concepts,
    \item choosing between modular and integral structures,
    \item sizing components and selecting materials, and
    \item prototyping, manufacturing, and inspection.
\end{enumerate*}
There are several key challenges that engineers face during this process. These include dynamically changing technologies and markets, trade-offs between desired qualities, the need to process, analyze, and synthesize lots of information, and time pressure during the process which necessitates quick decisions often without complete information~\citep{ulrich_product_2020}. Large multimodal language models with vision capabilities may be well-posed to automate tasks and support engineers with these challenges. For example, such models could sweep through thousands of sketches stored in knowledge databases and enable engineers to search through them using natural language. It could also analyze CAD models as they are constructed to include design-for-manufacturing rules early on, potentially reducing the manufacturing challenges in later stages of the process.

\paragraph{Objectives and Contributions}
This work is a preliminary and broad exploration of the capabilities of VLMs for engineering design tasks that require textual and visual inputs and sets to establish standardized evaluation tasks. We select diverse engineering tasks that could be automated by VLMs and perform qualitative and quantitative analyses of GPT-4V, summarized in Fig.~\ref{fig:contributions}. We then use our quantitative benchmark to evaluate a leading open-source VLM, LLaVA~1.6~34B~\citep{liu2024llavanext}. We discuss our findings and implications for using VLMs within engineering design. We aimed, wherever possible, for larger sample sizes and quantitative analyses, despite the lack of API at the start of our investigation. Specifically, our overall contributions are:

\begin{itemize}
    \item We developed and performed quantitative experiments through more than 1000 queries to evaluate GPT-4V and create benchmarks for future VLMs related to (i) design similarity, (ii) early-stage sketch description, (iii) understanding of engineering drawing and CAD generation, (iv) topology optimization result analysis, (v) assessment of manufacturability, (vi) machining feature identification, (vii) defect identification, (viii) textbook problems, and (ix) spatial reasoning.
    \item We developed and performed qualitative case studies of GPT-4V's performance related to (i) early-stage sketch description generation, (ii) material selection, and (iii) topology optimization understanding.
    \item We created and released datasets for future evaluations, including the input images, input prompts, and answers for all eight quantitative experiments described above on the project webpage \url{https://decode.mit.edu/projects/vlms4design/}.
    \item We ran comparative experiments using the open-source VLM LLaVA~1.6~34B to demonstrate the use of these tasks and datasets for benchmarking VLMs in engineering design.
\end{itemize}

\paragraph{Outline of the work}
Section~\ref{sec:concepts} delves into conceptual design, focusing on design sketches and text descriptions. 
Section~\ref{sec:detailed} focuses on tasks related to the detailed design stage. In particular, we discuss material selection, engineering drawing analysis and CAD generation. We also use the specific case of topology optimization to the model's general understanding and the support it can offer for result analysis.  Section~\ref{sec:manufacturing} assesses the general knowledge of manufacturing and tests the performance on inspection tasks. Section~\ref{sec:education} investigates the performance of GPT-4V on textbook problems, providing some insights into its over-arching engineering knowledge and skills. Section~\ref{sec:spatial} investigates the performance of GPT-4V on spatial reasoning tests, looking into an important skill across design tasks. Finally, Section~\ref{sec:llava} compares the quantitative results from GPT-4V and LLaVA~1.6~34B, and Section~\ref{sec:discussion} offers a broader perspective of the capabilities of VLMs for engineering design and discusses its limitations.

\section{Experimental Setup}
We developed most of our prompts within the ChatGPT user interface, specifically harnessing the capabilities of the \emph{September 25, 2023 update}. It should be noted that the principal content of the paper and the primary experiments were conducted using this particular version\footnote{The majority of the research and experimental work for the initial version of this paper was carried out before November 2, 2023, leveraging GPT-4V released on September 25, 2023. 
}, with a specific focus on the vision feature. 
We do not include internet access, plugins, or a code interpreter unless explicitly mentioned (for example see~\ref{to:volume_fraction_estimation}). By the nature of using the ChatGPT web interface, we do not control or vary the ``temperature'' or ``top-k'' parameters of the responses. Each repeated experiment was run in a new chat, which we denote as \textit{context} in this work. When using the GPT-4V API and LLaVA~1.6~34B we used a temperature and top-k of 1.0.

Our methodology is outlined by the following key points:

\begin{enumerate}[(1)]
\item  \textbf{Emphasis on Images.} Our primary focus in utilizing VLMs is for image-based tasks. This is where our efforts and resources are concentrated, aiming to explore the model's understanding and analysis of visual data.

\item \textbf{Short Text Prompts.} We utilize short, straightforward prompts that prioritize vision tasks. This is to ensure a focus on visual analysis over the complexity of prompt engineering or creating lengthy custom instructions.

\item \textbf{Transparent Prompts and Answers.} In each section, we provide examples of our exact prompts and the exact response from the model, unless shortened with ellipses. 
\end{enumerate}

By maintaining these standards, we hope to provide clear experiments that can be used as benchmarks for future VLM evaluations in engineering design. Each section contains the additional methodological aspects related to each topic.

\section{Conceptual Design}\label{sec:concepts}
\paragraph{Overview and Motivation} Conceptual design is an early stage of the product development process, during which designers identify customer needs, establish target specifications, and then generate, select, and test many concepts~\citep{ulrich_product_2020}. Experts estimate that 70-80\% of a product's life-cycle costs are determined during the early design stages~\citep{CORBETT198693,pahl2007engineeringdesign}, highlighting the importance of decisions made during this stage. There exists a rich body of research examining concept generation during this stage~\citep{bryant2005generation, toh2019generation, das2022sketch}, creativity evaluation in conceptual design~\citep{amabile1982social, shah2003metrics, RN682, baer2019assessing, cseh2019scattered}, cost estimates and environmental assessments during conceptual design~\citep{saravi2008cost, FU2020407}, and, recently, multimodal machine learning in conceptual design~\citep{yuan2021multimodal,  edwards2021design, song2023aemultimodal, su2023multimodal}.

In conceptual design, two of the primary design modalities are hand-drawn sketches and textual descriptions. Often these modalities are combined, and early-stage designs are represented as a sketch with accompanying text. A vision language model (VLM) such as GPT-4V has great potential to be used as a tool during conceptual design. Some of the main tasks in conceptual design include generating sketches and descriptions of design ideas, comparing and iterating upon those ideas, and ultimately selecting a design to move forward with. In the following experiments, we explore how GPT-4V can act as an aide for these tasks. Specifically, we see how the multimodal capabilities enable GPT-4V to perform engineering tasks when both design sketches and text prompts are included as input. For effective multimodal learning, it is important to have designs with both image and text modalities, and sufficient datasets of these designs. 
There exist a few multimodal conceptual design datasets that we use in the following experiments. However, an overarching theme in machine learning within the engineering domain is that most datasets are small, which poses a challenge for data-driven models. Large pretrained models like GPT-4V can help overcome this challenge because they have been trained on a plethora of information (although the exact training information of GPT-4V is not yet released), meaning that they have general knowledge about the world. Another challenge is faced during concept selection. Experts suggest generating a multitude of conceptual designs and then down-selecting through design evaluation~\citep{ulrich_product_2020}; however, the evaluation step is often performed by experts, which takes time and resources~\citep{baer2019assessing}. GPT-4V may be able to help engineers during the conceptual design stage by utilizing general knowledge and sketch understanding to interpret and compare designs, move between design representations (text and image), and perform concept selection tasks. We aim to evaluate these capabilities in the following experiments.

\subsection{Design Similarity}
\label{cd:design_similarity}

Determining if two designs are similar is an important part of conceptual design. Assessing design similarity can act as a proxy for assessing design novelty, which is a common criterion in concept selection~\citep{ahmed2018triplets}. Novelty expresses that a concept is rare, ingenious, imaginative, or surprising, as well as radical or transformational~\citep{verhaegen2012originality}. However, novelty evaluation is often subjective. It can be easier for humans to articulate why they might rate designs as similar than why they would rate one design as more novel than another~\citep{ahmed2018triplets}. 
For this reason, past work has studied how humans assess similarity of concepts~\citep{ahmed2018triplets} as a method to build idea maps and identify novelty. Similarity comparisons can also help explore the design space by identifying clusters of similar ideas, potentially helping with faster design space exploration.
Recently, researchers have compared how human evaluations of similarity compare to computationally determined similarity, and found that they diverge based on the level of abstraction of a product~\citep{kosa2022triplets}.

Three main challenges arise with human evaluations of similarity:
\begin{enumerate}
\item \textbf{Evaluation speed and cost} Human evaluations are very expensive. Both the time and the cost of these evaluations are exacerbated as the number of designs increases since the number of similarity queries scales with $n^2$ the number of designs. 
    \item \textbf{Self-consistency} Humans may make different similarity assessments when they repeat the same evaluation.
    \item \textbf{Transitive violations} Given designs A, B, and C, one cannot say that A is more similar to B, B is more similar to C, but that C is also more similar to A. This would violate the transitive property of inequality, since if AB < AC (where AB is a measure of similarity between design A and design B) and BC < BA, then these imply that CB < CA, so saying CA < CB cannot be true. A violation of this sort can be tested when the same three designs, which we call a triplet, are assessed for similarity multiple times.
\end{enumerate} 

In the following experiments, we evaluate if GPT-4V can effectively assess the similarity of designs, i.e. with high self-consistency and a low number of transitive violations. We are replicating this experiment which was completed by human raters in~\citet{ahmed2018triplets}, and here novelty is defined as distance from other designs, and is not impacted by relevance to the design task. Our methodology, an example of our prompt, and the table of results are provided below.

\begin{figure}[ht]
    \centering
    \includegraphics[width=\linewidth]{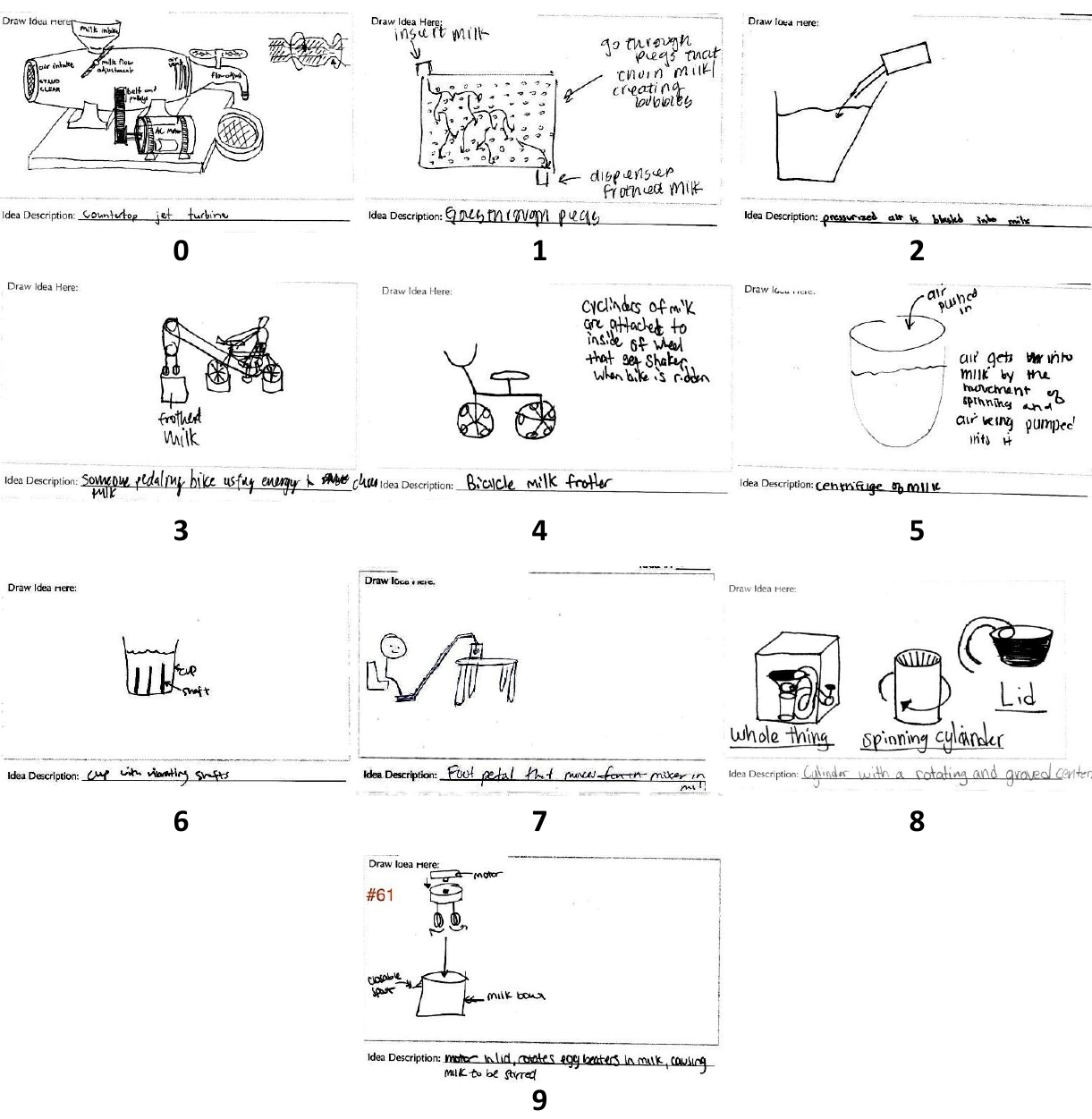}
    \caption{Ten conceptual designs of novel milk frothers. We task GPT-4V with assessing the similarity of these designs to one another. The handwritten descriptions at the bottom of each design are referred to as ``text descriptions.''}
    \label{fig:numbered_images}
\end{figure}

\begin{figure}
    
\begin{contextbox}[]{Design Similarity}
\begin{promptbox}{(Design Similarity)}
 Which design is more similar to design A?
    \begin{center}
    \includegraphics[width=0.6\linewidth,keepaspectratio]{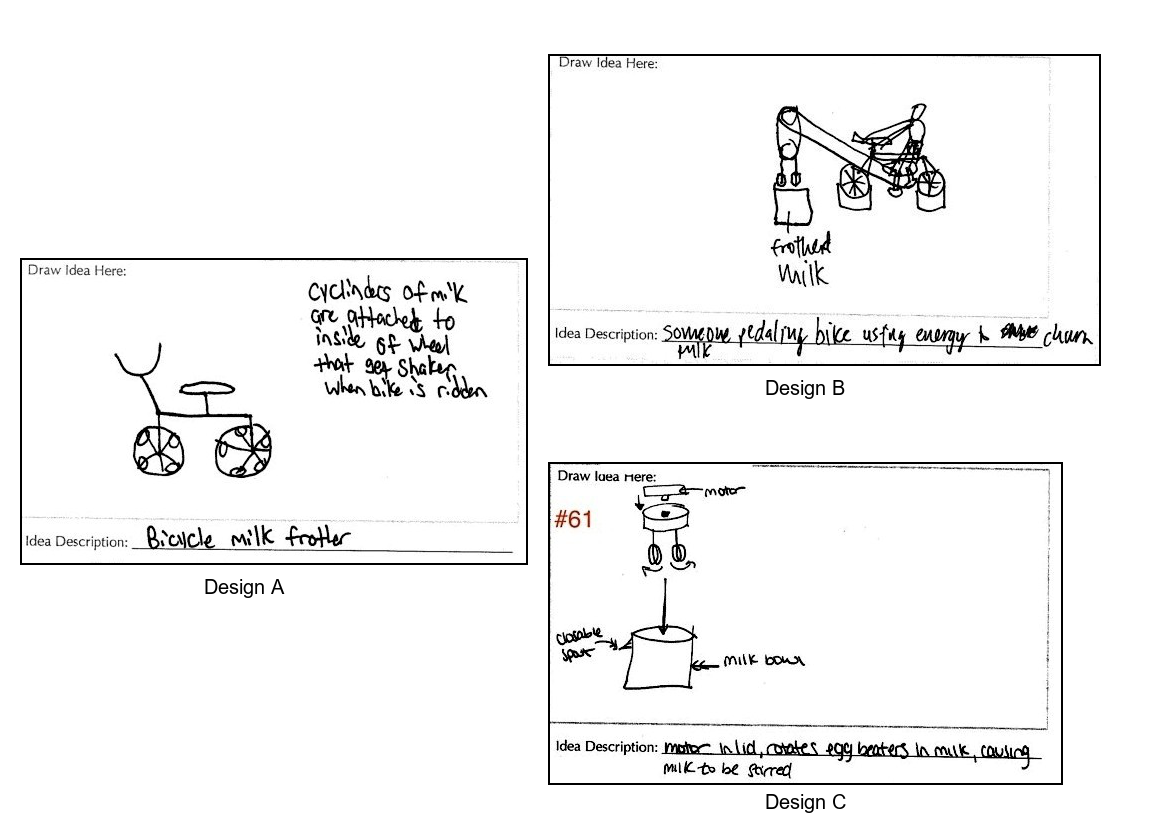}    
    \end{center}
    
\tcblower

Design B is more similar to Design A. \hl{Both involve bicycles and the concept of using pedaling or riding to froth milk. Design C is distinct as it involves a motor and an egg beater mechanism.}

\end{promptbox}
\end{contextbox}

    \caption{Assess design similarity.}
    \label{fig:design_similarity}
\end{figure}

\paragraph{Methodology} 
We tasked GPT-4V with performing the same experiment that eleven human raters performed in~\citet{ahmed2018triplets}. We have 10 early-stage design sketches from~\citep{STARKEY201647,miller2016design} as shown in Fig.~\ref{fig:numbered_images}. Each design is defined by a hand-drawn sketch, possibly annotated, and a handwritten description, referred to as ``text description.''  We group them into groups of three, which we call triplets. As shown in Fig.~\ref{fig:design_similarity}, we provide a triplet of design sketches labeled A, B, and C, and ask GPT-4V which design is most similar to A. Since we have 10 designs, we can make 360 triplets such that each design is considered design A for all 36 combinations of the other 9 designs as designs B and C. When given 360 of these triplet examples, we assessed if GPT-4V commits transitive violations. We then repeated 50 of the examples to evaluate GPT-4V's self-consistency.

\begin{table}[h]
\centering
\caption{A summary of the self-consistency and number of transitive violations of GPT-4V when evaluating 360 triplets for which designs are more similar. We compare the results with the corresponding values for eleven human raters reported in~\citet{ahmed2018triplets}. GPT-4V has higher self-consistency than the human raters and its number of transitive violations equals the lowest human rater value.}
\label{tab:triplet_similarity}
\begin{tabular}{@{}ccc@{}}
\toprule
\textbf{Rater} & \textbf{Self-consistency} & \textbf{Transitive Violations} \\ \midrule
1  & 91.6\%  & 5  \\
2  & 50.0\%  & 5  \\
3  & 83.3\%  & 5  \\
4  & 75.0\%  & 10 \\
5  & 58.3\% & 10 \\
6  & 58.3\% & 20 \\
7  & 41.6\% & 8  \\
8  & 41.6\% & 20 \\
9  & 58.3\% & 11 \\
10 & 75.0\% & 12 \\
11 & 58.3\% & 5  \\
\textbf{GPT-4V} & \textbf{94.0\%}                                & \textbf{5}                                          \\ \hline
\end{tabular}
\end{table}

\begin{figure}
    \centering
    \includegraphics[width=0.8\linewidth]{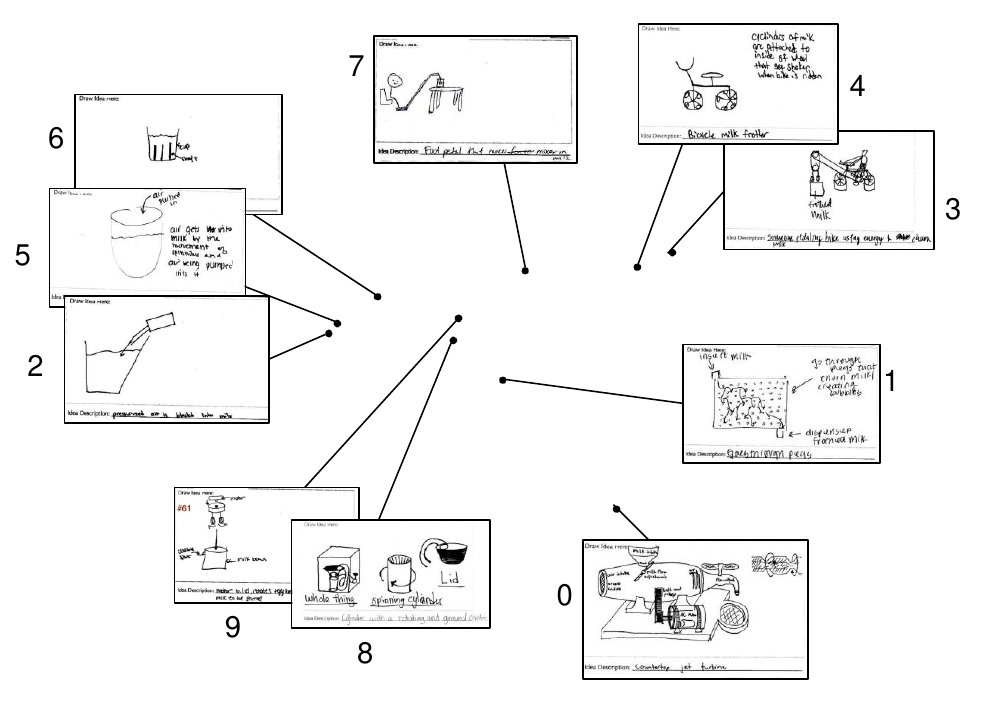}
    \caption{A map of the milk frother design sketches where sketches that are closer to each other are more similar. These are based on the responses by GPT-4V for 360 triplet similarity queries. We observe that the map clusters similar designs together and places unique designs further away from other designs.}
    \label{fig:similarity_scatter}
\end{figure}

In addition, we utilize the generalized non-metric multidimensional scaling (GNMDS) technique introduced by ~\citep{pmlr-v2-agarwal07a} to find 2-D embeddings of design sketches using triplet responses and generate a visualization of the ten designs where designs that are closer to each other are considered more similar. This technique is the same as used in~\citet{ahmed2018triplets} to generate a map of these designs from human ratings. The resultant map, referred to as the idea map of GPT-4V, is shown in Fig.~\ref{fig:similarity_scatter}. The idea map shows a striking resemblance to human-generated idea maps reported in literature. We observe that the three designs that show cups filled with milk are grouped together (designs 2, 5, and 6) as well as the two bicycle-based designs (designs 3 and 4). This clustering of similar designs was also observed in the idea map of all human raters combined, shown in Fig. 8 in~\citet{ahmed2018triplets}. GPT-4V's idea map also places Sketch 0 further away from all other sketches, denoting that it was perceived as most dissimilar. Coincidentally, the most novel sketch identified by the aggregated human ratings was also Sketch 0. Sketch 0, proposing a countertop jet turbine to froth milk was also the most novel sketch rated by the expert in their work. This serves as a validity check, demonstrating to us that GPT-4V's assessment of the similarity of sketches aligns with human ratings. We note that in past work, each human rater has a different map, and GPT-4V creates a unique map as well. The variability in individual human idea maps is likely influenced by diverse criteria for judging similarity. Consequently, establishing a definitive standard for sketch similarity is challenging. Therefore, we compared our results to the aggregated map by eleven human raters to gauge how GPT-4V's assessments conformed with collective human wisdom.

These initial findings pave the way for future research avenues. First, GPT-4V's capability to create idea maps is not only scalable, and ideal for evaluating a large number of design items, but also overcomes limitations faced by previous studies reliant on time-consuming human ratings. Second, the use of triplet queries for generating idea maps extends beyond sketches to other design forms like prototypes, 3D models, or multimedia, offering a novel approach to evaluating design similarity. These maps are valuable tools for designers, enabling them to better understand their design domain and leverage GPT-4V for more effective exploration of the design space.

\paragraph{Discussion: Design Similarity}

From the quantitative experiments using a total of 410 queries, with results summarized in Table~\ref{tab:triplet_similarity}, we find that GPT-4V is able to assess the similarity of designs with greater or equal performance as human raters. In the 360 trials, GPT-4V made only five transitive violations, which matches the lowest number of transitive violations made by any of the eleven human raters. Additionally, in our trials, GPT-4V was self-consistent 94\% of the time, which is greater than any of the human raters. A naive model could still be self-consistent without actually understanding much about a design sketch. So to gain insight into GPT-4V's similarity assessment, we plot the designs using the GNMDS embedding technique, shown in Fig.~\ref{fig:similarity_scatter}. We observe sensible clustering of three design sketches whose major features are cups with milk, and two design sketches whose major features are bicycles. 
While future work should explore how well VLMs perform at this task for other datasets, these results offer a promising suggestion that GPT-4V can effectively assess the similarity of conceptual design sketches. Future VLMs can be evaluated using these same methods to compare them against this version of GPT-4V. We have provided the dataset including all triplets for this purpose. These experiments were meant to test understanding, analysis, and evaluation. The results suggest that GPT-4V is able to understand and analyze design sketches in order to assess their similarity. Furthermore, assessing similarity among design triplets is a mundane and repetitive task. GPT-4V's ability to perform this task may mean that human raters do not have to, which can save time and resources both in dataset creation and in the design process.

\subsection{Design Descriptions}

Through these experiments, we aim to evaluate how well GPT-4V understands the different representations of a design, in this case, textual and sketch. Understanding sketches is a first step in being able to evaluate and compare them, which is one of the end goals of the conceptual design phase~\citep{ulrich_product_2020}. We task GPT-4V with matching a design to its correct description given a number of options, and we also task GPT-4V with generating textual descriptions of a design given just a sketch. We specifically chose to perform description matching (section \ref{cd:matching}) in the form of multiple-choice questions because this allows for quantitative analysis. Additionally, we provide the exact questions and results so that future VLMs can be similarly evaluated. In fact, we evaluate GPT-4V, and LLaVA~1.6~34B in this manner; results are shown in Tables \ref{tab:description_matching}, and \ref{tab:gpt_llava}. While these description matching and generating tasks do not directly translate to common engineering tasks during the design process, we believe that GPT-4V's performance on them sheds light on its ability to understand information from one modality (sketch) and then synthesize information in another modality (generating text). Furthermore, if a tool can automatically generate accurate and useful textual descriptions of conceptual design sketches, this could allow engineers to (1) create an easily-searchable catalog of early-stage designs, and (2) more easily generate multimodal datasets of paired sketches and text descriptions, which are necessary for multimodal machine learning in the engineering domain~\citep{song2023multi}. Automatically generating relevant textual descriptions for hand-drawn sketches can also help communicate design ideas to design team members and potential stakeholders, which is a primary role of sketching~\citep{das2022sketch}. Lastly, it can help human raters judge design ideas for creativity, novelty, quality, and other common design metrics~\citep{shah2003metrics}.

\subsubsection{Match a Description to a Design}
\label{cd:matching}

Given an image of an early-stage design sketch, and four different design description options, we test if GPT-4V can identify the correct description. We analyze GPT-4V's performance on these simple tasks to gain a basic understanding of whether more challenging description generation tasks are possible. Our methodology, four examples of our prompts, and a table of our results are provided below.

\begin{figure*}
\begin{contextbox}[width=0.49\textwidth,box align=top,before=,after=\hfill]{Design Descriptions}
\begin{promptbox}{(Matching a Design to its Description: Providing the full design including handwritten text.)}
Which text description matches the given design? 
\begin{enumerate}[A.,itemsep=2pt,topsep=2pt,parsep=0pt,partopsep=0pt]
    \item Motor in lid, rotates egg beaters in milk, causing milk to be stirred 
    \item Cylinder with a rotating and grooved center 
    \item Pressurized air is blasted into milk 
    \item None of the above
\end{enumerate}
    \begin{center}
    \includegraphics[width = 0.65\linewidth, keepaspectratio]{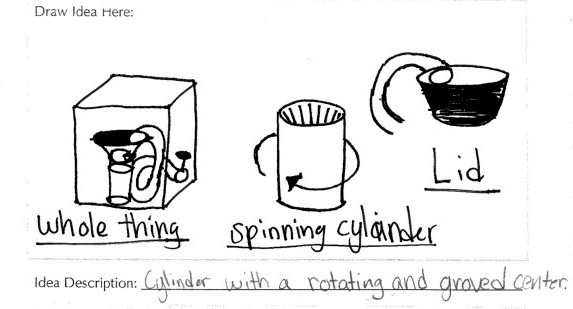}    
    \end{center}
\tcblower

\hl{B. Cylinder with a rotating and grooved center.}

\end{promptbox}
\end{contextbox}%
\begin{contextbox}[width=0.49\textwidth,box align=top,before=,after=]{Design Descriptions}
\begin{promptbox}{(Matching a Design to its Description: Providing the design, but removing the handwritten text description.)}
Which text description matches the given design? 
\begin{enumerate}[A.,itemsep=2pt,topsep=2pt,parsep=0pt,partopsep=0pt]
    \item Bicycle milk frother
    \item Motor in lid, rotates egg beaters in milk, causing milk to be stirred
    \item Goes through pegs
    \item None of the above
\end{enumerate}
    \begin{center}
    \includegraphics[width = 0.65\linewidth, keepaspectratio]{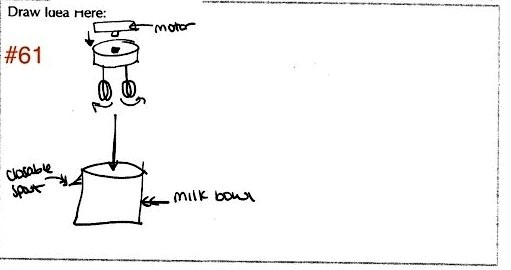}    
    \end{center}
    
\tcblower

\hl{B. Motor in lid, rotates egg beaters in milk, causing milk to be stirred.}

\end{promptbox}
\end{contextbox}

\vspace{2mm}
\begin{contextbox}[width=0.49\textwidth,box align=top,before=,after=\hfill]{Design Descriptions}
\begin{promptbox}{(Matching a Design to its Description: Providing the design, but removing the handwritten text description, and removing the ``None of the above'' option.)}
Which text description matches the given design? 
\begin{enumerate}[A.,itemsep=2pt,topsep=2pt,parsep=0pt,partopsep=0pt]
    \item Someone pedaling bike using energy to churn milk
    \item Cylinder with a rotating and grooved center 
    \item Motor in lid, rotates egg beaters in milk, causing milk to be stirred 
\end{enumerate}

\begin{center}
\includegraphics[width = 0.65\linewidth, keepaspectratio]{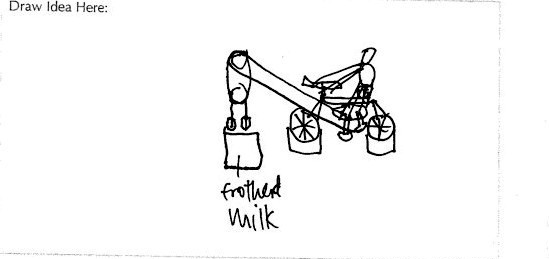}
\end{center}
    
\tcblower

\hl{A. Someone pedaling bike using energy to churn milk}
\end{promptbox}
\end{contextbox}
\begin{contextbox}[width=0.49\textwidth,box align=top,before=]{Design Descriptions}
\begin{promptbox}{(Matching a Design to its Description: Providing the design, but removing the handwritten text description, and removing the ``None of the above'' option.)}
Which text description matches the given design? 
\begin{enumerate}[A.,itemsep=2pt,topsep=2pt,parsep=0pt,partopsep=0pt]
    \item Cup with vibrating shafts
    \item Countertop jet engine
    \item Foot pedal that moves mixer in milk
\end{enumerate}
    \begin{center}
    \includegraphics[width = 0.65\linewidth, keepaspectratio]{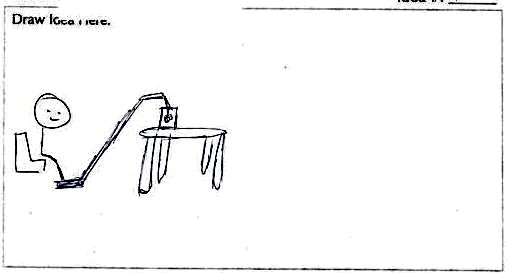}    
    \end{center}
    
\tcblower

D. A person sitting and looking at a computer on a table.

\hle{Note: None of the given options (A, B, C) match the provided design.}

\end{promptbox}
\end{contextbox}

    \caption{Match a design to the correct description.}
    \label{fig:design_description_prompts}
\end{figure*}

\paragraph{Methodology}
We assessed if GPT-4V can match a design sketch to its correct text description for three different cases:
\begin{enumerate}
    \item We provide the whole image including the handwritten text description, as well as four description options including ``None of the above''. Figure~\ref{fig:design_description_prompts} upper left.
    \item We provide the image with the handwritten text description removed, as well as four description options including ``None of the above''. Figure~\ref{fig:design_description_prompts} upper right.
    \item We provide the image with the handwritten text description removed, and only three text description options, removing the ``None of the above'' option. Figure~\ref{fig:design_description_prompts} lower left and lower right.
\end{enumerate}

 We ran 90 total queries: three trials each comprised of ten multiple choice questions for each of three different cases\footnote{Trial 1 used the \textit{September 25th update} of GPT-4V; Trials 2 and 3 used the \textit{November 6th update} of GPT-4V.} We included one multiple-choice question for each of the designs in Fig.~\ref{fig:numbered_images}. Examples of these questions are shown in Fig.~\ref{fig:design_description_prompts}. As the answer options, we provided three text descriptions from among the 10 design sketches. Table~\ref{tab:description_matching} displays the results for each of the three cases, as well as the final score out of 10. 

\begin{table}[ht]
\caption{Results for the three multiple-choice design description matching experiments. 90 queries were run across three trials, ten queries for each of the three cases. The full results for Trial 1 are displayed, as well as the scores for all three trials.}
\label{tab:description_matching}
\begin{tabularx}{\linewidth}{@{}lYYYY@{}}
\toprule
  &%
  \textbf{Correct Answer} &
  \textbf{With Text Description } &
  \textbf{No text description} &
  \textbf{No text description, no ``None of the above''} \\ \midrule
0              & B         & B              & B             & B             \\
1              & A         & A              & A             & A             \\
2              & C         & C              & D             & B             \\
3              & A         & A              & A             & A             \\
4              & B         & B              & B             & B             \\
5              & C         & C              & D             & B             \\
6              & C         & C              & D             & C             \\
7              & C         & C              & D             & D            \\
8              & B         & B              & A             & A             \\
9              & B         & B              & B             & B             \\ \midrule
\multicolumn{2}{@{}l}{\textbf{Trial 1 Score}} & \textbf{10/10} & \textbf{5/10} & \textbf{6/10}
\\

\multicolumn{2}{@{}l}{\textbf{Trial 2 Score}} & \textbf{10/10} & \textbf{5/10} & \textbf{8/10} 
\\

\multicolumn{2}{@{}l}{\textbf{Trial 3 Score}} & \textbf{10/10} & \textbf{6/10} & \textbf{7/10}
\\ \midrule

\multicolumn{2}{@{}l}{\textbf{Average}} & \textbf{100\%} & \textbf{53.3\%} & \textbf{70\%}
\\ \bottomrule
\end{tabularx}%

\end{table}

GPT-4V perfectly matches designs to their descriptions when the handwritten description is provided in the sketch. When the handwritten description is removed, GPT-4V's errors are typically a result of selecting ``None of the above,'' and when that option is removed, and it is forced to select a description, its performance increases (from 5.33/10 to 7/10). These results, combined with the results of LLaVA 1.6 (see Table~\ref{tab:gpt_llava}), suggest that the presence of textual context in sketches significantly enhances model accuracy, underscoring the importance of integrating text with visual data. The variability in model performance, particularly GPT-4V's superiority in scenarios without handwritten descriptions, suggests that model choice should be tailored to specific task requirements. Additionally, GPT-4V's tendency to select ``None of the above'' when uncertain highlights its cautious approach in ambiguous situations, reflecting a strategy to manage uncertainty. This behavior, along with the contrasting error patterns between GPT-4V and LLaVA 1.6, points to the need for deeper understanding and improvement in how different models process and interpret visual information, especially in the absence of textual cues. These findings are crucial for optimizing the use of VLMs in conceptual design contexts.

\subsubsection{Generate a Design Description from a Sketch}
\label{cd:generating_desc}

Given an image of an early-stage design sketch, can GPT-4V generate a relevant and accurate design description?

\begin{figure}
\begin{contextbox}[]{Design Descriptions}

\begin{promptbox}{(Generating a Design Description)}

A student was asked to develop a new innovative product that froths milk in a short amount of time. Please describe this design.
    \begin{center}
    \includegraphics[width=0.5\linewidth, keepaspectratio]{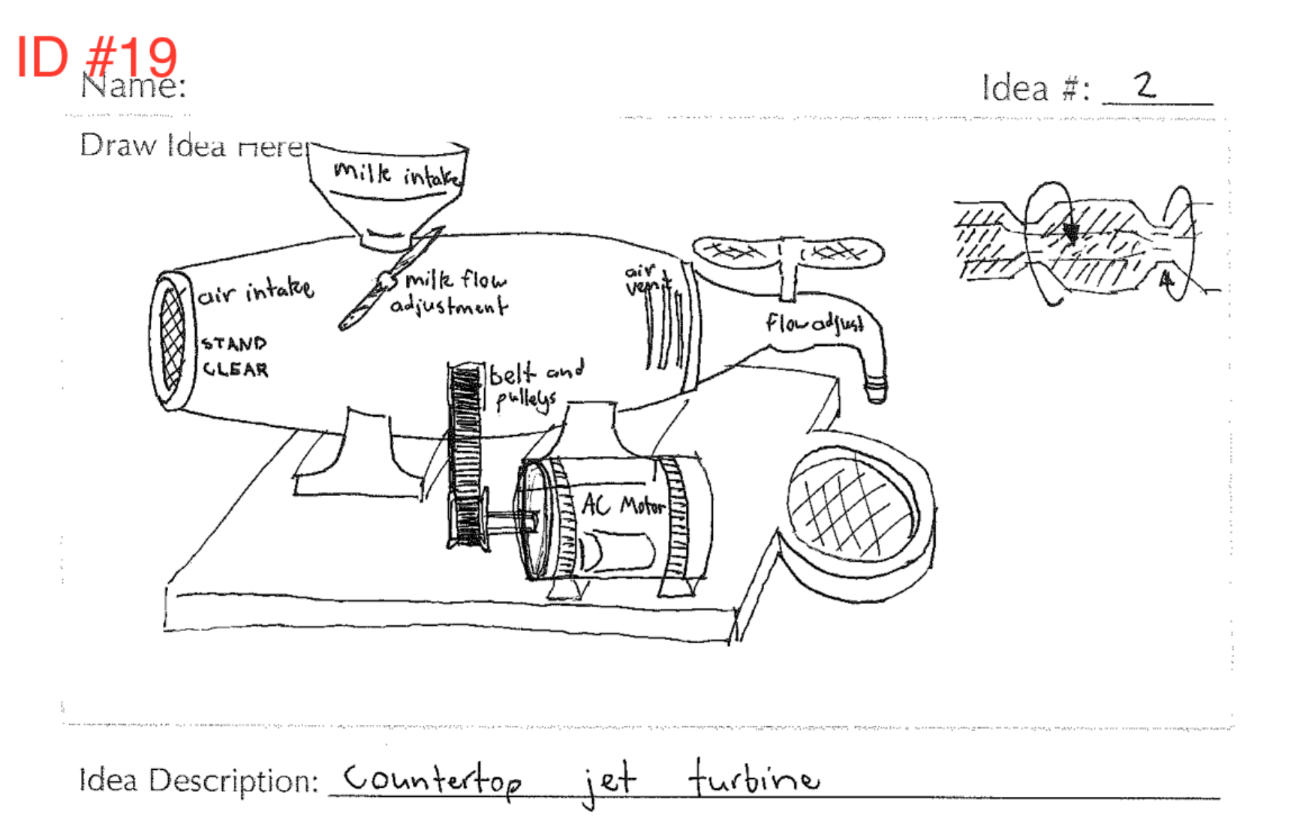}    
    \end{center}
    
\tcblower

The design presented is titled ``Countertop Jet Turbine.'' It appears to be a device intended to froth milk quickly using a turbine mechanism. Here's a breakdown of its features:

\begin{enumerate}[labelindent=0mm,leftmargin=*]
    \item \textbf{Milk Intake:} This is where milk is poured into the device. \hl{It seems to be designed in a funnel shape to ensure easy pouring without spillage.}
    \item \textbf{Air Intake:} Positioned on the side of the main body, this feature likely serves the purpose of drawing in air to aid in the frothing process.
    \item \textbf{Milk Flow Adjustment:} \hl{An adjustment knob or switch appears to be present, which probably allows the user to control the flow of milk or the intensity of frothing.}
    \item \textbf{Air Vent:} Located on the top side, the air vent might serve to release any excess air or to control the pressure within the device.

    \item \textbf{Flow Adjust:} Positioned near the air vent, this seems to be another control, possibly to manage the flow of frothed milk out of the device or to adjust the consistency of the froth.

    \item \textbf{Belt and Pulleys:} This part seems to be connected to an AC Motor. \hl{The belt and pulley system likely drives the turbine mechanism, converting the motor's rotational movement into the desired frothing action.}

    \item \textbf{AC Motor:} The power source of the device. \hl{This motor would drive the turbine at high speeds to froth the milk.}

    \item \textbf{Safety Warning - ``STAND CLEAR'':} This warning label suggests that the device might operate at high speeds or produce a forceful stream of frothed milk, and users should be cautious when it's running.
\end{enumerate}

\hl{In summary, the ``Countertop Jet Turbine'' appears to be a high-powered milk frother that utilizes a jet turbine mechanism to aerate and froth milk in a short amount of time. The multiple adjustment controls suggest that users have a good degree of control over the frothing process.}

\end{promptbox}
\end{contextbox}
    \caption{Generate a design description.}
    \label{prompt:generate_description}
\end{figure}

We performed this experiment for five early-stage design sketches with varying drawing scores. The drawing scores are based on a Consensual Assessment Technique evaluation of students' milk frother designs~\citep{STARKEY201647, miller2016design}. The scores can range from 1-7, but within the dataset of sketches, the scores range from 1-6. Table~\ref{tab:generate_descriptions} shows the results for these design sketches. The selected sketches were chosen at random from among all sketches with a similar drawing score. We show how GPT-4V responds when simply asked to describe the design, versus when provided with a description of the original design task given to the students and then asked to describe the design. For brevity, in both cases, we prompt GPT-4V to respond in three sentences. The exact prompts are provided in the column headers of Table~\ref{tab:generate_descriptions}.

\begin{table*}[ht]
    \caption{GPT-4V generated descriptions from design sketches. We also include the expert-rated drawing score of each sketch. The sketches are ordered by descending drawing score.}
\label{tab:generate_descriptions} 
    \centering
    \footnotesize
    \begin{tabular}{@{} b{3cm} m{1.2cm} b{6.2cm} b{6.2cm} @{}}%
        \toprule
        \textbf{Image} & \textbf{Drawing Score} & \textbf{Please describe this design in 3 sentences.} &\textbf{A student was asked to develop a new innovative product that froths milk in a short amount of time. Please describe this design in 3 sentences.}\\ 
        \midrule
\includegraphics[width=3cm]{countertopjetengine_19_2.png}
   & 5.5 &%
  The design depicts a countertop jet turbine. The apparatus features components such as an \hl{air intake, a milk intake, milk flow adjustment, an AC motor connected through belt and pulleys, and an air vent. Safety measures are indicated with a ``STAND CLEAR'' label near the air intake, and a flow adjustment knob is situated atop the device.} & %
  The design depicts a ``Countertop Jet Turbine'' that \hl{rapidly froths milk. It features an air intake for drawing in air, a milk intake for pouring in the milk, and adjustable settings for milk flow and air venting.} The device operates using \hl{an AC motor with belts and pulleys to generate the required frothing action, and it comes with a warning to ``STAND CLEAR'' during operation.} \\ 
  \midrule
\includegraphics[width=3cm]{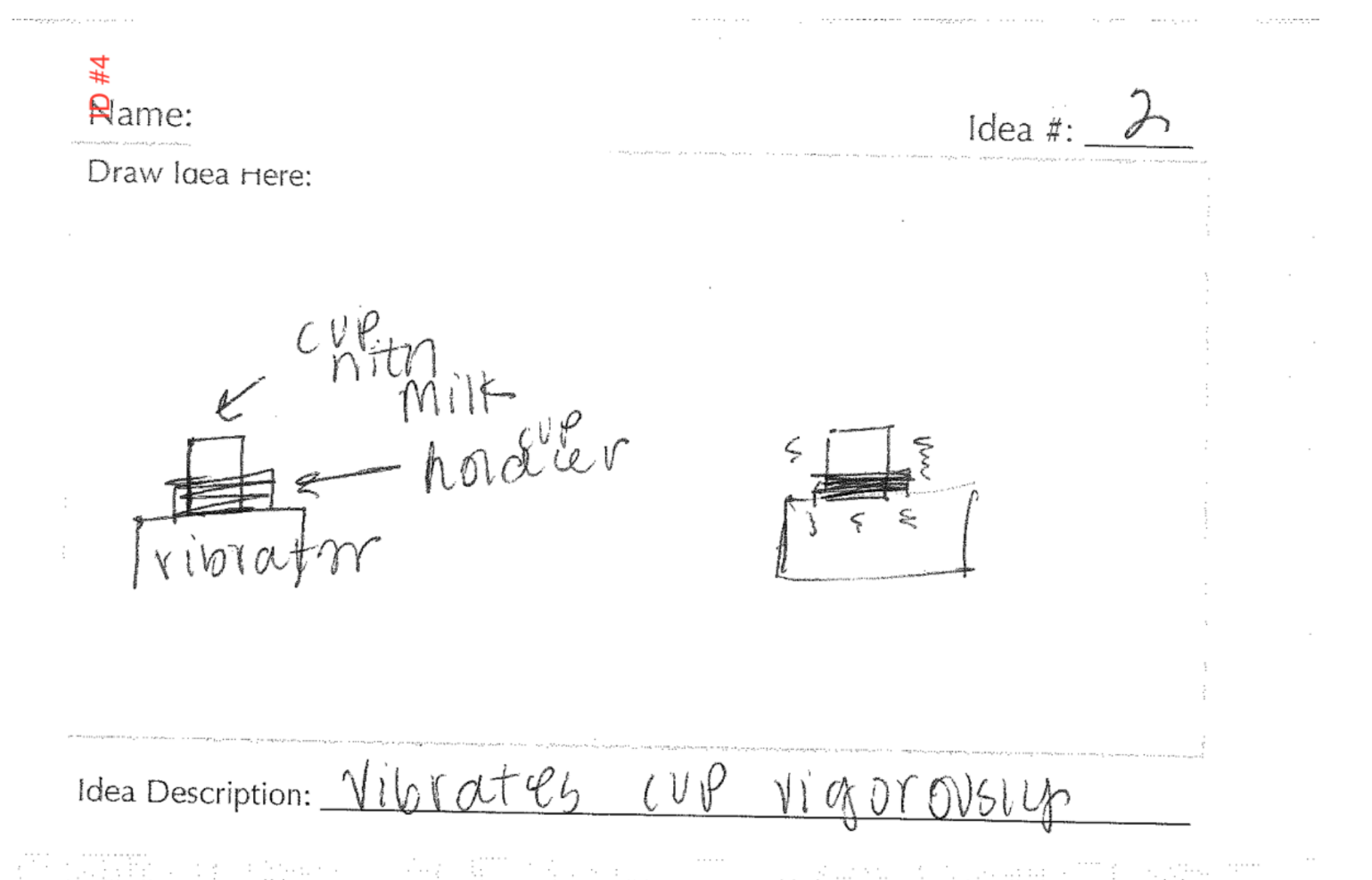}
   & 3.5 &
  This design illustrates a concept for \hl{a cup with a built-in vibration mechanism. The cup features an attachment or compartment labeled ``vibrator,'' suggesting it can shake or stir its contents.} Additionally, there's \hl{an emphasis on the cup being a ``milk holder,'' possibly indicating its primary use for beverages like milk or shakes.} &
  
  The student's design presents a ``Vibrates Cup Vigorouscup,'' which appears to be \hl{a cup with an integrated vibrator mechanism.} This mechanism is \hl{designed to rapidly froth milk when it's placed inside the cup.} The design also features a \hl{milk holder, ensuring that the milk is contained and frothed efficiently within the designated area.} \\ 
  \midrule
\includegraphics[width=3cm]{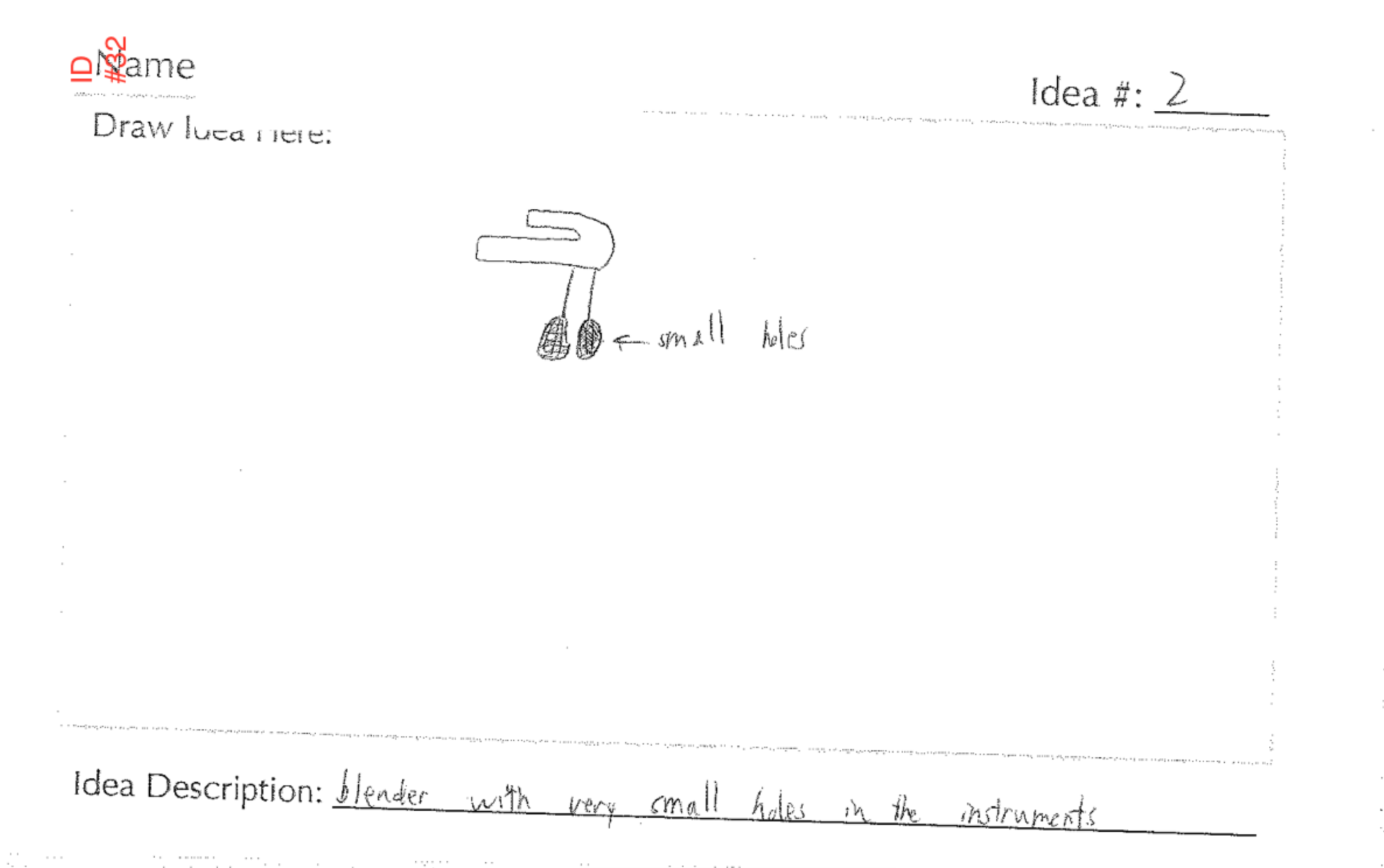}
    & 1.5 &
  This is a sketch of an innovative design labeled ``Idea \#2.'' The drawing depicts \hl{a blender-like device that has small holes in its instruments, possibly for fine blending or sieving.} The accompanying description suggests \hl{it's a blender with very small holes in the instruments, which might provide a unique blending mechanism or texture to the end product.}&
  The design showcases \hl{a unique blender specifically tailored for frothing milk.} It features \hl{a specialized blending apparatus with very small holes, ensuring thorough frothing within a short duration. The compact and ergonomic design of the blender ensures efficient milk frothing while minimizing spillage.} \\
        \bottomrule
    \end{tabular}
    
\end{table*}

\paragraph{Discussion: Design Descriptions}

In this section, we aimed to assess GPT-4V's ability to match different representations of an early-stage conceptual design and to generate one representation from another. The tasks we generated to explore these capabilities were: match a design sketch with its correct textual description, and generate a textual design description from a sketch. For each of these tasks, we gave various forms of the sketch to understand how the amount of handwritten text and the drawing skill in each sketch would affect these results. The quantitative results from the description matching experiment, shown in Table~\ref{tab:description_matching}, provided a basic understanding of whether the later description generation tasks were possible at all. The results showed that given an entire design sketch including the handwritten text description, GPT-4V was able to match the sketch to the text description for 10/10 of the questions across all three trials. This result essentially just assured us that GPT-4V could comprehend the hand-written text in these drawings. 

With this point verified, we next tested description matching if we removed the handwritten text description from the image. In this case, GPT-4V was able to match the sketch to the text description for 5.33/10 of the questions on average. While this is still better than random chance, which would be an average of 2.5/10, this result demonstrated how important providing both modalities, text, and image, is in this design stage. We noticed that many of the incorrect answers were GPT-4V selecting the ``None of the above'' option, suggesting that none of the descriptions matched the design. In fact, 4/5 of the incorrect answers for Trial 1 and 2, and 3/4 incorrect answers for Trial 3 occurred this way.  This is sensible to us, as the design sketches are often visually simple compared to their textual descriptions. An example of this is design 5 in Fig.~\ref{fig:numbered_images}, which visually looks like a cup of milk, but whose text description is ``Centrifuge of milk.'' With these results in mind, we tested how well GPT-4V would match the sketch to the description if we removed the ``None of the above'' option. This led to an improved average of 7/10 correct. One interesting example is shown in the lower right image of Fig.~\ref{fig:design_description_prompts}, in which GPT-4V generated its own option because the model determined that the sketch did not match any of the provided options, despite one being correct. 

We further explored GPT-4V's capability to generate textual descriptions from design sketches. The results are shown in Table~\ref{tab:generate_descriptions}. We explore how well the model generates design descriptions for three designs with different ``drawing scores,'' or levels of perceived drawing ability. For each design, we task GPT-4V with generating a description using two different prompts, one with more information:
\begin{enumerate}
    \item Please describe this design in 3 sentences.
    \item A student was asked to develop a new innovative product that froths milk in a short amount of time. Please describe this design in 3 sentences.
\end{enumerate}

Qualitatively, we assess that the model is able to generate useful and accurate text descriptions of designs even for sketches with very low drawing scores. We highlight in green the parts of the description that we believe showed understanding and relevance. GPT-4V extrapolates beyond just the given text to describe both the form and the function of the designs. This result once again highlights the importance of having both the text and image modalities in the design, as the generated descriptions pull heavily from the handwritten text to contextualize the sketch and explain how the design might function. Figure~\ref{prompt:generate_description} demonstrates GPT-4V's ability to understand, infer, extrapolate, and generate information about a design from its sketch. For example, the sketch includes a labeled belt and pulley system, and GPT-4V includes this in the generated description: ``\textbf{Belt and Pulleys:} This part seems to be connected to an AC Motor. The belt and pulley system likely drives the turbine mechanism, converting the motor’s rotational movement into the desired frothing action.'' We assess that GPT-4V is able to effectively generate textual descriptions when provided with a detailed sketch including handwritten text. This is true for both prompt types, the one that just requests ``Please describe this design in 3 sentences,'' as well as the prompt that provides additional information about the task. We do not observe a difference in the design descriptions that result from these different prompts, but further research could explore this more. It is important to note that LLMs, like GPT-4V, may hallucinate. This means GPT-4V may perceive patterns or objects in the input that are imperceptible to humans, producing outputs that do not make sense. Engineers should perform checks on generated descriptions, and additional research should be done regarding generating text descriptions for sketches in many domains. However, from our results, GPT-4V is able to generate sensible text descriptions from early-stage sketches. This ability may help engineers (1) create an easily-searchable catalog of early-stage designs, and (2) more easily generate multimodal datasets of paired sketches and text descriptions.

\subsection{Concept Selection}

A core component of conceptual design is concept selection~\citep{okudan2009concept, miller2020selection}. There are various concept selection methods, ranging from those based on decision matrices, to uncertainty modeling, to heuristics~\citep{okudan2009concept}. One of the most widely used concept selection methods for engineers is the Pugh Evaluation Matrix, sometimes called the Pugh chart~\citep{pugh1991total, pugh1995concept}. 

A Pugh Chart, also known as a Pugh Matrix, is a decision-making tool used in engineering and design. It involves comparing multiple options against a set of criteria, using a baseline for reference, to determine the most suitable choice. Each option is scored based on how well it meets each criterion compared to the baseline, facilitating an objective evaluation of alternatives.
The first step in creating a Pugh chart is defining selection criteria, which will be used to evaluate and compare concepts. The method may vary, but a common practice is to then select a benchmark design, and score all other designs qualitatively based on how they compare to the benchmark for each of the selection criteria. 

In the following experiments, we explore GPT-4V's ability to both generate selection criteria given a design prompt, and use concept selection methods on design sketches. We utilize a case study presented in the Concept Selection chapter of \textit{Product Design and Development} by~\citep{ulrich_product_2020}. This case study provides a design task: 
\begin{itemize}
    \item[] ``A medical supply company retained a product design firm to develop a reusable syringe with precise dosage control for outpatient use.'' 
\end{itemize}

Seven design sketches of reusable syringes are also included along with a Pugh chart and selection criteria. We use these as qualitative benchmarks with which we compare GPT-4V's ability to generate selection criteria and a Pugh chart.

\subsubsection{Generating Selection Criteria}
\label{cd:selection_criteria}

In Fig.~\ref{fig: concept_selection_criteria}, we explore if GPT-4V can generate selection criteria that would be used to evaluate designs, given a description of a design task.
As a baseline to assess the generated criteria, we utilized the selection criteria and Pugh chart provided in~\citep{ulrich_product_2020}; these were made for this design task and used the same seven design concepts that we used. Table~\ref{tab:gpt_selection_criteria} shows the ground truth selection criteria from the textbook, and the matching selection criteria generated by GPT-4V (if applicable). As seen in Table~\ref{tab:gpt_selection_criteria}, GPT-4V generated matching selection criterion for each of the ground truth criterion, sometimes via multiple criteria. For example for the ground truth criterion ``Ease of manufacture,'' the two matching criteria generated by GPT-4V are ``Cleaning and Sterilization: Ease of disassembly and reassembly for cleaning'' and ``Scalability: Design should be adaptable for mass production.''

\begin{figure}
\begin{contextbox}[]{Concept Selection}
\begin{promptbox}{(Generating Selection Criteria)}
A medical supply company retained a product design firm to develop a reusable syringe with precise dosage control for outpatient use. Please list the selection criteria that should be used to perform early-stage concept selection.

\tcblower

Certainly! \hl{Early-stage concept selection for a reusable syringe with precise dosage control for outpatient use involves considering various factors to ensure safety, efficiency, and user-friendliness.} Here's a list of selection criteria that should be considered:

\hl{\textbf{Safety and Biocompatibility:}}
        \begin{itemize}
        \setlength\itemsep{-0.3em}
            \item Material used should be biocompatible and non-reactive with the medication.
            \item The design should prevent overfilling and accidental discharge.
            \item Features to prevent cross-contamination, especially since it's reusable.
        \end{itemize}

\hl{\textbf{Ease of Use:}}[...]

\hl{\textbf{Precision and Accuracy:}}[...]

\hl{\textbf{Durability and Reusability:}}[...]

\hl{\textbf{Cleaning and Sterilization:}}[...]

\hl{\textbf{Cost-effectiveness:}}[...]

\hl{\textbf{Compatibility:}}[...]

\hl{\textbf{Environmentally Friendly:}}[...]

\hl{\textbf{Storage and Portability:}} [...]

\hl{\textbf{Scalability:}}[...]

\end{promptbox}
\end{contextbox}
    \caption{Generating selection criteria for a design task.}
    \label{fig: concept_selection_criteria}
\end{figure}

\begin{table*}[ht]
\centering
\caption{Selection criteria provided by the \textit{Product Design and Development} textbook and the GPT-4V equivalent.}
\label{tab:gpt_selection_criteria}
\begin{tabularx}{\textwidth}{@{}l X@{}}
\toprule
\multicolumn{1}{@{}l@{}}{\textbf{Textbook Selection Criteria}} & \multicolumn{1}{@{}l@{}}{\textbf{~~~GPT-4V Equivalent}}                                                                                                                                              \\ \midrule
Ease of handling                                     & Ease of use: \textit{Ergonomic grip and handling}                                                                                                                                                   \\
Ease of use                                         & Ease of use                                                                                                                                                                                 \\
Readability of dose settings                        & Ease of use: \textit{Clear markings and indicators for dosage                                    }                                                                                                   \\
Dose metering accuracy                              & Precision and accuracy                                                                                                                                                                      \\
Durability                                          & Durability and reusability                               \\
Ease of manufacture                                 & Cleaning and Sterilization: \textit{Ease of disassembly and reassembly for cleaning} Scalability: \textit{Design should be adaptable for mass production} \\
Portability                                          & Storage and Portability                                                                                                                                                                     \\ \bottomrule
\end{tabularx}
\end{table*}
\subsubsection{Creating a Pugh Chart}
\label{cd:pugh_chart}

\begin{figure}
\begin{contextbox}[]{Concept Selection}
\begin{promptbox}{Generating a Pugh Chart}
Given concepts A-G and the following selection criteria, make a Pugh chart to select a concept. The reference concept is concept D. 

Selection criteria: Ease of handling, Ease of use, Readability of settings, Dose metering accuracy, Durability, Ease of manufacture, Portability.

    \begin{center}
    \includegraphics[width =0.8\linewidth,keepaspectratio]{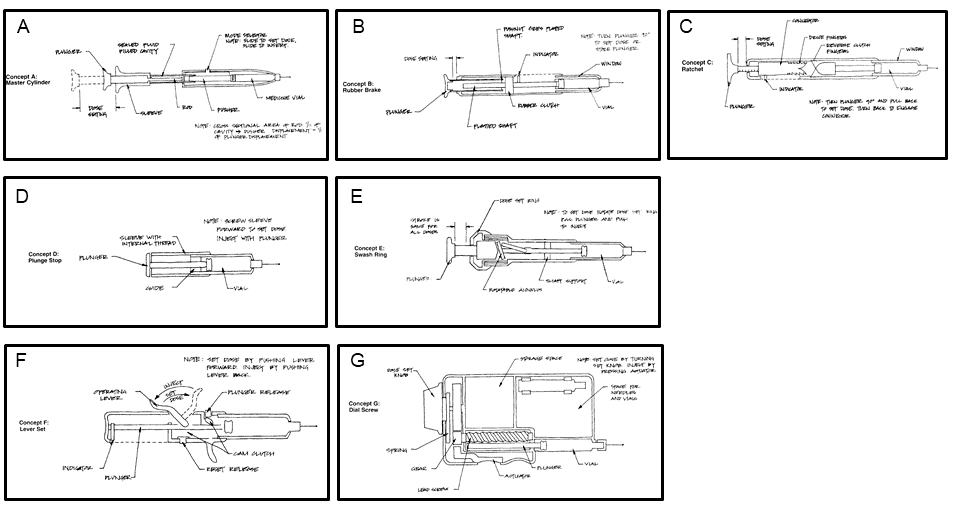}    
    \end{center} 
{\tiny
\textit{Images reproduced with permission from the authors.}}    
\tcblower

\hl{Creating a Pugh chart involves comparing various concepts against a reference concept across different selection criteria, and assigning scores to indicate whether the concept is better, worse, or the same as the reference.}

Here's an explanation on how to create a simplified Pugh chart, given the selection criteria you've provided:

\begin{enumerate}[labelindent=0mm,leftmargin=*]
    \item \textbf{Set up the matrix:} \hle{List the concepts (A-G) in rows and the criteria (ease of handling, ease of use, etc.) in columns.}
    \item \hl{\textbf{Choose a reference concept:}} In this case, Concept D is the reference and should be marked with a ``0'' (zero) score across all criteria since it's the standard for comparison.
    \item \textbf{Score the concepts:} \hl{Evaluate each concept against the reference for each criterion. You can use ``+1'' for better than the reference, ``0'' for the same as the reference, and ``-1'' for worse than the reference.}
    \item \textbf{Total the scores:} Sum the scores for each concept.
\end{enumerate}

Since I can't create a visual Pugh chart directly within this platform, I'll outline the matrix in text form:
 [\textit{Generated an empty table with concepts as the row headers and selection criteria as the column headers. The Concept D row was filled with zeroes.}]
\end{promptbox}
\end{contextbox} 
    \caption{Generating a Pugh Chart for concept selection.}
    \label{fig: concept_selection_pugh}
\end{figure}

In Fig.~\ref{fig: concept_selection_pugh}, we explore GPT-4V's ability to perform concept selection by generating a Pugh Chart. In particular, given selection criteria and images of various designs, can GPT-4V analyze and evaluate the designs and format this evaluation in a Pugh chart? The results are discussed in the following section. 

\paragraph{Discussion: Concept Selection}
Through these experiments, we explored GPT-4V's ability to perform two common concept selection tasks: generating selection criteria given a design task and making a Pugh chart given several design concepts. We found that GPT-4V was able to assess a design task and generate many relevant selection criteria, as shown in Fig.~\ref{fig: concept_selection_criteria}. For example, given this design task ``A medical supply company retained a product design firm to develop a reusable syringe with precise dosage control for outpatient use,'' GPT-4V generated selection criteria such as ``Safety and Biocompatibility,'' ``Ease of Use,'' and ``Precision and Accuracy.'' These criteria highlight that the design must be user-centered and safe in healthcare settings. 

Table~\ref{tab:gpt_selection_criteria} shows the baseline selection criteria provided for the design task from~\citep{ulrich_product_2020}, as well as the equivalent selection criteria generated by GPT-4V, if applicable. We observe that for each of the seven criteria that the baseline Pugh chart used, GPT-4V outputs an equivalent criterion. It is important to note that some of GPT-4V's equivalent criteria were subcategories, such as GPT-4V's ``Ease of Use: Clear markings and indicators for dosage,'' which mapped to the textbook's ``Readability of dose settings.'' These results demonstrate that GPT-4V is able to generate many relevant selection criteria, but that an engineer should still read the raw output and select relevant criteria as well as separate subcategories of certain criteria. 

When tasked with creating a Pugh chart given selection criteria and several designs, we found that GPT-4V understood what a Pugh chart was and how to generate one, however, it was often reluctant to create one given the limited information, as shown in Fig.~\ref{fig: concept_selection_pugh}. GPT-4V was able to generate an empty Pugh chart with the correct matrix format (not shown in Fig.~\ref{fig: concept_selection_pugh}, which switched the typical rows and columns), and also understood that it would be filled with qualitative comparisons of concepts, with one reference concept. However, it would only fill in the Pugh chart with hypothetical values given the lack of information about each concept. For example, in one trial, GPT-4V stated ``Since I cannot physically interact with the concepts to evaluate them against the criteria you've provided, such as "Ease of handling" or "Durability", I am unable to make informed decisions about the scores. These criteria often require subjective assessments or empirical testing that cannot be performed without direct interaction with the designs.'' In every instance of running the task in Fig.~\ref{fig: concept_selection_pugh}, GPT-4V refused to fill in the Pugh chart with anything other than random hypothetical values. Perhaps if an engineer provided more information about each concept, GPT-4V would have been able to generate an accurate Pugh chart, however, it failed to do so within our task format. 

Overall, our findings suggest that GPT-4V can be potentially effective in assisting human designers in identifying key factors that should be considered in the design process. However, while GPT-4V can generate equivalent criteria to those used in traditional methods, its outputs may need refinement, such as categorizing subcriteria. In terms of creating a Pugh chart, GPT-4V understands the concept and can format the chart correctly, but its reluctance to fill in the chart without extensive information indicates a limitation. This suggests that while GPT-4V can be a useful tool for structuring and initiating the concept selection process, human input remains crucial for detailed analysis and decision-making. 
The implications for practitioners are clear: VLMs such as GPT-4V can be a valuable aid in the initial stages of design concept evaluation, but they may require careful oversight and additional information to realize their full potential in more complex decision-making tasks.

\subsection{Summary} A concise summary of our assessment areas for GPT-4V's conceptual design abilities, and our findings for each are provided below.

\begin{enumerate}[(i)]
    \item \textbf{Assessing Design Similarity~\ref{cd:design_similarity}} 
    How does GPT-4V's consistency in assessing design sketch similarity compare to human benchmarks?
    \begin{itemize}
    \item We measure consistency using two measures from~\citep{ahmed2018triplets}--- self-consistency and transitive violations in assessing sketch triplet queries.
    The model is able to assess design similarity with higher self-consistency than human raters (94\% compared to 62.8\% average for human raters) and as few transitive violations as the top human raters.
    \end{itemize}
    \item \textbf{Matching Design Representations~\ref{cd:matching}} 
    Can GPT-4V accurately match design sketches with their text descriptions under varying information conditions?
    \begin{itemize}
        \item We ran 90 queries: three trials each comprised of ten multiple choice questions for each of three different cases. We found the following results. When provided the entire design sketch including a handwritten description, the model matched a design sketch to its appropriate text description 10/10 times for all three trials, however with the handwritten description removed, the score dropped to an average of 5.33/10, which we can compare to a score of 2.5/10 for randomly matching. Incorrect answers were often in the form of choosing ``None of the above,'' so when given the same task without the ``None of the above'' option, the score rose to an average of 7/10.
    \end{itemize}
    \item \textbf{Generating Design Descriptions~\ref{cd:generating_desc}} 
    Is GPT-4V capable of generating effective descriptions for early-stage design sketches?
    \begin{itemize}
        \item Qualitatively, we find that GPT-4V is able to generate accurate and useful design descriptions given hand-drawn sketches.
    \end{itemize}
    \item \textbf{Generating Selection Criteria~\ref{cd:selection_criteria}} 
    How effectively does GPT-4V generate concept selection criteria in engineering design?
    \begin{itemize}
        \item In our case study, we find that when provided a design task GPT-4V generates useful selection criteria that match those generated by design professionals.
    \end{itemize}
    \item \textbf{Generating a Pugh Chart~\ref{cd:pugh_chart}} 
    What is the extent and limitation of GPT-4V's ability to generate Pugh charts for conceptual design evaluation?
    \begin{itemize}
        \item GPT-4V understands what a Pugh chart is, and can provide examples of the formatting, but often will not fill in the Pugh chart, or simply provide a ``looks-like'' Pugh chart given just a design task and design sketches. The model cites that it cannot fill in the Pugh chart without additional context about the designs, suggesting it may be able to if provided with more information.
    \end{itemize}
\end{enumerate}

\section{System-Level And Detail Design}\label{sec:detailed}
\paragraph{Overview and Motivation}
After conceptual design, engineers move into the system-level and detail design portion of the product development process~\citep{ulrich_product_2020}. During this phase, engineers flesh out the specifics of the design as they select materials, develop CAD models, iterate to optimize the design, abide by constraints, and create engineering drawings for subsequent manufacturing steps. For example, an experienced mechanical engineer, embarking on the design of a new lightweight bicycle frame, would necessarily go through a material selection phase -- to identify materials that are both strong and light -- a CAD generation phase -- to develop the geometric details of the design -- and a design optimization phase -- to refine the design to best meet a set of objectives. These steps are very common and critical to the detail design stage and inform the structure of this section.

The generation of system-level and detail designs draws upon many skill sets, including spatial reasoning, knowledge of specific CAD software programs, and physics-based principles. Many of the tasks that engineers perform during this step of the product development process are inherently visual. They must consult material charts and graphs, read and create engineering drawings, develop 3D geometric models in CAD GUIs, and visually check for geometric constraint violations. Given VLMs' image-processing capabilities and emergent technical knowledge, we test, in the follow sections, the models' abilities to support visual-based detail design tasks.

\subsection{Material Selection}
\label{sec:material_selection}
Oftentimes, the selection of material comes early-on in the detail design phase, as material choice informs both the design and the utilized manufacturing method. Material selection requires balancing various constraints and requirements, such as material strength, stiffness, cost, density, embodied energy, electrical resistivity, and thermal conductivity~\citep{ashby2016materials}. Choosing a material that meets an extensive list of requirements and constraints often requires cross-referencing multiple tables or charts, such as Ashby charts~\citep{ashby2016materials}. Ashby charts enable engineers to visually represent the trade-offs between various material properties for different families of materials. Provided with these charts for different material properties, LLMs have the potential to condense material information and identify materials that meet certain criteria. Several groups have explored GPT's ability to assist with material considerations. In \citep{saka2023gpt}, the authors used the GPT API to integrate ChatGPT into the building information modeling process to assist with material selection for components of a building. In \citep{makatura_how_2023}, the authors looked at GPT-4's ability to propose a manufacturing process for a part based on material selected. In \citep{buehler2023melm}, the authors trained a model, MeLM (Mechanics Language Model), which was used for material-related tasks, like proposing microstructure designs that meet certain stress-strain responses. In this section, we conduct three independent experiments involving Ashby charts and material selection. To analyze the consistency of responses, each experiment is repeated three times.

\begin{figure}
\begin{contextbox}[]{Ashby Chart Look Up}
\begin{promptbox}{(Identifying materials that meet stiffness and density constraints.)}
This is an Ashby chart used for the selection of materials in engineering design. I am looking for a material that has a density from 7-10 $\frac{Mg}{m^3}$ and a Young's Modulus greater than 100 GPa. Based on the chart, which materials meet these constraints? \\
\begin{center}
\includegraphics[height=3.3cm, keepaspectratio]{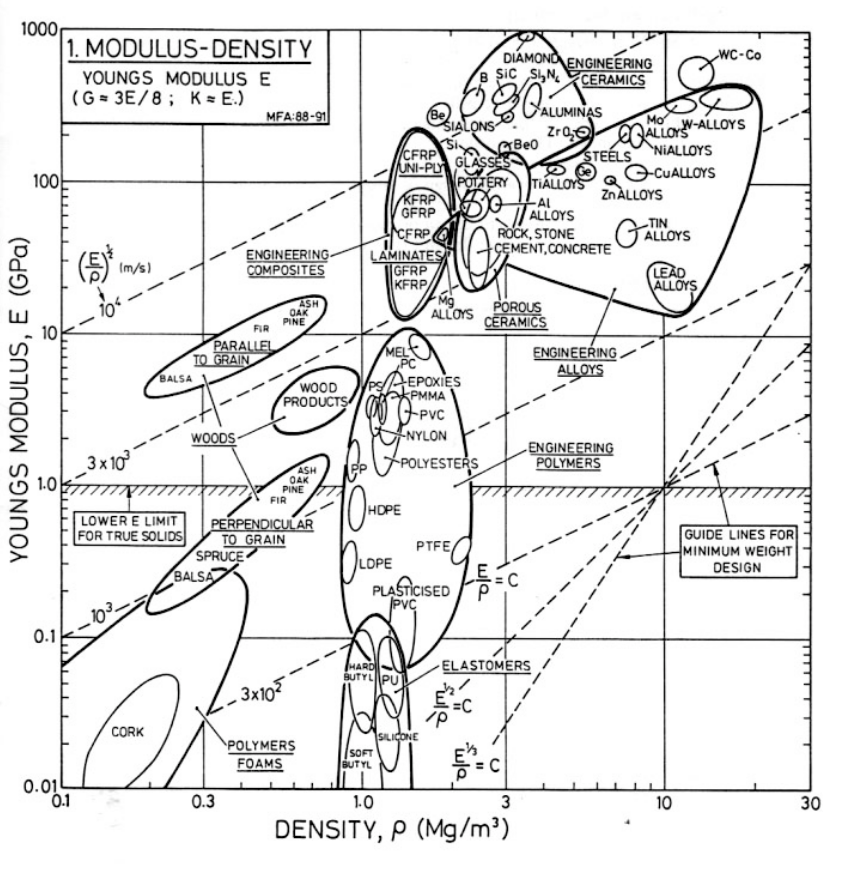}     
\end{center}
{\tiny
\textit{Image from \citep{mit_ocw_unified_engg}} used under CC BY-NC-SA 4.0.}  
\tcblower

Based on the Ashby chart you've provided, the materials that meet the constraints of having a density between 7--10 Mg/$\text{m}^3$ and a Young's Modulus greater than 100 GPa are located in the \hl{top right} quadrant of the graph. Here's how to interpret the chart for your specified ranges:

\begin{enumerate}[leftmargin=*,labelindent=0mm,]
    \item \hl{Identify the horizontal band corresponding to a Young's Modulus (E) greater than 100 GPa. This is above the horizontal line marked with 100 GPa on the left-hand side vertical axis.}

    \item \hl{Locate the vertical band corresponding to a density ($\rho$) between 7-10 Mg/$\text{m}^3$. This band lies between the vertical lines marked 7 and 10 on the horizontal axis at the bottom.}
\end{enumerate}

\hl{The materials that fall within the overlap of these two bands are your candidates.} From the chart, it looks like the following materials fall within this range:

\begin{itemize}[leftmargin=*,labelindent=0mm,itemsep=1pt]
    \item \hl{Steels (various types)}
    \item \hle{Titanium alloys (Ti alloys)}
    \item \hl{Nickel alloys (Ni alloys)}
    \item \hle{Tungsten carbide-cobalt composites (WC-Co)}
\end{itemize}

These materials are marked within ovals or circles and fall within the intersection of the specified Young's Modulus and density ranges. It is important to note that the exact positioning within the oval represents variations within the material category, so specific grades or compositions of steel, titanium alloys, nickel alloys, and WC-Co may need to be looked up to ensure they meet the exact requirements of your application.

\end{promptbox}
\end{contextbox}
\caption{Ashby chart look-up: Identifying materials that meet stiffness and density constraints.}
\label{ashby-1}
\end{figure}

\begin{figure}    
\begin{contextbox}[]{Ashby Chart Cross-Referencing}
\begin{promptbox}{(Identifying materials that meet stiffness, strength, and density constraints.)}
These are two Ashby charts that are used for the selection of materials in engineering design. I am looking for a material that has a density from 1.0 to 3.0 Mg/$\text{m}^3$, a Young's Modulus between 0.01 and 0.1, and a strength of 3 MPa. Based on the provided charts, which materials meet these constraints? \\
\begin{center}
\includegraphics[height=3.3cm, keepaspectratio]{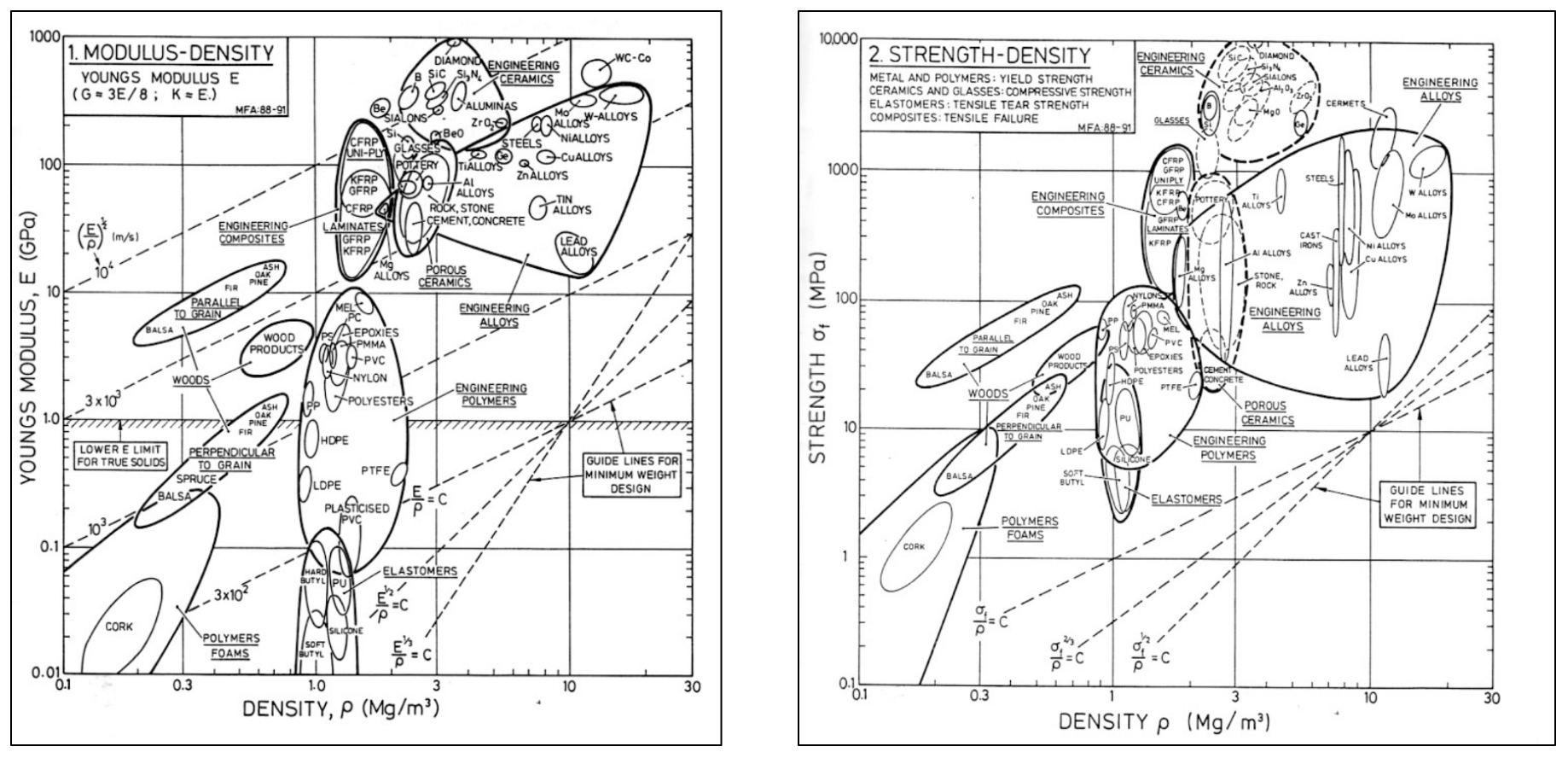}     
\end{center}

{\tiny
\textit{Images from \citep{mit_ocw_unified_engg} and \citep{mit_ocw_unified_engg_zm21}} used under CC BY-NC-SA 4.0.} 
\tcblower

Based on the Ashby charts provided:

For the modulus-density chart on the left, \hl{materials that fall within a density of 1.0 to 3.0 Mg/$\text{m}^3$ and a Young's Modulus between 0.01 and 0.1 GPa are located in the bottom center portion of the graph}. \hle{From the chart, it seems that 'Polymers Foams' and 'Cork' could potentially meet these criteria}.

\hl{For the strength-density chart on the right, you are looking for materials with a density from 1.0 to 3.0 Mg/$\text{m}^3$ and a strength of 3 MPa. The materials that meet these criteria would be plotted around the 3 MPa line, which is not explicitly drawn on this graph.} However, materials just above the 'Polymers Foams' region could potentially meet the strength requirement of 3 MPa.

It's important to note that the exact overlap between the modulus and strength requirements cannot be determined precisely from these charts without more detailed scales, but \hle{'Polymers Foams' appear in both regions of interest and might be a good starting point for further investigation}. It's always recommended to check the detailed properties of these materials from a materials database or literature to ensure they meet the specific engineering requirements.

\end{promptbox}
\end{contextbox}
\caption{Ashby chart cross-referencing: Identifying materials that meet stiffness, strength, and density constraints.}
\label{ashby-2}
\end{figure}

\begin{figure}
\begin{contextbox}[]{Material Selection for a Beam}
\begin{promptbox}{(Selecting materials based on design requirements.)}
I am looking to select a material for a beam I am designing. I want the beam to be both stiff and light. Attached is an Ashby chart, which can be used for the selection of materials in engineering design. Based on this chart and the information provided, can you suggest materials that could be used for the beam? \\
\begin{center}
\includegraphics[height=3.3cm, keepaspectratio]{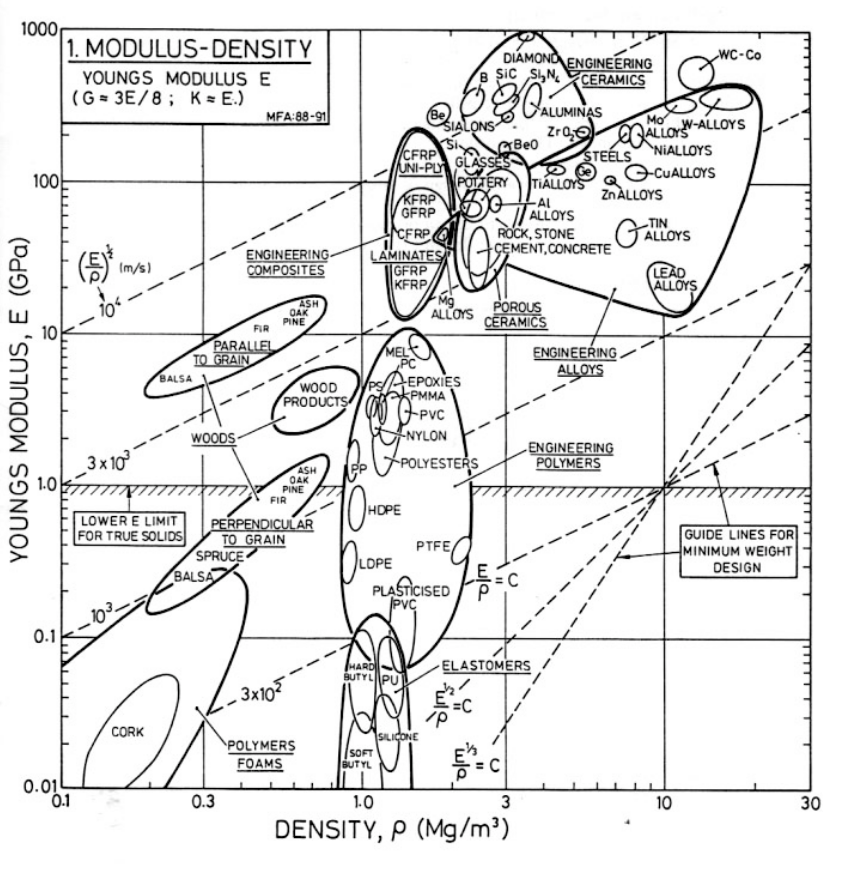}     
\end{center}
{\tiny
\textit{Image from \citep{mit_ocw_unified_engg}} used under CC BY-NC-SA 4.0.}
\tcblower

\hl{The Ashby chart you've provided is used to select materials based on their stiffness (Young's modulus) and density. For a beam that is both stiff and light, you would want materials that have a high Young's modulus to density ratio. These materials would appear towards the top left of the chart, as this region represents materials with high stiffness and low density.}

On the chart, materials are clustered into groups. The \hl{``Engineering Composites'' group, which includes materials like CFRP (Carbon Fiber Reinforced Plastic) and GFRP (Glass Fiber Reinforced Plastic)}, stands out as particularly suitable for applications requiring high stiffness and low weight. Composites are known for their excellent strength-to-weight ratios and are commonly used in aerospace and automotive applications for this reason.

Another group that might be relevant is the \hl{``Engineering Alloys,''} especially those towards the left of the group indicating lower density. This could include certain aluminum alloys, which are known for their use in lightweight structures where stiffness is also a requirement.

To narrow down the selection further, you would consider additional factors such as cost, manufacturing processes, environmental resistance, and compatibility with other materials used in the design. It would also be important to ensure the selected material can withstand the specific loading conditions of your beam design.

\end{promptbox}
\end{contextbox}
\caption{Application of Ashby charts to the material selection for a beam}
\label{ashby-3}
\end{figure}

\paragraph{Methodology}
For the material selection experiments, we provide GPT-4V with various Ashby charts, which are commonly used by engineers to evaluate trade-offs between different materials. We conduct three different types of experiments: Ashby chart look-up, Ashby chart cross-referencing, and Ashby chart material selection for a beam. Each experiment type is repeated three times. One of the repetitions of each of the three experiment types can be seen in Fig.~\ref{ashby-1} (look-up), Fig.~\ref{ashby-2} (cross-referencing), and Fig.~\ref{ashby-3} (material selection for a beam).

For the look-up experiment, we provide the model a density vs. Young's modulus Ashby chart. We ask the model to identify materials that have a density between 7-10 $\frac{Mg}{m^3}$ and a Young's Modulus greater than 100 GPa. The purpose of this experiment is to assess whether GPT-4V can perform a simple ``look-up'' of feasible materials from the chart. 

For the cross-referencing experiment, we give GPT-4V two Ashby charts, one showing density vs. Young's modulus and another showing density vs. strength. We then ask GPT-4V to cross-reference the two charts, identifying materials that have a density in between 1-3 $\frac{Mg}{m^3}$, a Young's modulus between 0.01 and 0.1 GPa, and a strength of 3 MPa. The purpose of this experiment is to understand if GPT-4V can synthesize information from two material charts together. 

For the material selection for a beam experiment, we provide the model a density vs. Young's modulus Ashby chart. We ask GPT-4V to help us select a material for a hypothetical beam, given general requirements that the beam must be both stiff and light. The purpose of this experiment is to understand if GPT-4V can translate the general requirements into material requirements and propose appropriate material families based on those requirements.

\begin{table*}[h]
    \caption{Summarized results for the GPT-4V material selection experiments}
    \label{tab:full_material_results}
    \footnotesize
    \begin{tabular}{@{}l|ccc@{}} \toprule
         \textbf{Repetition \#} &  \makecell{\textbf{Ashby Chart} \\ \textbf{Look-Up}} & \makecell{\textbf{Ashby Chart} \\ \textbf{Cross-Referencing}} & \textbf{Material Selection for a Part} \\ 
         \midrule
         Rep 1 & \makecell{\hl{Steels} \\ \hl{Ni-alloys} \\ \hle{Ti-alloys}\\ \hle{WC-Co}} & \makecell{\hle{Polymer Foams}} & \makecell{\hl{Top left of chart} \\ \hl{Engineering Composites} \\ \hl{Engineering Alloys}} \\
         \midrule
         Rep 2 & \makecell{\hl{Steels} \\ \hle{Ti-Alloys}} & \makecell{\hle{*Polymers} \\ \hle{Foams} \\ \hle{Woods w/ Lower Densities}} & \makecell{\hl{Top left of chart} \\ \hl{Engineering Composites} \\ \hl{Engineering Ceramics} \\ \hl{Engineering Alloys} \\ \hl{Wood Products}}  \\
         \midrule
         Rep 3 & \makecell{\hl{*Steels} \\ \hl{Ni-Alloys} \\ \hle{WC-Co}} & \makecell{\hle{*Polymers} \\ \hle{Woods} \\ \hle{Polymer Foams}} & \makecell{\hl{Top left of chart} \\ \hl{Engineering Composites} \\ \hl{Engineering Ceramics} \\ \hl{Engineering Alloys} \\ \hl{Wood Products}}  \\
         \bottomrule
         \multicolumn{4}{@{}p{\textwidth}@{}}{\footnotesize *For these responses, GPT-4V expressed concern about making definitive material choices based on the resolution of the provided Ashby charts.} \\
    \end{tabular}
\end{table*}

\paragraph{Discussion: Material Selection} Overall, we conclude that GPT-4V performs well at identifying broad material families that exhibit general properties (e.g. low density), but performs less well when given specific requirements or constraints (e.g., density between 1.0 and 3.0). This finding is illustrated by the results of the three experiments. The responses from all experiments and repetitions can be seen in Table~\ref{tab:full_material_results}. One repetition of each experiment is displayed in-full in Fig.~\ref{ashby-1}, \ref{ashby-2}, and \ref{ashby-3}. For the Ashby Chart Look-Up experiment (see Fig.~\ref{ashby-1}), we would expect the correct answer---materials that have a density between 7 and 10 $\frac{Mg}{m^3}$ and that have a Young's modulus greater than 100 GPa---to be steel, Ni-alloys, and Cu-alloys. (Zn-alloys and Mo-alloys lie on the border of the feasible region). For all three repetitions, GPT-4V correctly answered that steels would be feasible materials. For two out of the three times, it also mentioned that Ni-alloys would meet the specified requirements. However, in all three iterations it also included materials in its answer---either Ti-alloys or WC-Co---that do not meet our specifications;  Ti-alloys have a density less than 7 $\frac{Mg}{m^3}$ and WC-Co materials have a density greater than 10 $\frac{Mg}{m^3}$. In none of the iterations did GPT-4V mention Cu-alloys in its answer, although this material group meets both the stiffness and density specifications. 

GPT-4V performed poorly in our Ashby Chart Cross-Referencing Experiment (see Fig.~\ref{ashby-2} for the full response from one of the repetitions). The correct answers to the question were the soft butyl and elastomer materials, which have densities between 1.0 to 3.0 $\frac{Mg}{m^3}$, Young's moduli between 0.01 and 0.1, and a strength of 3 MPa. Across the three repetitions, GPT-4V never identified either of these materials as meeting our requirements. Overall, it tends to conclude that polymer foams, foams, polymers, or woods would be suitable choices, but these materials do not meet our specifications. Polymer foams, for example, don't meet the density requirement; many polymer foams have densities between 0.1 and 0.3 $\frac{Mg}{m^3}$, suggesting that GPT-4V possibly confuses our 1.0 - 3.0 $\frac{Mg}{m^3}$ density specification with this 0.1 to 0.3 range. It is important to note that for two of the three repetitions of this experiment, GPT-4V was hesitant to provide an answer due to the ``resolution" of the provided images.

Our Ashby Chart Look-Up and Cross-Referencing experiment reveals possible areas of improvement needed in handling precise numerical data and complex information synthesis. The model's struggle with accurately interpreting numerical constraints, as evidenced by these two experiments, highlights a shortfall in applying exact numerical ranges. Furthermore, its inability to effectively cross-reference and synthesize data from multiple sources underscores a challenge in processing multi-dimensional information. This issue is particularly pertinent in engineering where precision and multi-faceted data analysis are crucial.

GPT-4V performs much better when asked to propose potential material families for a beam that needs to be both lightweight and stiff (see Fig.~\ref{ashby-3}). For all three experiment repetitions, GPT-4V correctly translates the stiffness specification into a high Young's modulus requirement and the light-weight specification into a low-density requirement. For all trials, it correctly asserts that materials we would want to consider are towards the top left of the provided Ashby chart and proposes engineering composites and engineering alloys (and for two of the three repetitions, also proposes engineering ceramics and wood products).  We conclude that while GPT-4V struggles to identify materials that meet specific numerical requirements, it is much better at proposing material families that meet general specifications. 

Overall, the use of GPT-4V in material selection in engineering design showcases its potential as a supportive tool in the preliminary stages of decision-making and as an educational aid in materials science. Its ability to suggest material families based on general requirements can streamline the initial phases of design, allowing engineers to focus on finer details. This integration points towards a future where AI complements traditional engineering tools, enhancing the efficiency of design workflows. However, it also raises important ethical and practical considerations, such as over-reliance on VLM models without knowing their limitations and ensuring AI-generated recommendations of material align with safety standards and environmental concerns. For example, the varying answers across the repeats highlight the non-deterministic nature of ChatGPT's interface, which engineers need to consider.

Transitioning from the exploration of GPT-4V's capabilities in material selection, the research now shifts focus to another critical aspect of engineering design: VLM's ability to interpret complex engineering drawings and contribute to the generation of Computer-Aided Design (CAD) models.

\subsection{Engineering Drawing Analysis and CAD Generation}
\label{sec:cad_and_engineering_drawings}

A critical step of the detailed design process is the generation of 3D models. Computer-aided design (CAD) software enables the creation of detailed solid models, allowing engineers to encode precise part dimensions and assembly relations between parts \citep{nyemba2022computer}. These CAD models pave the way for design for manufacturing, since detailed engineering drawings with manufacturing specifications are typically created from the 3D models \citep{nyemba2022computer}. CAD models are also useful for the different ways in which designs and parts can be visualized (e.g. cross-sections, wireframe views, etc.), enabling engineers to easily consider different aspects of their design \citep{nyemba2022computer}. We hypothesized that GPT-4 with vision would be better able to assist with CAD generation and engineering drawing analysis than GPT-4 since these two design forms -- CAD models and engineering drawings -- are inherently visual mediums. 

We gain inspiration from the work of researchers who have explored the potential of GPT to assist with converting text into CAD \citep{makatura_how_2023, nelson2023utilizing}. For example, Makatura et al. devoted a large section of their work to the exploration of GPT-4's ability to generate CAD designs from text. They looked at text to scripts that would generate 2D designs (DXFs and SVGs), demonstrating relative success in the design of 2D pieces of a cabinet. Another work then performed several case studies to illustrate GPT-4's ability to convert text into scripts for 3D designs, using both CSG-based CAD languages and sketch-based CAD languages~\citep{makatura_how_2023}. These experiments showcased mixed success, oftentimes requiring prompts to be engineered with specific function signatures. The authors noted reasoning challenges, specifically when it came to spatial reasoning. They also cited iterative ability as both a capability and a drawback of GPT-4, as they found the model was sometimes able to correct errors through continued chat iterations but that more iterations also led to limited memory of previous information in a chat.
However, a key limitation of past work is that it relied on text-only LLMs, while CAD is inherently a task that has significant visual aspects. In our study, we focus on evaluating the capabilities of VLMs.

\paragraph{Methodology}

\begin{figure*}
\begin{contextbox}[label=cad_and_drawings]{Experiment 1---Engineering Drawing Analysis and CAD Generation Experiment}
\begin{minipage}{0.498\linewidth}\vspace{0pt}%
\begin{promptbox}[]{(P1 - Part Description from an Engineering Drawing)}
 This is an engineering drawing of a part, showing four views of the part. Can you tell me what the part is?
    \begin{center}
    \includegraphics[width=0.57\linewidth,keepaspectratio]{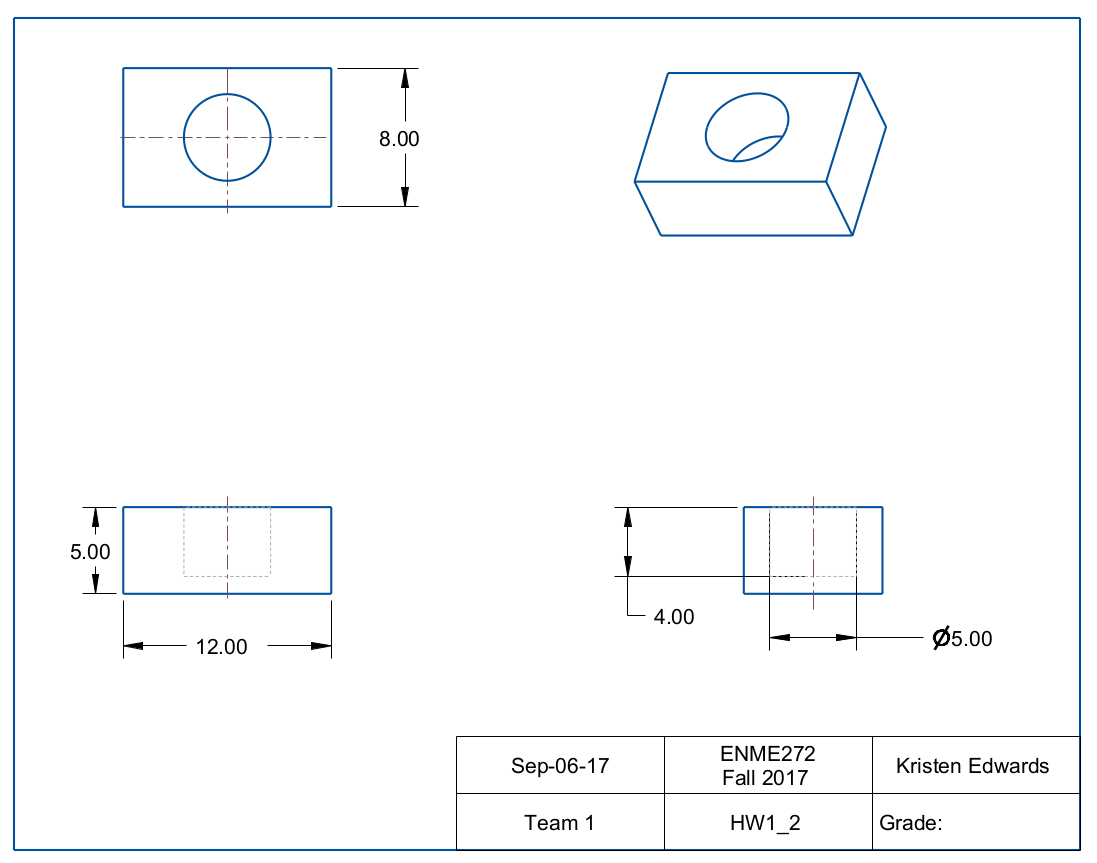}    
    \end{center}
\tcblower

Certainly! Based on the provided engineering drawing, the part depicted is a \hl{rectangular block (or prism) with a cylindrical} \hle{through-hole}. Here's a breakdown:

\begin{enumerate}[leftmargin=*,labelindent=0mm]
    \item The top-left view shows a top view of the part. \hl{We can see a circle, which indicates a cylindrical hole} \hle{through the part}. The dimensions indicate that the length of the rectangular block is 12.00 units and the width is 5.00 units.

    \item The bottom-left view is a front view, where the height of the block is shown to be 8.00 units.

    \item The bottom-right view is a right-side view, showing the height of the block as 4.00 units and the diameter of the cylindrical hole as 5.00 units.

    \item The top-right view is an isometric or 3D view that gives a perspective of how the part looks in 3D.
\end{enumerate}

So, to summarize, it's a \hl{rectangular block with dimensions 12.00 units x 5.00 units x 8.00 units with a cylindrical hole} of diameter 5.00 units \hle{running through} its shorter width (5.00 units).

\end{promptbox}
\begin{promptbox}[]{(P2 - Dimension Extraction from an Engineering Drawing)}
 The part depicted in the engineering drawing is a block with a blind hole. Based on the engineering drawing views, can you please extract all dimensions of the part that are shown on the drawing, assigning each dimension an appropriate name? Only list dimensions that appear in the drawing; do not extrapolate from them to create more dimensions.

\tcblower

Of course! Based on the provided engineering drawing, here are the dimensions of the part:

\begin{enumerate}[leftmargin=*,labelindent=0mm]
    \item \hl{Length of the Block: 12.00 units} (Shown on the bottom-left view)
    \item \hl{Width of the Block: 5.00 units} (Both top-left and bottom-left views indicate this width)
    \item \hl{Height of the Block (from top-left view): 8.00 units}
    \item \hle{Height of the Block (from bottom-right view): 4.00 units} (This is likely the depth of the blind hole, as it doesn't span the entire height)
    \item \hl{Diameter of the Blind Hole: Ø5.00 units} (Shown on the bottom-right view)
\end{enumerate}

From the information provided, the hole is a blind hole because it doesn't go all the way through the block, as indicated by the different heights in the top-left and bottom-right views.
\end{promptbox}
\begin{promptbox}{(P3 - CAD Generation 1)}
You are correct that the height of the block is 8.00 units, but it is not correct to say that the height of the block is also 4.00 units. The 4.00 unit dimension represents the depth of the blind hole. Now that you have extracted the dimensions and given that you still have the engineering drawing, please use the CAD scripting language, CadQuery, to generate a script that encodes the CAD of the part.

\tcblower

Of course, I apologize for the oversight. Let's proceed with creating a CadQuery script for the part based on the provided engineering drawing.

Here's a CadQuery script to generate the CAD model of the part:

\begin{lstlisting}[basicstyle=\tiny]
(*@\hl{import cadquery as cq} @*)
# Parameters from the drawing
(*@\hl{length = 12.00} @*)
(*@\hl{width = 5.00} @*)
(*@\hl{height = 8.00} @*)
(*@\hl{hole\_diameter = 5.00} @*)
(*@\hl{hole\_depth = 4.00} @*)
# Create a block with the specified dimensions
(*@\hl{block = cq.Workplane("XY").box(length, width, height)} @*)
# Create a blind hole in the block
(*@\hle{block\_with\_hole = block.faces(">Z").workplane().circle(hole\_diameter / 2).cutBlind(-hole\_depth)} @*)
[...]
\end{lstlisting}

You would need to have CadQuery installed and integrated with your CAD system to run the script. [...]
When you run this script with CadQuery, it will generate a block with the specified dimensions and a blind hole according to the drawing.

\end{promptbox}
\end{minipage}\hfill%
\begin{minipage}{0.498\linewidth}\vspace{0pt}%
\begin{promptbox}[]{(P4 - CAD Generation 2)}

That's not quite right – the CAD of the part created using the OpenSCAD script you just gave me does not match the part shown in the original engineering drawing I provided you. The attached photo shows four views of the part's CAD that was created using the CadQuery script you just provided. Each view shows a different orientation of the CADed part, as seen by the coordinate system shown in each view. The views show the CADed part in a shaded-with-hidden-edges style. Based on these CAD views, can you update the CadQuery script to fix any issues that you see, so that the part created using the script matches the part shown in the original drawing I provided?

 \begin{center}
    \includegraphics[width=0.62\linewidth]{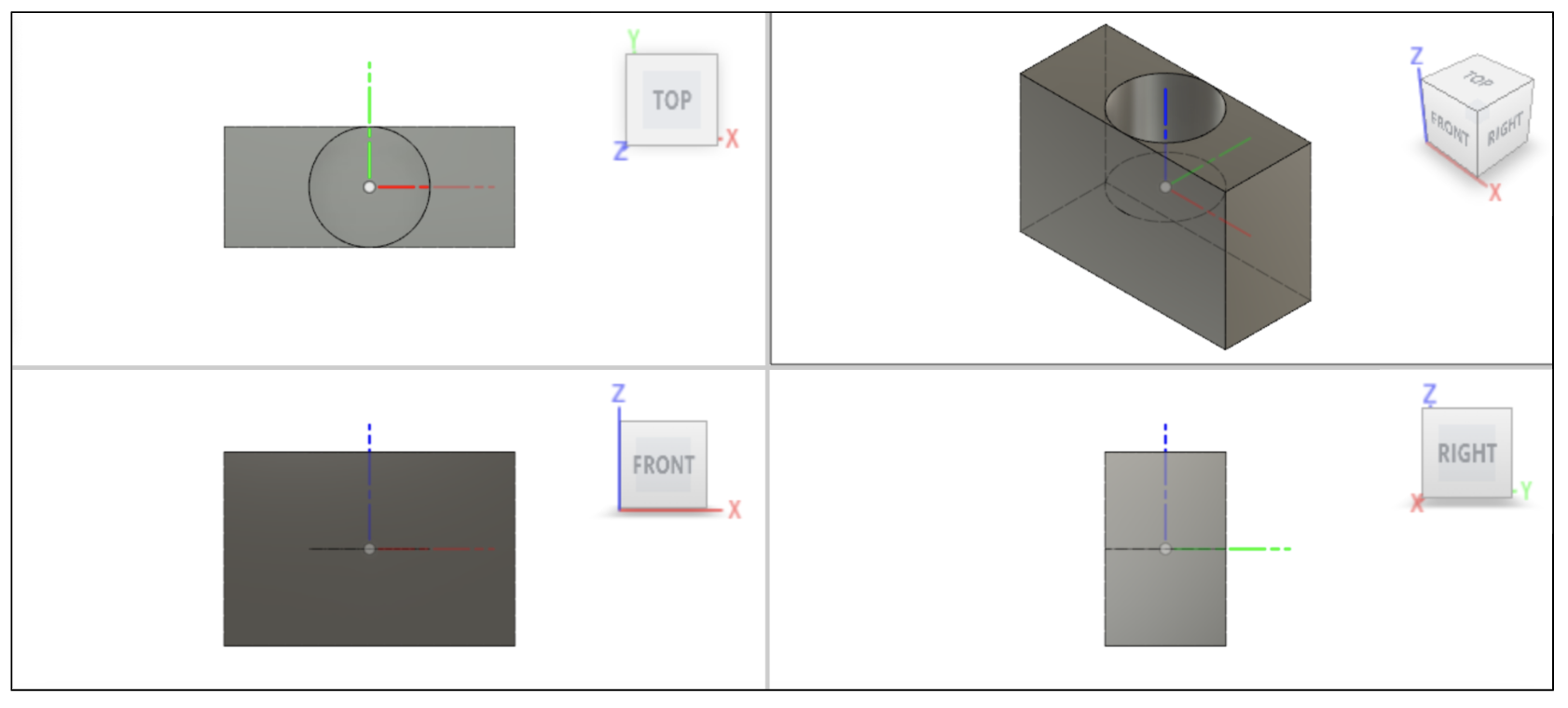}    
    \end{center}
\tcblower

Certainly, I apologize for the oversight. Let's correct the CadQuery script.

From the CAD views you provided and comparing it with the original engineering drawing, I observe the following:

\begin{enumerate}[leftmargin=*,labelindent=0mm]
    \item \hle{The hole is a through-hole and not a blind hole. This means the depth of the hole is equivalent to the height of the block.}
    \item \hle{The hole is off-centered in both X and Y directions.}
\end{enumerate}

Let's create a new CadQuery script with these corrections:

\begin{lstlisting}[basicstyle=\tiny]
import cadquery as cq
# Parameters from the drawing
(*@ \hl{length = 12.00} @*)
(*@ \hl{width = 5.00} @*)
(*@ \hl{height = 8.00} @*)
(*@ \hl{hole\_diameter = 5.00} @*)
(*@ \hle{hole\_depth = height} @*)  # making it a through-hole
(*@ \hle{\# Assuming half the width of the block minus half the hole diameter for the Y offset (based on drawing) \\
offset\_y = (width - hole\_diameter) / 2} @*)
(*@ \hle{\# Assuming half the length of the block minus half the hole diameter for the X offset (based on drawing) \\
offset\_x = (length - hole\_diameter) / 2} @*)
# Create a block with the specified dimensions
(*@ \hl{block = cq.Workplane("XY").box(length, width, height)} @*)
(*@ \hle{\# Create a through-hole in the block at the specified offsets \\
block\_with\_hole = block.faces(">Z").workplane().center(offset\_x, offset\_y).circle(hole\_diameter / 2).cutThruAll()} @*)
[...]
\end{lstlisting}

\end{promptbox}%
\begin{promptbox}[]{(P5 - CAD Generation 3)}
That's not quite right – the CAD of the part created using the OpenSCAD script you just gave me does not match the part shown in the original engineering drawing I provided you. The attached photo shows four views of the part's CAD that was created using the CadQuery script you just provided. Each view shows a different orientation of the CADed part, as seen by the coordinate system shown in each view. The views show the CADed part in a shaded-with-hidden-edges style. Based on these CAD views, can you update the CadQuery script to fix any issues that you see, so that the part created using the script matches the part shown in the original drawing I provided?

 \begin{center}
    \includegraphics[width=0.62\linewidth]{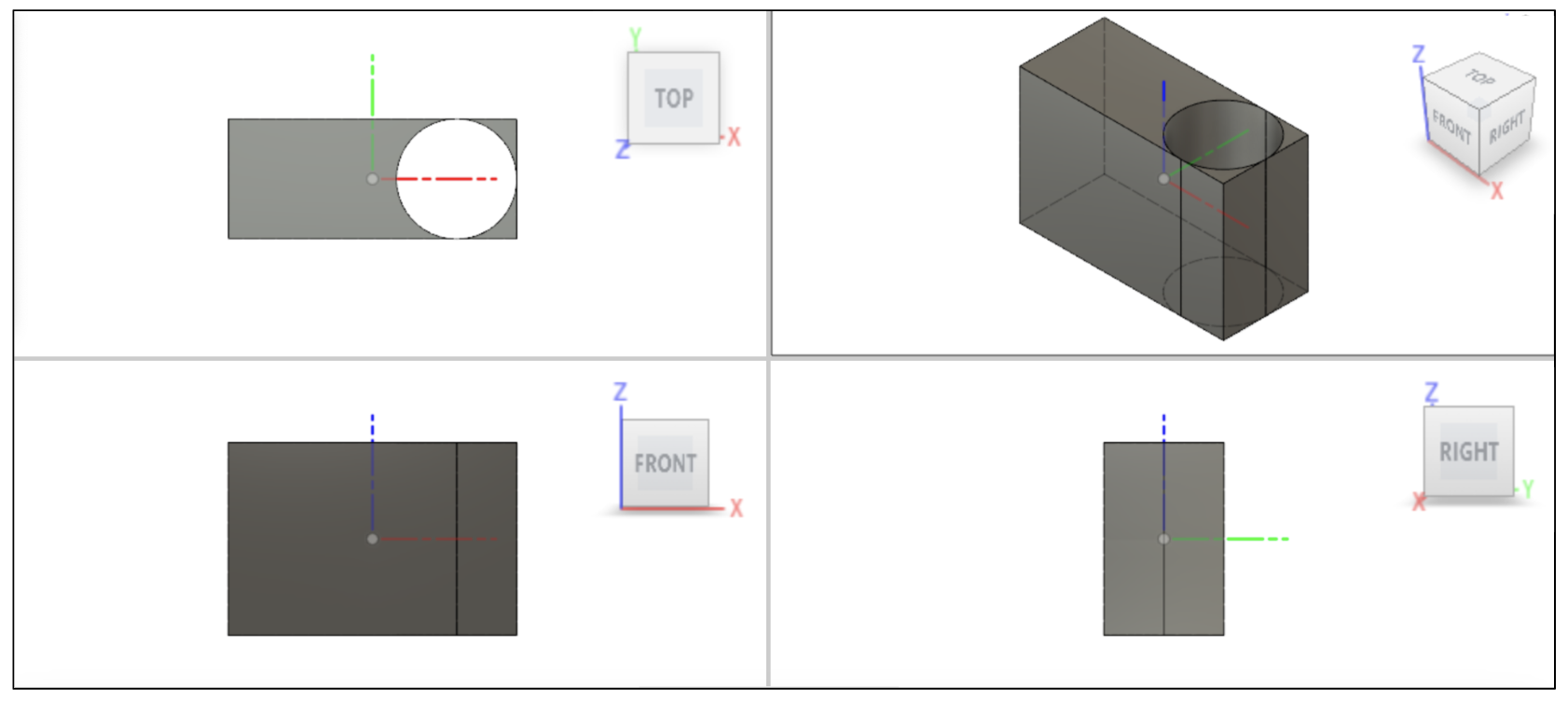}    
    \end{center}
\tcblower

Certainly! Based on the provided CAD views, let's identify and rectify the discrepancies between the CAD views and the original engineering drawing:

\begin{enumerate}[leftmargin=*]
    \item Top View: The CADed part has a cylindrical cutout at the left side, but it's not completely cutting through. \hle{The original engineering drawing indicated that this cylinder should be a through hole.}
    \item Front View: \hle{The rectangular cutout looks fine in relation to the original drawing. However, we can't confirm the depth of this cutout from this view alone.}
    \item Right View: The circular cutout seems to be centered vertically, which seems accurate as per the original drawing.
\end{enumerate}

Given these observations, let's make corrections to the CadQuery script:

\textit{[Incorrect code based on the above observations]}
\end{promptbox}%
\end{minipage}

\end{contextbox}
\caption{Overview of the prompts and answers for one experiment of engineering drawing analysis and CAD generation using OpenSCAD. Only the three first CAD generation iterations are shown.}
\label{cad-generation}
\end{figure*}

To assess GPT-4V’s ability to analyze engineering drawings and generate CAD, we utilized a seven-prompt experiment structure (carried out in the same context window). An example of a full experiment can be seen in Fig.~\ref{cad-generation}. The first two prompts (P1 and P2) of each experiment assess GPT-4V’s ability to analyze engineering drawings. We test the model on two aspects: 1) its ability to describe a part based on an engineering drawing and 2) its ability to extract dimensions from an engineering drawing. We correct any responses to these questions that the model answers incorrectly, so we can independently score the next part of the experiment. In prompts three through seven (P3 - P7), we evaluate the model's ability to generate a script that encodes the CAD of a part. We ask the model to do this based on the previously provided engineering drawing, the previously extracted part dimensions, and a CAD scripting language that we specify. In this part of the experiment, we score the model on the CAD that its script generates. If the CAD is not correct on the first attempt, we feed it back views of the generated CAD and ask it to iterate to fix any discrepancies it sees between the generated CAD and the original engineering drawing. In this way, our feedback is much like the model's built-in Code Interpreter, but instead for CAD. We repeat this iterative process until five different CAD generation attempts have been made. 

In total, we ran nine experiments, each with seven prompts (P1-P7) conducted sequentially in a single chat context. Three groups of three experiments used identical prompts (conducted for repetition), and the difference between the three groups of experiments is in the CAD scripting language specified. We now further elaborate on the experiment structure and the method for scoring each experiment:

The prompts for these queries are identical across all nine experiments.

\begin{itemize}
    \item \textbf{Part Description from an Engineering Drawing} - Prompt 1 (P1). GPT-4V is given an engineering drawing of a block with a blind hole, as seen in Fig.~\ref{cad-generation}, P1. We ask the model to provide a description of the part. We chose to use this block-with-blind-hole part as the subject of our experiments since it represents one of the most basic yet functional parts that can be created using CAD, necessitating only two sketches and basic cut/extrude operations. The drawing follows typical engineering drawing conventions and was created in an undergraduate-level engineering course.\footnote{ The engineering drawing of the block with blind hole shown in Fig.~\ref{cad_and_drawings} was created as an assignment for the ``ENME272: Introduction to Computer Aided Design" course at the University of Maryland.} As such, it is a drawing that we would expect undergraduate-level mechanical engineers to readily understand.
    
    \begin{itemize}
        \item Scoring (1 point possible): We assign 1 point if GPT-4V correctly mentions that the part is a ``block with a hole'' or ``block with a blind hole.'' Any mention of a ``through'' hole receives no points as it shows an incorrect understanding of the underlying geometry.
    \end{itemize}

    \item \textbf{Dimension Extraction from an Engineering Drawing} - Prompt 2 (P2). We next ask GPT-4V to extract the dimensions shown in the engineering drawing, assigning them appropriate names. We specifically ask GPT-4V to not extrapolate to create dimensions that are not explicitly called out in the drawing.

    \begin{itemize}
        \item Scoring (10 points possible): 1 point is awarded for each of the five numbers shown on the drawing (8.00, 5.00, 12.00, 4.00, and \diameter5.00) that GPT-4V successfully extracts. Another point is awarded for each of the five dimensions to which it assigns an appropriate name. For the block dimensions — 8.00, 5.00, and 12.00 — we accept any assignment of [length, width, height] or [depth, width, height] to the three dimensions since the assignment of these labels depends on how the block is oriented. For the 4.00 dimension and the \diameter5.00 dimension, we expect labels of ``hole depth'' and ``hole diameter'' respectively, or equivalent names. For any dimensions GPT-4V lists beyond those shown in the drawing, we subtract 1 point for not following instructions.
    \end{itemize}

\end{itemize}

The prompts for the following queries vary across the nine experiments, as indicated below.

\begin{itemize}
    \item \textbf{CAD Generation 1} —  Prompt 3 (P3). Continuing in the same context window where P2 left off, we correct any dimensions that GPT-4V extracted from the drawing incorrectly, and we then ask GPT-4V to generate a CAD script of the block-with-hole part based on the engineering drawing provided in P1 and the dimensions it extracted in P2. For three of the experiments (experiments 1 – 3), we ask GPT-4V to do this using the CadQuery scripting language; for another three experiments (experiments 4 – 6), we ask GPT-4V to do this with a different scripting language, FeatureScript; and for the last three experiments (experiments 7 – 9), we ask GPT-4V to use the CAD scripting language OpenSCAD. Note that each language offers unique features and advantages:

    \begin{itemize}
        \item CadQuery: An open-source CAD scripting module built-in Python, CadQuery is easy to use for those already familiar with Python.
        \item FeatureScript: The scripting language of the CAD software Onshape -- a free cloud-based CAD software -- FeatureScript is integrated into Onshape, enabling both traditional CAD modeling and custom, script-defined, parametric modeling.
        \item OpenSCAD: Another open-source CAD scripting language built in C++, OpenSCAD is integrated into the CAD software FreeCAD and provides granular control over a model.
    \end{itemize}

By utilizing these three scripting languages, our research aims to comprehensively assess GPT-4V's ability to adapt to different CAD scripting environments and to evaluate its versatility in translating engineering drawings into functional CAD models.

    \begin{itemize}
        \item Scoring (6 points possible): We assign 1 point if the generated script has no errors when run. We award 1 point for each of the following features that the generated CAD possesses: the block has correct dimensions; the CAD has a hole on the largest block face; the hole is centered on a face; the hole has the correct depth; and the hole has the correct diameter. We subtract a point for each extra, incorrect feature that is present in the generated CAD (e.g. a second hole, a cutout in the block, etc.).
    \end{itemize}

    \item \textbf{CAD Generation 2 - 5}:  Prompts 4 - 7 (P4 - P7). If the CAD generated by the previous prompt has a syntax error when the code is run, we provide it to GPT-4V and ask it to fix the script. If the script runs but the generated CAD doesn’t have a perfect score, we ask GPT-4V to correct discrepancies between the generated CAD and the engineering drawing. We ask it to do this by providing it with an image with four views of the CAD generated from the previously provided script (see \ref{cad_and_drawings} P4 for an example). These views show the CAD with hidden lines and coordinate systems visible for each view. If the CAD is still not receiving a perfect score by P7 (CAD Generation 5), we end the experiment.

    \begin{itemize}
        \item Scoring (6 points possible for each prompt, P4 – P7): Scoring for these prompts is identical to the scoring for P3.
    \end{itemize}

\end{itemize}

Please note that the scoring system used here primarily serves as an illustrative example to simplify the understanding of an aggregate score for readers. Practitioners may adjust the weights assigned to each question type based on specific requirements. We have made the raw data for each response available, facilitating benchmarking and allowing for flexibility in evaluating performance when varying the importance of different factors. 

\begin{table*}[h]
    \centering
    \caption{Summarized results from section \ref{sec:cad_and_engineering_drawings}}
    \label{tab:drawing_analysis}
    \footnotesize
    \begingroup
    \setlength{\tabcolsep}{3pt}
    \begin{tabularx}{\textwidth}{@{}X|cccccccccc@{}} \toprule
         Experiment Name &  Exp 1 & Exp 2 & Exp 3 & Exp 4 & Exp 5 & Exp 6 & Exp 7 & Exp 8 & Exp 9  \\ 
         \midrule
         \textbf{(P1) Part Description}& \textbf{0/1} & \textbf{0/1} & \textbf{0/1} & \textbf{1/1} & \textbf{0/1} & \textbf{0/1} & \textbf{0/1} & \textbf{0/1} & \textbf{0/1} \\
         \midrule
         8.0 Dim. Extraction and Label & 2 & 2 & 2 & 2 & 2 & 2 & 2 & 2 & 1* \\
         5.0 Dim. Extraction and Label & 2 & 2 & 2 & 2 & 2 & 2 & 2 & 2 & 2 \\
         12.0 Dim. Extraction and Label & 2 & 2 & 2 & 2 & 2 & 2 & 2 & 2 & 2 \\
         \diameter5.0 Dim. Extraction & 2 & 2 & 2 & 2 & 2 & 2 & 2 & 2 & 2 \\
         4.0 Dim. Extraction and Label & 1* & 2 & 2 & 2 & 2 & 2 & 2 & 2 & 1* \\\
         Additional Dim. Extraction & 0 & -1 & 0 & 0 & 0 & 0 & 0 & 0 & 0 \\
         \midrule
        \textbf{(P2) Dimension Extraction} & \textbf{9/10} & \textbf{9/10} & \textbf{10/10} & \textbf{10/10} & \textbf{10/10} & \textbf{10/10} & \textbf{10/10} & \textbf{10/10} & \textbf{8/10} \\
         \bottomrule
    \end{tabularx}
    \endgroup
    {\footnotesize*These entries which have a score of 1 always correspond with successful dimension extraction but incorrect label assignment.}
\end{table*}

\begin{figure}[h]
    \centering
    \includegraphics[width=1.0\linewidth]{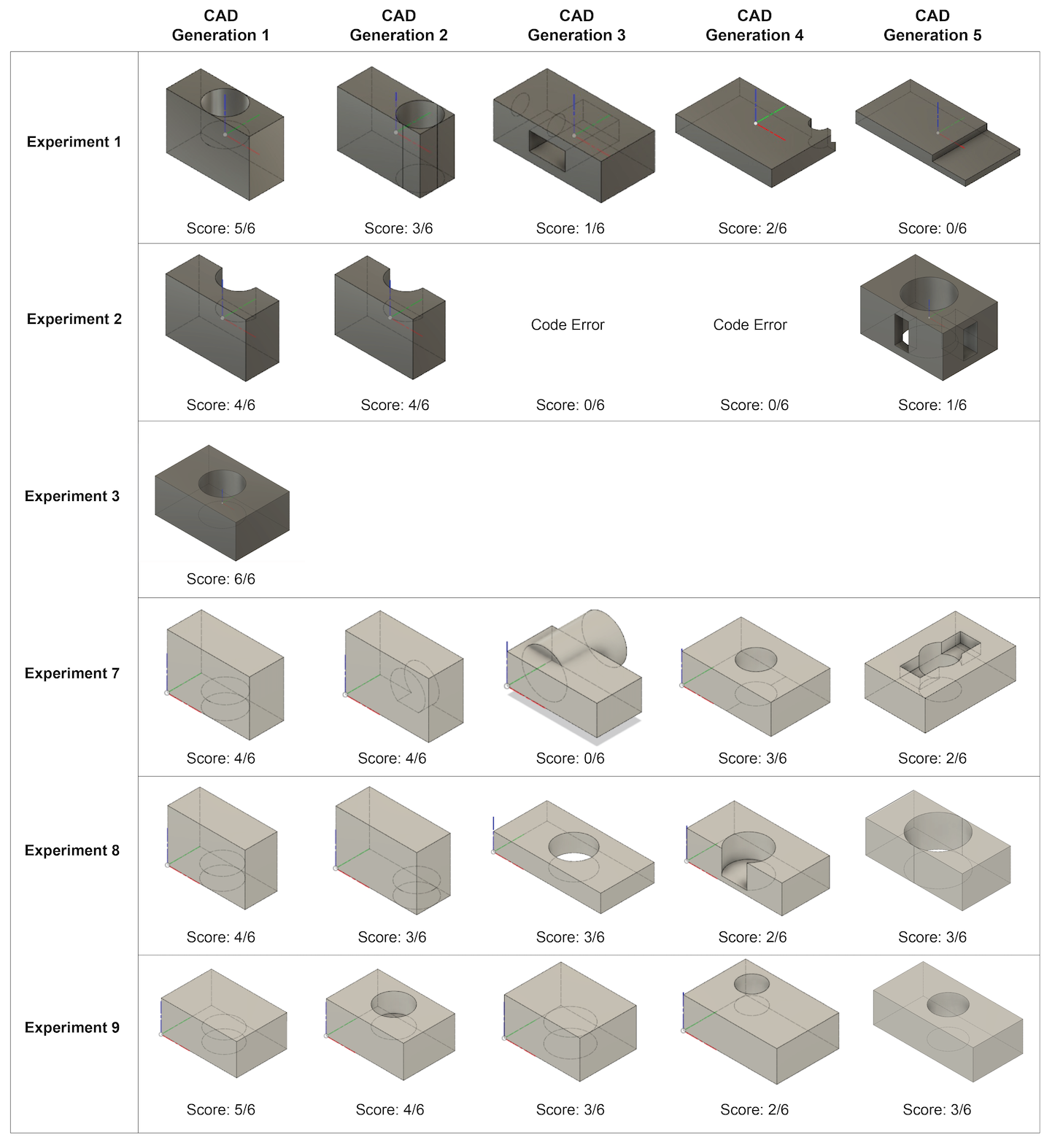}
    \caption{Results of the CAD Generation prompts, CAD Generation 1 - CAD Generation 5 (P3 - P7). Experiments 1 - 3 were generated using CadQuery and Experiments 7 - 9 were generated using OpenSCAD. Experiment 3 was the only experiment that generated perfect score CAD on the first iteration. Experiments 4 - 6, the FeatureScript experiments, are not shown here, since they had persistent code errors and never generated viable CAD.}
    \label{fig:cad_iterations}
\end{figure}

\paragraph{Discussion: Engineering Drawing Analysis and CAD Generation}
Based on the results from P1 and P2 (see Table \ref{tab:drawing_analysis}), which quantify GPT-4V's ability to analyze an engineering drawing, we conclude that the model generally understands the content in the drawing, but struggles with interpreting the drawing's details. For P1 in eight out of the nine experiments, GPT-4V incorrectly describes the part as a block with a hole ``through'' it. While it understands the part generally - a block with a hole - it does not pick up on the notation in the drawing that indicates the hole is blind rather than through. In the one experiment (experiment 4) where it received a correct score for the part description, it called the part a ``rectangular block with a cylindrical hole or recess in it'' and a ``generic block with a hole.'' While this qualifies as an accurate description, it does not demonstrate whether GPT-4V recognizes the blind hole in the drawing or not.

After being told that the part in question is a block with a blind hole (P2), GPT-4V is generally good at extracting dimensions from the drawing, receiving a perfect P2 score for six out of nine experiments (see Table \ref{tab:drawing_analysis}). Across all nine experiments, GPT-4V always extracts all five dimensions from the drawing. Two-thirds of the time it assigns all of the dimensions appropriate labels. It has the most trouble naming the hole depth dimension: in experiment 1 it calls it the ``height of the block (from the bottom right view)'' and in experiment 9 it calls it the ``width of the block.'' The relative difficulty in understanding what the 4.0 dimension represents in the drawing is consistent with its initial lack of understanding (in P1) that the drawing represents a block with a \textit{blind} hole. It is also interesting to note that GPT-4V is inconsistent in the labels it chooses for the three block dimensions -- varying between height/width/length, depth/height/width, height/depth/width, and depth/height/length -- perhaps reflecting a lack of consistent spatial reasoning or a lack of consistent norms used to label dimensions.

From the responses to P3-P7, we evaluate GPT-4V's ability to generate CAD using CAD scripting languages. We observe, see Fig.~\ref{fig:cad_iterations}, that GPT-4V rarely generates accurate CAD on the first attempt (P3), and CAD iterations (P4-P7) do not improve the CAD. For P3, only one out of the nine experiments (experiment 3, using CadQuery) leads to correctly generated CAD on the first attempt. For FeatureScript, GPT-4V cannot get out of syntax and function implementation errors for all five CAD generation iterations. The most common issue for P3 is not putting the hole on the correct face. We noticed this is because the hole extrusion direction is always linked with the dimension to which GPT-4V assigns the ``height'' label. The 5.0 block dimension is only assigned the height label three times, one of which is the sole experiment (experiment 3) where perfect score CAD is generated in P3. 

From the results from P4-P7, we conclude that visual feedback of the generated design from the previous prompt does not improve GPT-4V's CAD scripting ability. In fact, if GPT-4V generated incorrect CAD in P3, P4-P7 will never fully rectify the problematic CAD, and CAD Generation 5 (P7) will have a worse score than CAD Generation 1 (P3). A visualization of this finding can be seen in Fig.~\ref{fig:cad_iterations}. For the CadQuery and OpenSCAD experiments, a general reduction in CAD score occurs in CAD Generation 3 (P5), where GPT-4V consistently forgets the dimensions that it extracted in the original engineering drawing. 

In summary, we find that GPT-4V can pick up many aspects of the provided engineering drawing (e.g. general part depicted, many of the dimensions shown, etc.), but can struggle when it comes to understanding the details (e.g. recognizing the \textit{through} hole, labeling the through hole dimension, etc.). GPT-4V performs poorly when it comes to CAD generation, and we demonstrate that our attempts at visual, iterative improvements are unsuccessful. These findings imply that while GPT-4V can offer some assistance in preliminary design tasks, its current capabilities are not yet sufficient for detailed, precision-driven CAD work. Building on the findings from the evaluation of GPT-4V, future research should focus on enhancing the model's ability to interpret and process detailed engineering information. Another critical area for development is in CAD generation, where GPT-4V currently shows limitations. Future work should explore methods to improve the model's accuracy and efficiency in creating detailed CAD models, perhaps through advanced training techniques or integration with specialized CAD software. Additionally, there's a need to investigate how iterative feedback mechanisms can be better utilized by GPT-4V to make meaningful corrections and improvements in successive design iterations. Addressing these areas will be crucial in expanding the applicability of GPT-4V and similar VLM tools in more advanced and precision-dependent stages of the engineering design process.

In tandem with CAD generation and engineering drawing creation, engineers frequently aim to improve the design of the part, using iterative optimization approaches (e.g., using commercial tools such as nTop or SOLIDWORKS Simulation software). One of the commonly used iterative optimization approaches is structural topology optimization, which can help a designer reduce material usage while meeting some design requirements. In the next section, we turn to GPT-4V's ability to assist with topology optimization.

\subsection{Topology Optimization Analysis}\label{sec:to}

Structural topology optimization (TO) is a numerical approach for optimizing material distribution in structures under specific constraints, aiming for efficient material use while maintaining performance. The SIMP (Solid Isotropic Material with Penalization) method, a dominant approach in TO, models material properties using a density field, adjusting it iteratively to optimize design and adhere to stress or deformation constraints~\citep{bendsoe1989optimal}. In mechanical systems, the \emph{minimum compliance problem} focuses on finding a material density distribution, $\vx \in \mathbb{R}^n$, to minimize deformation under forces and boundary conditions~\citep{SIMP1988}. The problem is formulated as:

\begin{equation}
\begin{aligned}
\min_{\vx \in [0, 1]} \quad & c(\vx)  = F^T U(\vx)  \\
\textrm{s.t.} \quad & v(\vx)  = v^T \vx < \bar v\\
\end{aligned}
\end{equation}

Here, the objective is to minimize the compliance $c(\vx)$, with $F$ being the external load forces, $U(\vx)$ the displacements of nodes and solution of the equilibrium equation $K(\vx)  U(\vx)  = F$, and $K(\vx)$ the stiffness matrix which depends on the material distribution. The constraints include maintaining the volume fraction $v(\vx)$ below a specified limit $\bar v$ and ensuring the design variables $\vx$ remain within the bounds of 0 and 1, allowing for a gradient of material distribution from void to solid~\citep{SIMP1988, SigmundMaute2013}.

Optimal topologies are, however, often challenging to analyze for human experts. The topologies that result from the optimization process may be mathematically optimal, but present practical challenges in manufacturability and analysis and can be non-intuitive for human designers. We consider the analysis of topology optimization images as a test case to evaluate the use of visual-language models to assist humans in interpreting complex topologies. Note, in these experiments we do not ask the vision-language model to perform topology optimization, but rather to determine the inputs to or interpret the results of topology optimizations.

\subsubsection{Volume Fraction Estimation}
\label{to:volume_fraction_estimation}

In this experiment, we task the model with calculating the volume fraction from an optimized topology depicted in an image. This involves measuring the proportion of black material in the given domain and determining the relevant ratio. The challenge is initially approached using GPT-4V's visual analysis capabilities alone. Following this, GPT-4V employs its code interpretation abilities to address the task. We aim to obtain an accurate answer within a 5\% error threshold.

\begin{figure}

\begin{contextbox}[]{Volume Fraction Estimation}

\begin{promptbox}{(Quantitative analysis without tools)}

Can you estimate the volume fraction of this structure? Consider that white means the absence of material and the initial domain is a square of size 256x256.\\
\begin{center}
\includegraphics[height=0.1\textheight, keepaspectratio]{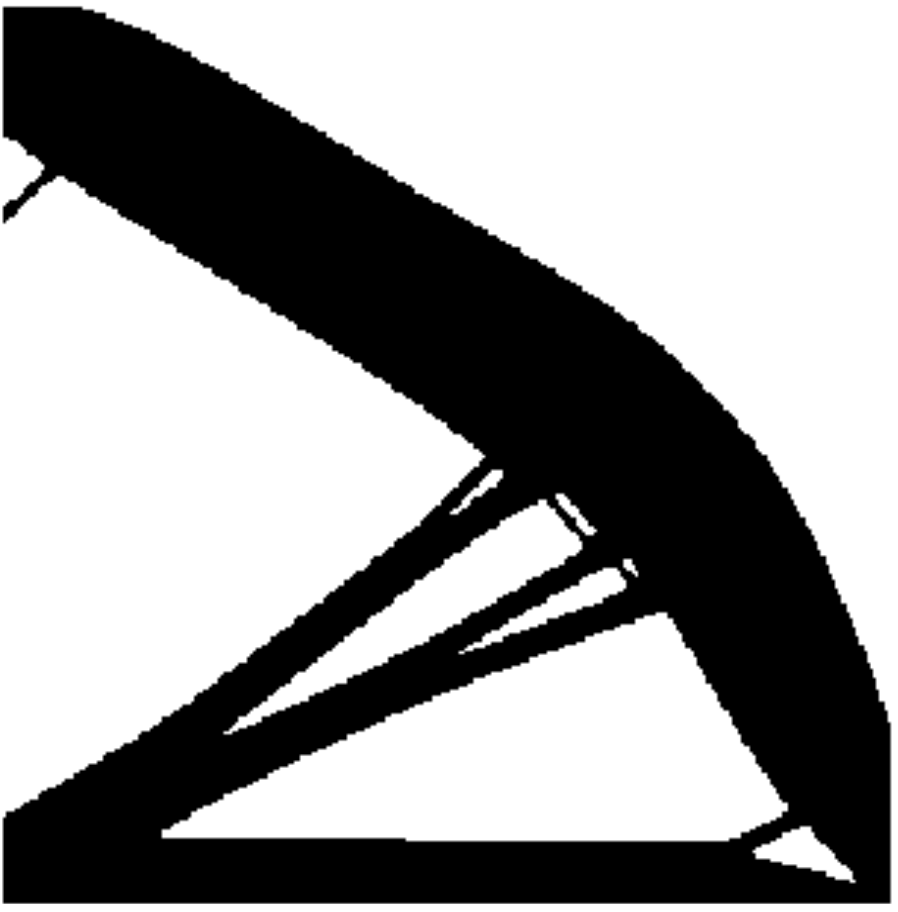}    
\end{center}
\textbf{Target:} 0.40

\tcblower

\begin{enumerate}[leftmargin=*,labelindent=0mm]
    \item To determine the volume fraction, \hl{we can evaluate the percentage of the black area (material presence) relative to the total area of the square.}
    
    \item The formula for volume fraction \( \phi \) is:
    \[ \phi = \frac{\text{area of the black region}}{\text{total area}} \]
    
    \item From the image, the total domain is \(256 \times 256\), so the total area \( A_{\text{total}} \) is 65536 pixels.
    
    \item Let's compute the area of the black region \( A_{\text{black}} \).
    
    \item After analyzing the image:
    \( A_{\text{black}} \) is \hle{approximately 42540 pixels} (this is an estimation based on the provided image).
    
    \item Plugging into the formula:
    \[ \phi = \frac{42540}{65536} \approx 0.649 \]
    
    \item Therefore, \hle{the estimated volume fraction of the structure is approximately 0.649} or 64.9\%.
\end{enumerate}

\textit{When asked to try again, the result is different (50.8\%) but still incorrect.}
\end{promptbox}

\begin{promptbox}{(Quantitative Analysis with tools - code interpreter)}

Can you compute the volume fraction of this structure (refer to previous figure in subsection)? White means absence of material and the square domain is 256x256 pixels.\\

\textbf{Input File:} Same image as above. \\ %

\textbf{Target:} 0.40

\tcblower

Sure! I'll start by loading the image and visualizing it. Then, I'll \hl{compute the volume fraction based on the number of white pixels (representing the absence of material) and the total number of pixels in the domain.}

[Code to load the image in Python]

Here's the loaded image. Now, I'll compute the volume fraction. The volume fraction will be determined by \hl{the ratio of non-white pixels (material presence) to the total number of pixels in the domain.}

[Code to compute the volume fraction]

\hl{\textbf{Result:} 0.3852859438817139}

\end{promptbox}
\end{contextbox}
\caption{Volume fraction estimation.}
\label{fig: to-volume_fraction_estimation}
\end{figure}

\paragraph{Methodology} In this experiment, we provide a VLM with an image of an optimized topology, and prompt the VLM to estimate the volume fraction of the structure. We provide the following contextual information: ``Consider that white means the absence of material and the initial domain is a square of size 256x256,'' as shown in the prompt in Fig.~\ref{fig: to-volume_fraction_estimation}. 

We perform this experiment for 100 optimized designs with no floating material as well as 50 non-optimized designs which have floating material. We run the experiment with these different prompting strategies, as seen in Table~\ref{tab:quantitative-to-table}: without domain-specific prompting (w/o Expertise), incorporating domain expertise in the prompt (w/ Expertise, i.e. "you are an expert engineer in the field of topology optimization"), and enriching the expert prompt with Chain-of-Thoughts rationale (w/ CoT, i.e. "answer using a step-by-step rationale"). The quantitative results are shown in Table~\ref{tab:quantitative-to-table}.

\begin{table}[ht!]
\centering
\caption{
Volume Fraction Error (VFE) is the percent error when predicting the volume fraction of an optimized design. Floating Material Error is for the classification task of determining if a design has floating material or not (50\% error would be a random guess).
}
\label{tab:quantitative-to-table}
\footnotesize
\begin{tabular}{@{}lccc@{}}
\toprule
& $\downarrow$ VFE (\%) & $\downarrow$ FME (\%)  & $\downarrow$ FME (50/50) (\%)\\
\midrule
w/o Expertise       & 45.71 $\pm$ 2.45  &  47.15 $\pm$ 1.11  & 42.08 $\pm$ 2.83 \\
w/ Expertise        & 44.13 $\pm$ 1.11  &  44.71 $\pm$ 1.15  & 39.75 $\pm$ 2.54 \\
w/ CoT              & 43.62 $\pm$ 0.77  &  43.90 $\pm$ 0.23  & 39.50 $\pm$ 2.49 \\
\bottomrule
\end{tabular}

\end{table}

We provide quantitative experiments in Table~\ref{tab:quantitative-to-table}, where the aim is to estimate Volume Fraction (VF) and the presence of Floating Material (FM). Table~\ref{tab:quantitative-to-table} presents the error metrics across 100 optimized topologies (first and second columns) and a balanced set of 50 optimized alongside 50 un-optimized topologies with randomized floating material (third column). It focuses on Volume Fraction Error (VFE) for assessing the accuracy in estimating material usage and Floating Material Error (FME) for identifying disconnected components in designs, with a baseline error expectation set at 50\% for random chance. %
We can compare our three prompting strategies (w/o Expertise, w/ Expertise, and w/ CoT). Integrating engineering expertise into the model's prompting strategy (w/ Expertise) improves the results, as seen by lower VFE than w/o Expertise in Table~\ref{tab:quantitative-to-table}. However, the addition of step-by-step explanations (CoT) yields only marginal improvements in accuracy.  
Overall, the capacity to estimate Volume Fraction and detect Floating Material is relatively poor.
This observation underlines the limitations of relying solely on a vision encoder for precise topology optimization tasks, such as estimating Volume Fraction and detecting Floating Material. It highlights the necessity of employing external analytical tools, like a code interpreter or dedicated vision modules, to achieve more accurate and reliable outcomes.

 \paragraph{Discussion: Volume Fraction Estimation}

In Fig.~\ref{fig: to-volume_fraction_estimation}, we challenge the model to quantitatively estimate the volume fraction of an optimized topology. The model accurately defines the task as ``\emph{evaluating the percentage of the black area (material presence) relative to the total area of the square}.'' However, its initial attempts to count black pixels and calculate the material percentage yield highly inaccurate results. This inconsistency persists across multiple trials, each providing different and incorrect answers. 

To address these limitations, we introduce a code interpreter (third prompt in Fig.~\ref{fig: to-volume_fraction_estimation}), enabling the model to use a Python script for the estimation. This approach significantly improves accuracy, bringing the estimate close to the target within a reasonable margin of error. 
This experiment highlights two key insights: Firstly, it underscores the limitations of the vision encoder (at least for the version of GPT-4V used in this study) in handling precise quantitative assessments based on images. Secondly, it demonstrates the effectiveness of integrating coding tools in overcoming these limitations, showcasing the synergistic potential of combining AI's interpretive capabilities with precise, code-based calculations for more accurate and reliable results.

\subsubsection{Technical  Caption Generation and Analysis}

\label{to:technical_knowledge}

\begin{figure}
\begin{contextbox}[]{Technical  Captioning}
\begin{promptbox}{(Generic Prompt)}
Can you describe this image? \\
\begin{center}
\includegraphics[width=0.7\linewidth]{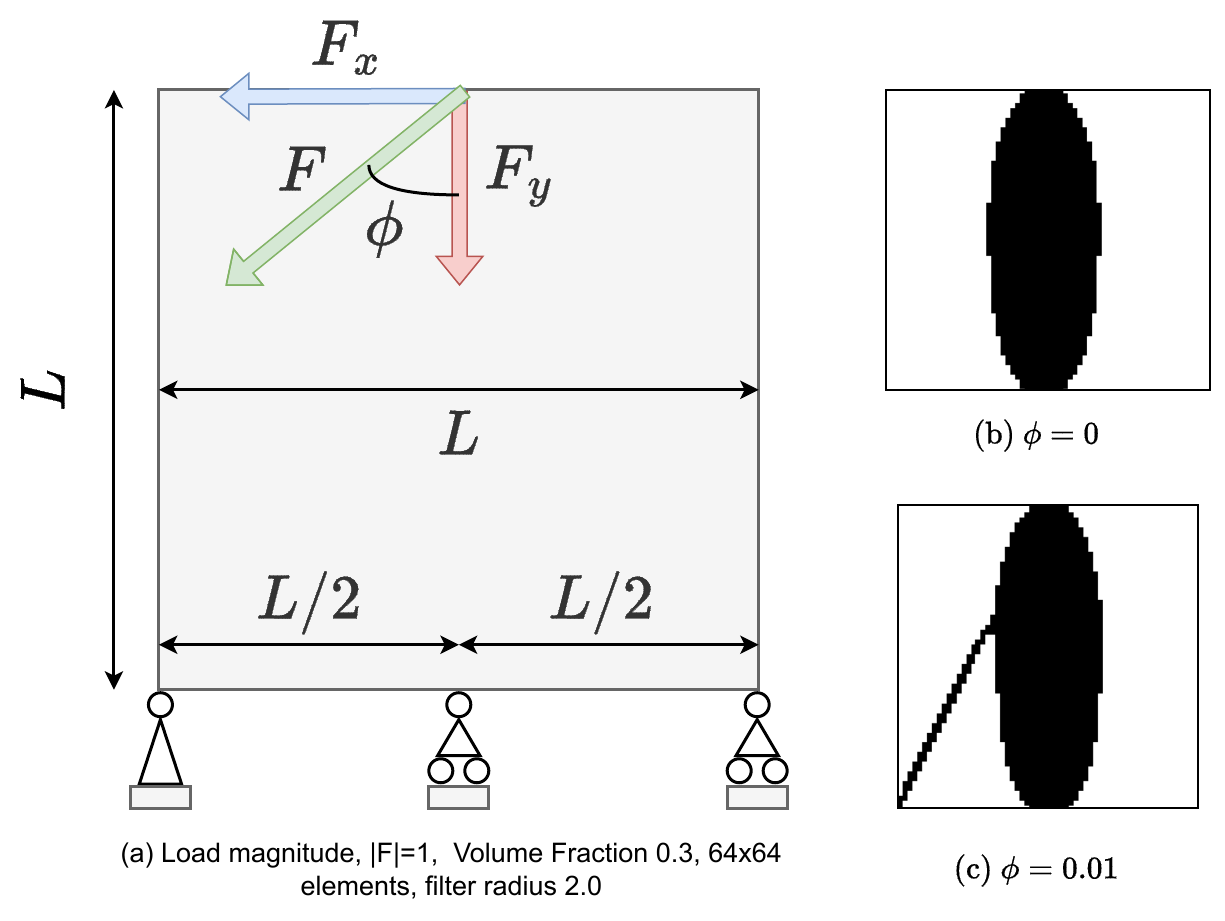}    
\end{center}

\textbf{Target:} Significant effect small changes in boundary conditions and loading can have on the optimal topology. The boundary conditions (a) with corresponding optimized designs for (b) and (c).

\tcblower

The image consists of three separate figures labeled (a), (b), and (c).
\begin{enumerate}[labelindent=0mm,leftmargin=*]
    \item (a) This figure depicts a rectangular frame with a length labeled as "L". The rectangle is divided into two halves by a center line, and each half is labeled "$\frac{L}{2}$". \hle{At the bottom corners of the rectangle, there are two hanging weights}. On the top of the rectangle, there's a \hl{force vector labeled "F" making an angle "$\phi$" concerning the horizontal axis.} This force "F" is decomposed into two components: a horizontal force "$F_x$" and a vertical force "$F_y$". The caption underneath mentions that the load magnitude $|F|$ is 1, the volume fraction is 0.3, there are 64x64 elements, and the filter radius is 2.0.

    \item (b) This figure shows a square with an oval-shaped black region in the center. The caption below the square mentions "$\phi=0$".

    \item (c) Similar to (b), this figure shows a square. However, the black oval-shaped region in the center is slightly tilted to the right and has a \hl{diagonal black stripe running from the bottom left to the top right}. The caption below the square indicates "$\phi=0.01$".
\end{enumerate}

It appears that \hl{the images might be related to some form of structural or mechanical analysis, possibly demonstrating how the structure or material responds to different load angles.}
\end{promptbox}

\begin{promptbox}{(Specialized prompt)}
    
Answer the same question (refer to the previous image) using this instruction:
You are an expert engineer with years of experience in topology optimization and engineering design.\\

\textbf{Target:} Significant effect small changes in boundary conditions and loading can have on the optimal topology. The boundary conditions (a) with corresponding optimized designs for (b) and (c).

\tcblower

The image represents a \hl{topology optimization problem, a technique frequently used in structural and mechanical design to determine the best material distribution under specific loading and boundary conditions.}

\begin{enumerate}[labelindent=0mm,leftmargin=*]
    \item (a) This figure illustrates a common boundary condition setup for a topology optimization problem. The depicted domain, a rectangle of size ``L'', represents a structure or component. \hle{Two external point loads, represented by weights, are applied at the bottom corners.} On the top, there's a \hl{force vector ``F'' applied at an angle ``$\phi$'' to the horizontal.} This force is decomposed into its horizontal ``$F_x$'' and vertical ``$F_y$'' components. It's essential to note the given parameters: the magnitude of the applied load is $|F|=1$, and the design space has been discretized into a 64x64 grid. \hl{The volume fraction of 0.3 suggests that only 30\% of the total domain will be occupied by material}, while the rest will be void. The mention of a filter radius of 2.0 indicates the usage of a \hl{filtering technique}, which is often employed in topology optimization to ensure mesh-independent results and \hl{to smoothen the material distribution}.

    \item (b) \hl{This figure depicts the optimized material layout for a direct vertical load}, as indicated by ``$\phi=0$''. The black region represents the material presence, suggesting a symmetric structure optimized to handle vertical loads efficiently. The shape's orientation and form are dictated by the combination of the loading, boundary conditions, and design domain, resulting in the most efficient load-bearing structure with minimal material usage.

    \item (c) Here, the optimized topology is presented for a slightly angled load, as denoted by ``$\phi=0.01$''. The slight tilt in the material region and the additional diagonal brace-like feature suggest an optimized design that caters to both vertical and small horizontal components of the applied force. This shows the adaptability of the topology optimization technique in producing efficient designs for varied loading conditions.
\end{enumerate}

In summary, \hl{these images provide a clear depiction of how topology optimization evolves structural designs according to specified load cases}, ensuring efficient material usage and optimal load-bearing capability.

\end{promptbox}
\end{contextbox}
\caption{Technical captioning of TO images.}
\label{fig: to-technical_knowledge}
\end{figure}

\paragraph{Methodology }In this experiment, we task a VLM with captioning a technical diagram using a basic prompt, with the diagram inspired by the experiment in Fig.~7 from~\citep{woldseth2022use}. Initially, we employ a generic prompt for captioning. Then, we enhance the task by incorporating details about the system's technical expertise, providing a more in-depth and knowledgeable description of the diagram. These can be seen in Fig.~\ref{fig: to-technical_knowledge}.
We would like to see that the model understands that small variations of constraint configurations (in this case load direction) can greatly change the optimized topology.

Additionally, we conducted a quantitative experiment to assess how effectively a VLM processes textual information within images, such as descriptive captions (Table~\ref{tab:to-more-quantitative}). We created ten problem setups modeled after the example in Fig.~\ref{fig: to-technical_knowledge}, varying aspects of the image such as the number of elements in the x and y axes, volume fraction, load magnitude, and angle of application, along with the positions and directions of loads and the types of boundary conditions.

Our focus was to evaluate GPT-4V's ability to extract specific details from captions—information that exists solely as text within the image. We also tested the model's accuracy in identifying the positions and directions of loads, as well as the types and placements of boundary conditions. Additionally, we examined the model's capability to generate a functional topology optimization Python script based on the given data.

The findings indicate that GPT-4V performs well in interpreting and retrieving data from image captions. However, it shows limitations in accurately locating loads and boundary conditions. The Python scripts generated by the model were structurally sound but typically included minor mistakes in indexing and variable names, reflecting areas needing improvement in the model's programming language generation capabilities.

\begin{table}[ht!]
    \centering
    \caption{This table presents whether the model was successful (a score of 1) or not (a score of 0) at caption analysis, loads and boundary positioning, and code validation when analyzing a topology optimization image. For caption analysis, the terms nelx and nely are the number of elements in x-direction and y-direction respectively; $|F|$ is the force module; VF is the volume fraction; R is the filter radius; $\phi$ is the angle of application for the load.}
    \label{tab:to-more-quantitative}
    \setlength{\tabcolsep}{3pt}
    \begin{tabular}{@{}c|cccccc|cc|c@{}}
    \toprule
    & \multicolumn{6}{c|}{Caption Analysis} & \multicolumn{2}{c|}{Position} & Validation \\
    Problem & nelx & nely & $|F|$ & VF & R & $\phi$ & Load & BC & Code Runs \\
    \midrule
      P 1   & 1 & 1 & 1 & 1 & 1 & 1 & 0 & 0 & 0 \\
      P 2   & 1 & 1 & 1 & 1 & 1 & 1 & 0 & 0 & 0 \\
      P 3   & 1 & 1 & 1 & 1 & 1 & 1 & 0 & 1 & 0 \\
      P 4   & 1 & 1 & 1 & 1 & 1 & 1 & 0 & 0 & 0 \\
      P 5   & 1 & 1 & 1 & 1 & 1 & 1 & 1 & 1 & 0 \\
      P 6   & 1 & 1 & 1 & 1 & 1 & 1 & 1 & 1 & 0 \\
      P 7   & 1 & 1 & 1 & 1 & 1 & 1 & 1 & 1 & 0 \\
      P 8   & 1 & 1 & 1 & 1 & 1 & 1 & 0 & 0 & 0 \\
      P 9   & 1 & 1 & 1 & 1 & 1 & 1 & 1 & 0 & 0 \\
      P 10  & 1 & 1 & 1 & 1 & 1 & 1 & 0 & 0 & 0 \\
      \midrule
       Avg. & 1.0 & 1.0 & 1.0 & 1.0 & 1.0 & 1.0 & 0.4 & 0.4 & 0.0\\
      \bottomrule
    \end{tabular}
\end{table}

Table \ref{tab:to-more-quantitative} presents the results from testing the GPT-4V model on ten different problem setups, which are modifications of the example shown in Fig.~\ref{fig: to-technical_knowledge}. These modifications include variations in load positions and directions, domain sizes, volume fractions, and load application angles (represented as $\phi$). The table evaluates the model's performance in three critical areas:
(i) \emph{Caption Analysis}. We evaluate how accurately the model interprets text within images, such as captions.
(ii) \emph{Load and Boundary Positioning}. We evaluate the model's precision in identifying the correct positions for loads and boundary conditions.
(iii) \emph{Code Validation for Topology Optimization}. We evaluate the effectiveness in generating a usable Python script for minimum compliance topology optimization, highlighting issues like indexing errors or incorrect variable naming which prevent the script from running.
Each entry in the table is scored as 1 (correct answer) or 0 (wrong answer, partially correct, or non-functional script), summarizing the model's capability to handle textual and spatial information in images under varying complexity and setup parameters.
nelx: number of elements in x-direction; nely: number of elements in y-direction; $|F|$: force module; VF: volume fraction; R: filter radius; $\phi$: angle of application for the load.

\paragraph{Discussion: Technical Caption Generation and Analysis}

In Fig.~\ref{fig: to-technical_knowledge}, we evaluate the model's ability to interpret a complex diagram. The model delivers a comprehensive and accurate analysis, adeptly linking forces, their angles of application, and optimized topology. It accurately identifies the image as ``\emph{a structural or mechanical analysis, demonstrating how the structure or material responds to varying load angles}.'' This insight into how loading direction affects topology is a correct deduction of a difficult physical problem, showcasing the model's proficiency in understanding boundary conditions, loads, and their impact on structures.

However, the model encounters difficulty with the boundary conditions at the bottom center and right of the diagram, mistakenly interpreting them as ``\emph{two hanging weights},'' which is an incorrect assessment of the boundary sketch. This misinterpretation is unexpected, particularly given the overall high-quality response and the accurate grasp of the problem's essence.

Further refining the prompt to emphasize engineering concepts, the model again provides a largely accurate response, delving deeper into topics like loads, volume fraction, and filtering radius. Yet, it repeats the same error concerning the boundary conditions, suggesting ``\emph{Two external point loads, represented by weights, are applied at the bottom corners.}'' This persistent mistake indicates a gap in the model's global understanding of the scenario, revealing a vulnerability to misconceptions in specific contexts.

\subsubsection{Invalid Design}
\label{to:invalid_design}

\paragraph{Methodology} The task involves identifying invalid designs, specifically floating materials, based on a given prompt. The objective is for the model to recognize, independently and without prior information, the presence of a disconnected component within a low-resolution design (64x64). Following this recognition, the model is expected to assess the design's overall validity and the quality of the low-resolution grid. Lastly, we test if the model is capable of suggesting potential improvements to rectify the identified issues.

\begin{figure}
\begin{contextbox}[]{Invalid Design Identification}

\begin{promptbox}{(Understanding)}

Can you describe this image?\\
\begin{center}
\includegraphics[height=0.1\textheight, keepaspectratio]{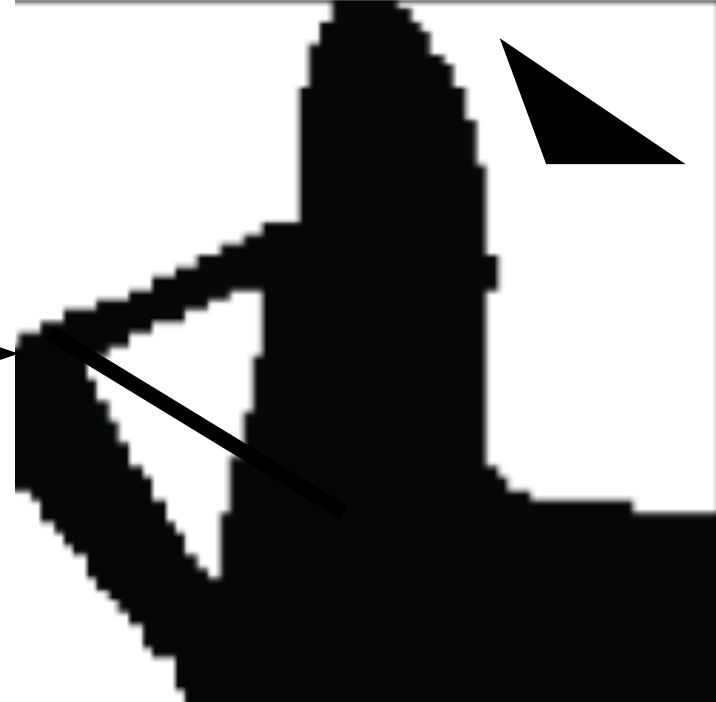}    
\end{center}

\tcblower

The image showcases a \hl{topology, possibly a result of a topology optimization process}. Here's a description of the depicted design:

The topology is primarily characterized by a black structure against a white background. The structure appears to have:

\begin{enumerate}[labelindent=0mm,leftmargin=*]
    \item \textbf{Central Tower:} \ldots

    \item \textbf{Diagonal Bracing:} \ldots

    \item \textbf{Horizontal Base:} \ldots

    \item \textbf{\emph{Isolated Triangle:}} Towards the \hl{top right corner of the image, there's an isolated triangular shape.} Its \hl{disconnection from the main structure} is intriguing. This shape could either be a byproduct of the optimization process, represent a region of minor stress concentration, or perhaps an artifact of the resolution used.
\end{enumerate}

\ldots

\end{promptbox}

\begin{promptbox}{(Evaluation)}

Do you think this (refer to previous design) is a "valid" topology?

\tcblower

Evaluating the provided topology:

The design consists of a primary central structure with distinct features, such as a diagonal brace and an isolated triangular shape. Here's an assessment:

\begin{enumerate}[labelindent=0mm,leftmargin=*]
    \item \textbf{Pixelation Sharp Corners:} The design seems pixelated, especially evident in the diagonal brace, which appears jagged or stair-like. Sharp corners or sudden changes in geometry, as suggested by this pixelation, could introduce stress concentrations in a real-world structure, potentially leading to failure.

    \item \textbf{Isolated Triangle:} \hl{The triangular shape towards the top right is disconnected from the main structure.} In practical manufacturing scenarios, this isolated feature might be problematic to manufacture, and its \hle{disconnected nature might render it ineffective in a load-bearing role.}
    
    \item \textbf{Material Usage:} \hle{The design seems to be focused on minimal material usage}, which is a typical goal of topology optimization. However, this also means that any structural weaknesses are amplified since there's less material to absorb or distribute unexpected stresses.
\end{enumerate}

In conclusion, while the design showcases interesting structural features that may be optimized for specific load cases, its pixelation and \hl{the presence of the isolated triangle raise questions about its practical validity} and real-world performance. Before pursuing manufacturing, it would be advisable to refine the design to smooth out pixelated regions, \hl{consider the connectivity of all features}, and validate its performance using finite element analysis or similar computational tools.

\end{promptbox}

\begin{promptbox}{(Optimization)}

How would you improve such a design (refer to previous design)?

\tcblower

Improving the design resulting from a topology optimization process often requires a combination of computational refinement and engineering judgment. Here's how the design could be improved:

\begin{enumerate}[labelindent=0mm,leftmargin=*]
    \item \textbf{Smoothen Pixelation:} [...]

    \item \textbf{Address Disconnected Components:}
    \begin{itemize}[labelindent=0mm,leftmargin=*]
        \item \hl{The isolated triangular shape should be either integrated into the main structure or removed if it doesn't provide any significant functional benefit.}
        \item \hle{A re-evaluation of boundary conditions and load cases can ensure that disconnected components are not formed.}
    \end{itemize}

    \item \textbf{Review \& Refine Structural Features:}[...]

    \item \textbf{Material \& Manufacturing Considerations:}[...]

    \item \textbf{Iterative Validation:}[...]

\end{enumerate}

\end{promptbox}
\end{contextbox}
\caption{Invalid design identification.}
\label{fig: to-invalid_design_identification}
\end{figure}

\paragraph{Discussion: Invalid Design}

In Fig.~\ref{fig: to-invalid_design_identification}, we task the model with identifying the presence of floating material in a design, specifically a detached triangle in the top right corner. The model accurately recognizes the issue as a result of a topology optimization process, correctly noting the disconnection of the ``\emph{isolated triangular shape from the main structure}'' in the top right.

When queried about the structure's validity, the model identifies the floating material but its response lacks full clarity on the implications of such a flaw. While it correctly points out that ``\emph{this isolated feature might be problematic to manufacture, and its disconnected nature might render it ineffective in a load-bearing role},'' it fails to emphatically state that a disconnected component invariably compromises structural integrity and manufacturability. 
In Table~\ref{tab:quantitative-to-table}, we present a quantitative analysis of our experiment on detecting Floating Material using the vision encoder, illustrating the complexities and challenges inherent in accurately identifying such presence or absence of small disconnected components in a broad range of cases.

Regarding the design's optimization objectives, the model suggests a focus on ``\emph{minimal material usage}.'' This is a common requirement in topology optimization, but it oversimplifies the broader range of performance requirements typically involved in such processes.

When asked about improving the design, the model sensibly proposes ``\emph{integrating the isolated triangle into the main structure or removing it if it lacks functional benefit}.'' This is a valid solution to address the floating material issue. However, its subsequent recommendation to ``\emph{re-evaluate boundary conditions and load cases}'' as a method to eliminate disconnected components is somewhat misguided. The more appropriate approach would involve refined optimization strategies and post-processing techniques.

In summary, while GPT-4V effectively identifies floating material and offers viable solutions, it falls short of fully understanding the criticality of disconnected components. Floating material or disconnected parts in a topology-optimized design invariably render it structurally unsound from an engineering point of view or unmanufacturable without further optimization or processing.

\subsection{Summary} In order to assess GPT-4V's performance in select detail design tasks, we performed evaluations on material selection using an Ashby chart, on engineering drawing analysis and CAD generation, and on supporting the understanding and analysis of topology optimization. Our findings are discussed below.

\begin{enumerate}[(i)]
    \item \textbf{Material Selection~\ref{sec:material_selection}} 
    Can GPT-4V effectively assist in material selection based on property charts and design requirements?
    \begin{itemize}
    \item We find that GPT-4V can be helpful at pointing out material families that meet general specifications, but that it struggles to identify materials that match specific numerical requirements.
    \end{itemize}
    \item \textbf{Engineering Drawing Analysis and CAD Generation~\ref{sec:cad_and_engineering_drawings}}
    How accurately can GPT-4V extract and interpret information from engineering drawings? What is GPT-4V's proficiency in generating and iteratively improving CAD scripts from engineering drawings?
    \begin{itemize}
        \item Only 11\% of the time was GPT-4V successful in describing a block with a blind hole part. Once told it was a block with a blind hole, GPT-4V was successful 67\% of the time in extracting all dimensions from the engineering drawing and assigning the dimensions appropriate names.
        \item Preliminary findings suggest that GPT-4V struggles with CAD generation of the block-with-blind-hole part, as it only succeeded once in nine attempts to generate correct CAD on the first try. Also, its iterative ability for CAD correction appears limited, as it doesn't successfully correct incorrect CAD in subsequent iterations.
    \end{itemize}

    \item \textbf{Topology Optimization Analysis~\ref{sec:to}} Can VLMs properly analyze aspects of an optimized topology, such as the volume fraction and the presence of floating material? Furthermore, can VLMs interpret images with technical captions to identify the correct positions for loads and boundary conditions?
    \begin{itemize}
        \item While GPT-4V correctly defines the concepts of volume fraction and floating materials in terms of topology optimization, Table~\ref{tab:quantitative-to-table} shows that GPT-4V has high error in estimating volume fraction or identifying floating material from images. 
        \item In terms of analyzing images and captions to identify the number and location of elements like forces and boundary conditions, Table~\ref{tab:to-more-quantitative} shows that GPT-4V could consistently extract values from captions, however, could only identify the position of loads and boundary conditions from images 40\% of the time. Lastly, GPT-4V could never generate executable topology optimization code. 
    \end{itemize}
\end{enumerate}

We highlight the strengths and weaknesses of GPT-4V's response by marking the more relevant and high-quality sections in green, and the incorrect, out-of-context, or low-quality parts in brown.

\section{Manufacturing and Inspection}\label{sec:manufacturing}
\paragraph{Overview and Motivation}

Here we focus on assessing the performance of GPT-4V in manufacturing and inspection-related tasks. Our motivation relies on the visual cues that engineers often use to understand the practical aspects of manufacturing complex geometric artifacts. This multimodal information requires expertise in understanding images as well as manufacturing knowledge. As GPT-4V shows potential for task-specific image analysis, we evaluate its potential for manufacturing and inspection. The field of manufacturing is broad and discussing the complete potential of multimodal LLMs for all manufacturing tasks is out of the scope of our work. To this end, we focus on selective manufacturing tasks that can provide useful insights in assessing the capabilities of these multimodal LLMs. Specifically, we focus on design for manufacturing (DfM) and post-manufacturing inspection tasks. Both of these topics are critical for manufacturing applications in industry and demand extensive domain-specific knowledge. We draw particular attention to understanding the manufacturability of 3D CAD models only from images. Note that manufacturability traditionally refers to the relative ease with which a part can be manufactured \citep{budinoff2019geometric, budinoff2021will, yang2015additive}. Ensuring the manufacturability of a new part is a major challenge and requires careful analysis and expertise. The potential of an automated tool for this purpose would increase manufacturing productivity by a large margin. Multimodal LLMs may help industries build next-generation tools for automating these types of tasks. Our analysis can be thought of as an early evaluation of multimodal LLMs and their manufacturing knowledge and reasoning. For brevity, we divide the Design for Manufacturing section into two parts: additive and subtractive manufacturing. Based on existing literature, we query GPT-4V with images of 3D CAD models and assess its manufacturability response against the ground truth.

\subsection{Design for Manufacturing}
Design for manufacturing (DfM) is a popular concept that studies the manufacturability of an engineering design~\citep{webb200845nm}. The DfM field is broad, as manufacturability is dependent on the materials used, the specific manufacturing method employed (e.g. additive, subtractive, etc.), and the particular tools utilized for manufacturing (e.g. which type of 3D printer). We explore GPT-4V's ability to assist with DfM for two popular manufacturing methods: additive and subtractive.

\subsubsection{Design for Additive Manufacturing (DfAM)}\label{mfg:dfam}

Additive manufacturing (AM) has become increasingly popular as a fabrication method in recent years \citep{attaran2017rise}. AM first became popular because of its usefulness in rapid prototyping, but it is also utilized for low quantities of design-varying parts in aerospace and automotive component manufacturing \citep{attaran2017rise}. Design constraints for AM vary considerably by the additive system used, and assessing manufacturability of a design often requires some experimentation. However, through manufacturing experience, engineers and machinists often develop manufacturing guidelines or rules for designing a part to be manufactured with a specific process. It would be challenging to quantitatively assess GPT-4V's ability to predict the 3D-printability of part, because just as understanding of manufacturability varies from person to person, it would vary from person to model. However, we can provide the model with a set of unambiguous rules pertaining to 3D printability and ask the model to assess the printability of a part based on those rules. Hubs, a ProtoLabs company that offers on-demand manufacturing, created a chart, entitled ``Design rules for 3D Printing,'' encoding common design rules for AM based on printer type (see Figure~\ref{dfm-additive}) \citep{hubs_3dprinting}. There are ten specific design rules for FDM-printed parts. For example, one rule states that supported walls for an FDM printer can have a minimum thickness of 0.8 mm. These rules are heuristics and exceptions can be found, but the chart enables us to assess GPT-4V's ability to apply common fabrication rules to a design.

\begin{figure*}
    \centering
    \includegraphics[width=0.5\linewidth]{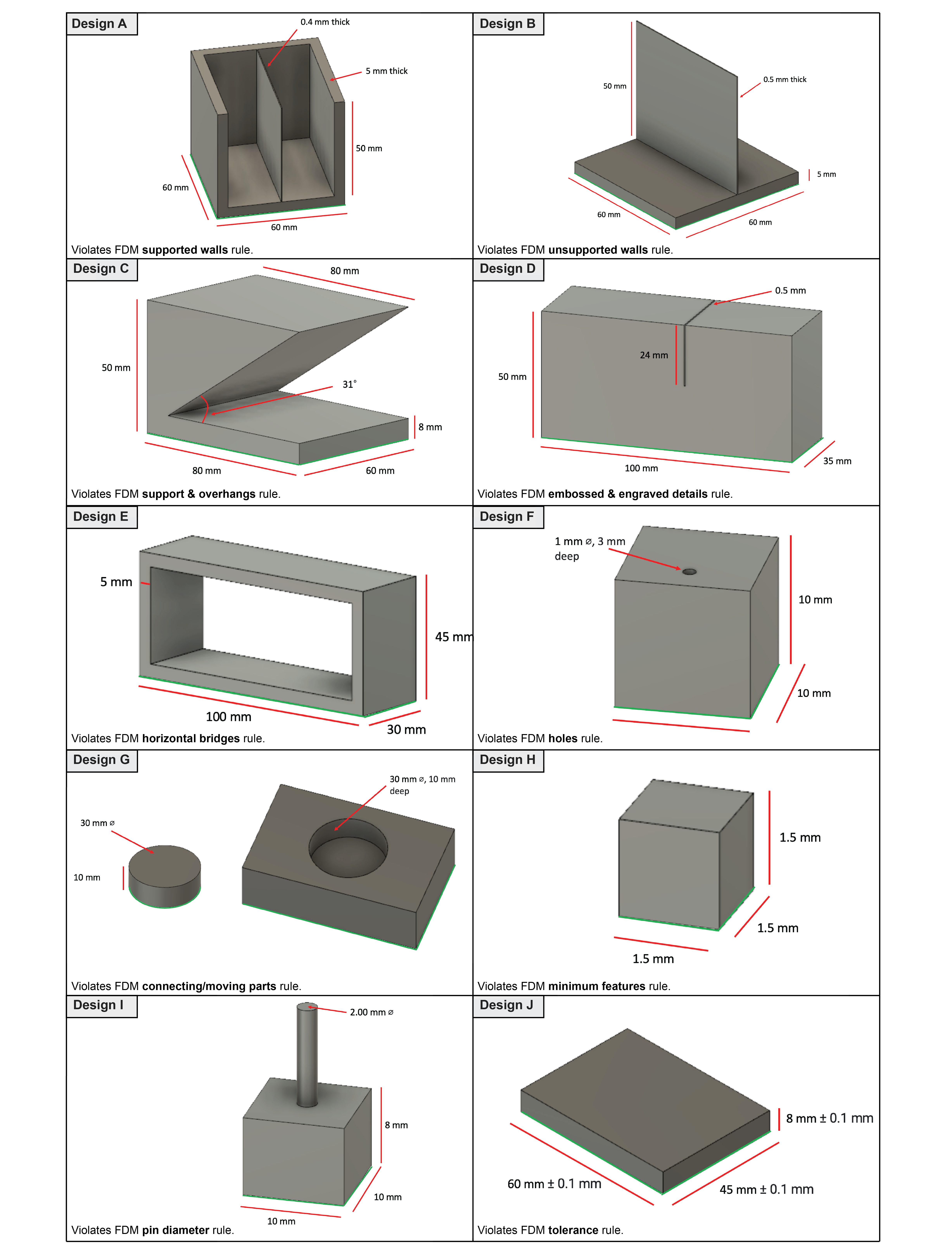}%
    \includegraphics[width=0.5\linewidth]{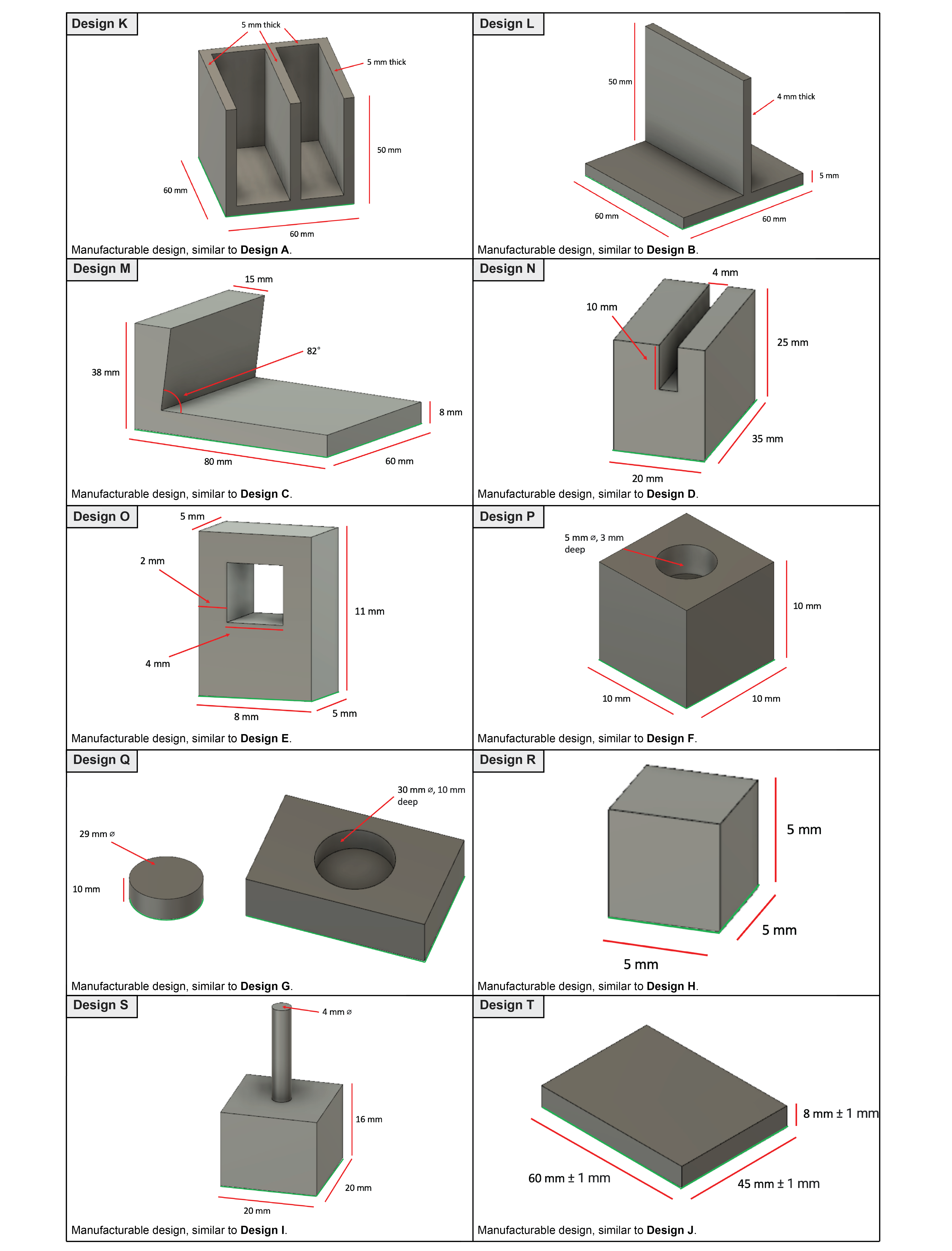}
    \caption{(A--J) Ten problematic designs, where each design violates one of the FDM AM rules on The Hub's chart for ``Design rules for 3D printing.'' The specific rule violated is noted below each design. (K--T) Ten manufacturable designs, each based on one of the problematic designs.}
    \label{fig:dfam_parts}
\end{figure*}

\begin{figure}
    \centering
    \includegraphics[width=0.8\linewidth]{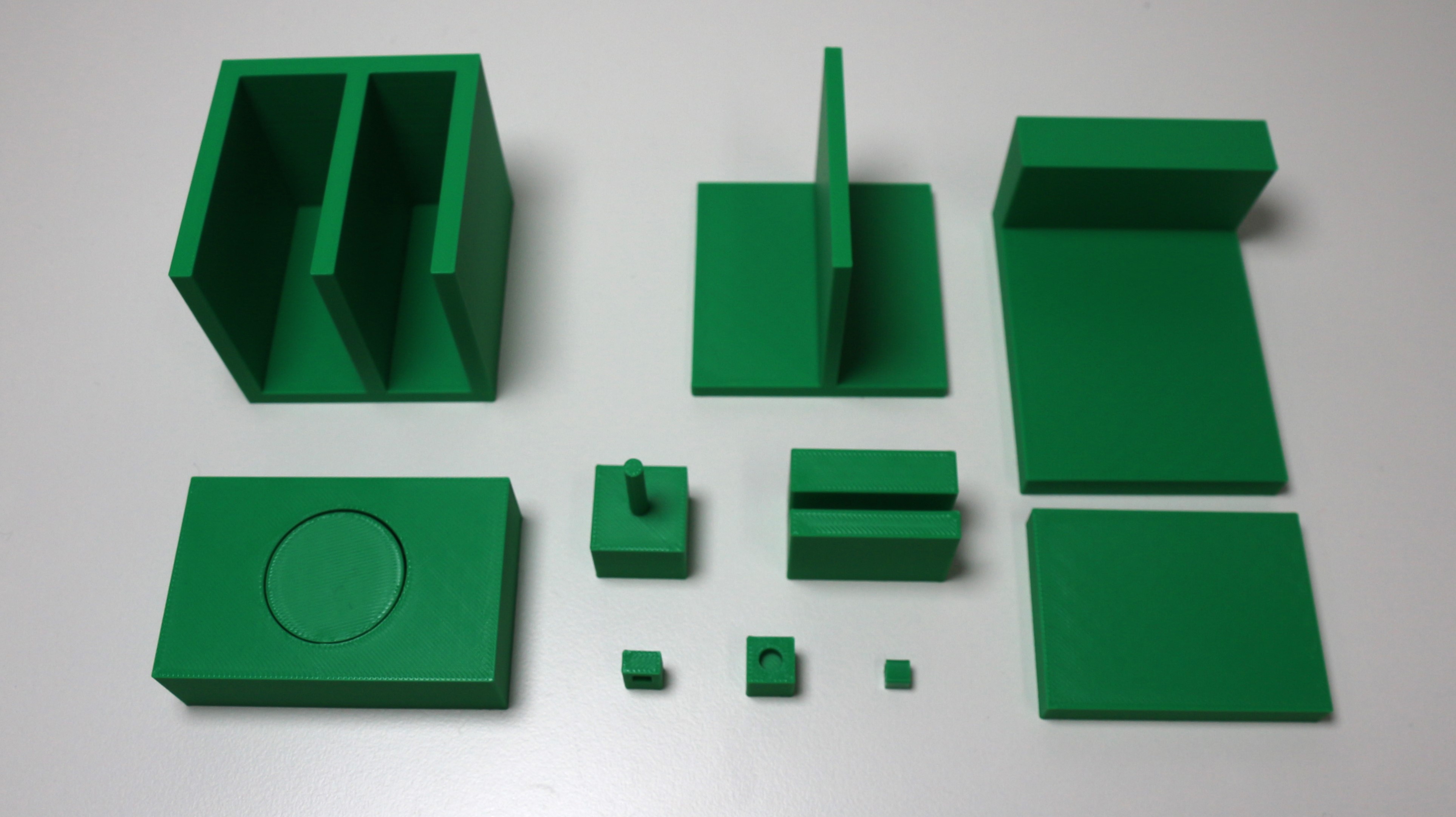}
    \caption{Parts from Fig.~\ref{fig:dfam_parts} (K--T) that we 3D-printed using a Carbon X1 Bambu printer.}
    \label{fig:printed_parts}
\end{figure}

\paragraph{Methodology} We assess GPT-4V's ability to understand and apply AM design rules by asking the model to predict success in 3D-printing various designs. For this task, we created a set of 20 designs, split into two sets: one set of problematic designs and another set of manufacturable designs, see Fig.~\ref{fig:dfam_parts}. For each of the ten design rules in The Hubs chart that pertain to FDM manufacturing, we created the ten problematic designs, each of which violates one of the ten rules. The other ten designs comprising the manufacturable designs set are \textit{similar} to problematic designs but actually pass all of the FDM rules in The Hubs chart. To confirm the intended manufacturability of the ten designs in the manufacturable set, we 3D-printed them using a Carbon X1 Bambu printer. All ten designs were printed successfully as shown in Fig.~\ref{fig:printed_parts}. 

We carried out 20 queries, each in a new context window and each corresponding with one of the 20 designs. For each query, we provided GPT-4V with the chart of the design rules and a dimensioned image---one of the twenty images shown in Fig.~\ref{fig:dfam_parts}---of the design we desired to print. We then asked GPT-4V, based on the provided design rules, to predict the success of 3D printing the part using an FDM printer. We asked the model to point to the specific design rule(s) violated if it believed the part would not print successfully. Sample queries can be seen in Fig.~\ref{dfm-additive}. To check repeatability, each of these queries was repeated three times for a total of 60 queries. We scored each response as follows:

\begin{enumerate}
    \item \textbf{Manufacturable? (max score 1):} If GPT-4V correctly answered if the part was manufacturable or not, we assigned a score of 1, otherwise, 0.
    \item \textbf{Correct Rule (max score 1):} This scoring metric is only applicable to designs in the problematic design set. If the rule that was in violation was named in GPT-4V's response, we assigned a score of 1, otherwise, 0.
    \item \textbf{\# Incorrect Rules (max score 0):} This scoring metric is only applicable to designs in the problematic design set. The number of rules GPT-4V mentioned that the model believed were violated, but which were truly not violated, corresponds with the negative value of this score. For example, if three rules were mentioned by GPT-4V which were not violated by the design, the score for this metric would be -3.
\end{enumerate}
A summary of all our results can be seen in Table \ref{tab:additive_table}.

\begin{figure}
\begin{contextbox}[]{}
\begin{promptbox}{}[When provided with Design C:]
Here is a chart of 3D printing design rules. We will refer to the rules in the chart by the column headers, so we will call the rules 1) support walls, 2) unsupported walls, 3) support \& overhangs, 4) embossed \& engraved details, 5) horizontal bridges, 6) holes, 7) connecting/moving parts, 8) escape holes, 9) minimum features, 10) pin diameter, and 11) tolerance. I've also attached the design of a part I plan to 3D print using an FDM printer. I do not intend to use support material. The face indicated in green will be in contact with the print bed. Assume no strict tolerance requirements unless they are indicated on the dimensions of the part. Also assume that the part does not need to mate with other parts. Based on the design rules, will the print be successful? Respond with only a simple "yes" or "no." If the answer is no, please concisely list the specific rule number(s) that is/are being violated. 

\vspace{10pt}

\includegraphics[height=5.5cm]{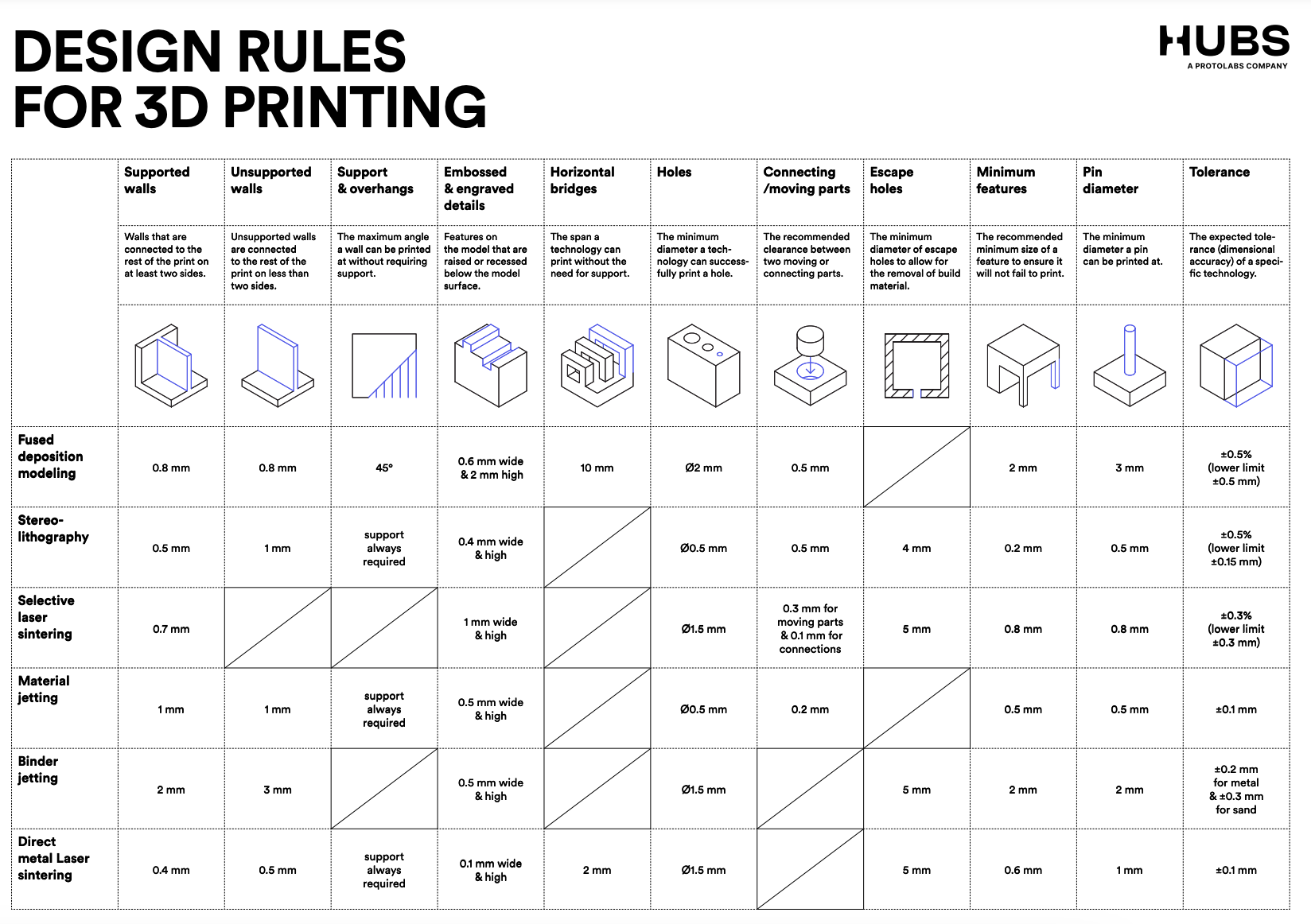}

Above image is reproduced here as a courtesy of HUBS \citep{hubs_3dprinting}.

\begin{minipage}[t]{0.5\linewidth}
Design C:

    \includegraphics[height=2.1cm]{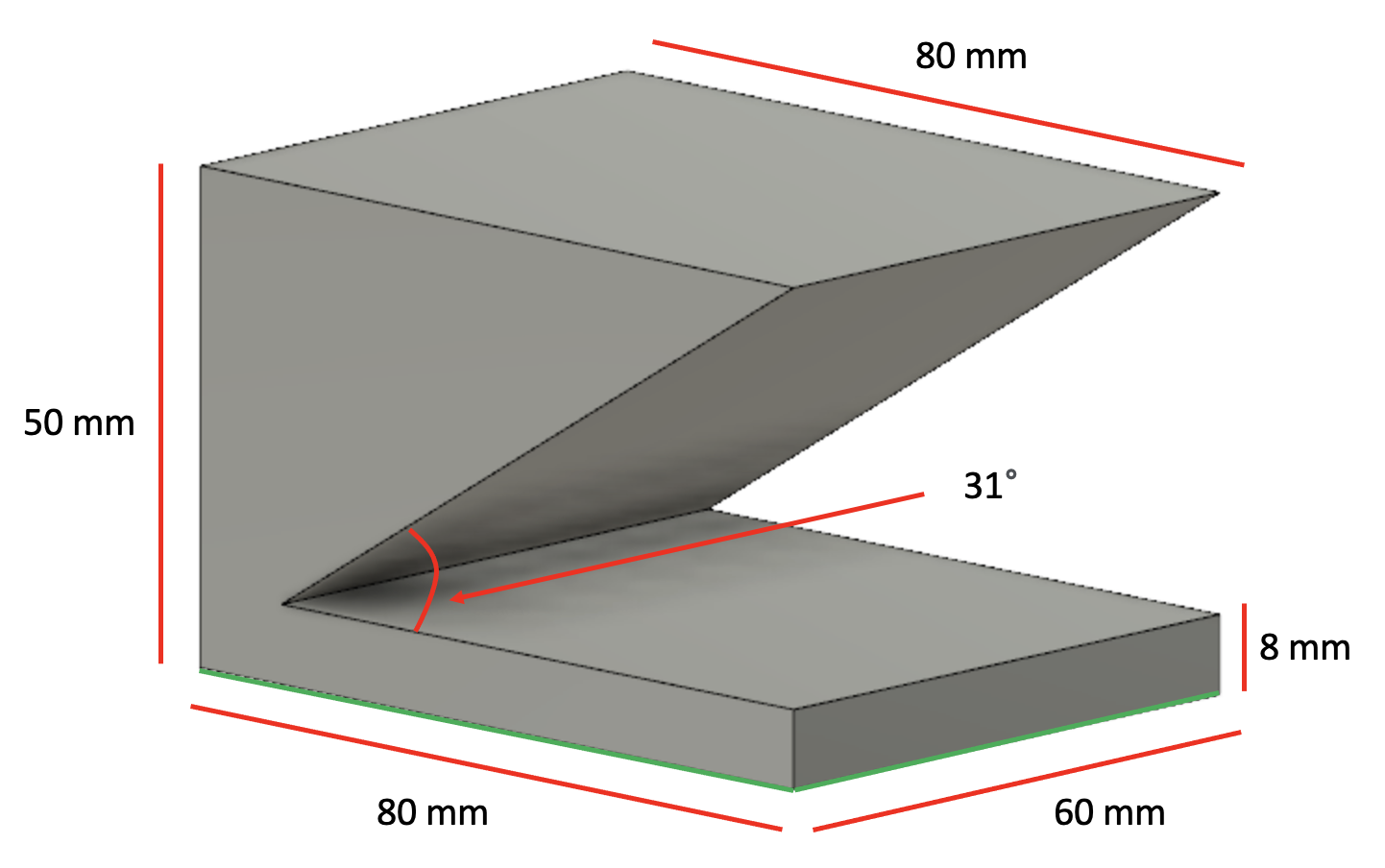} 
\end{minipage}%
\begin{minipage}[t]{0.5\linewidth}
Design M:

    \includegraphics[height=2.2cm]{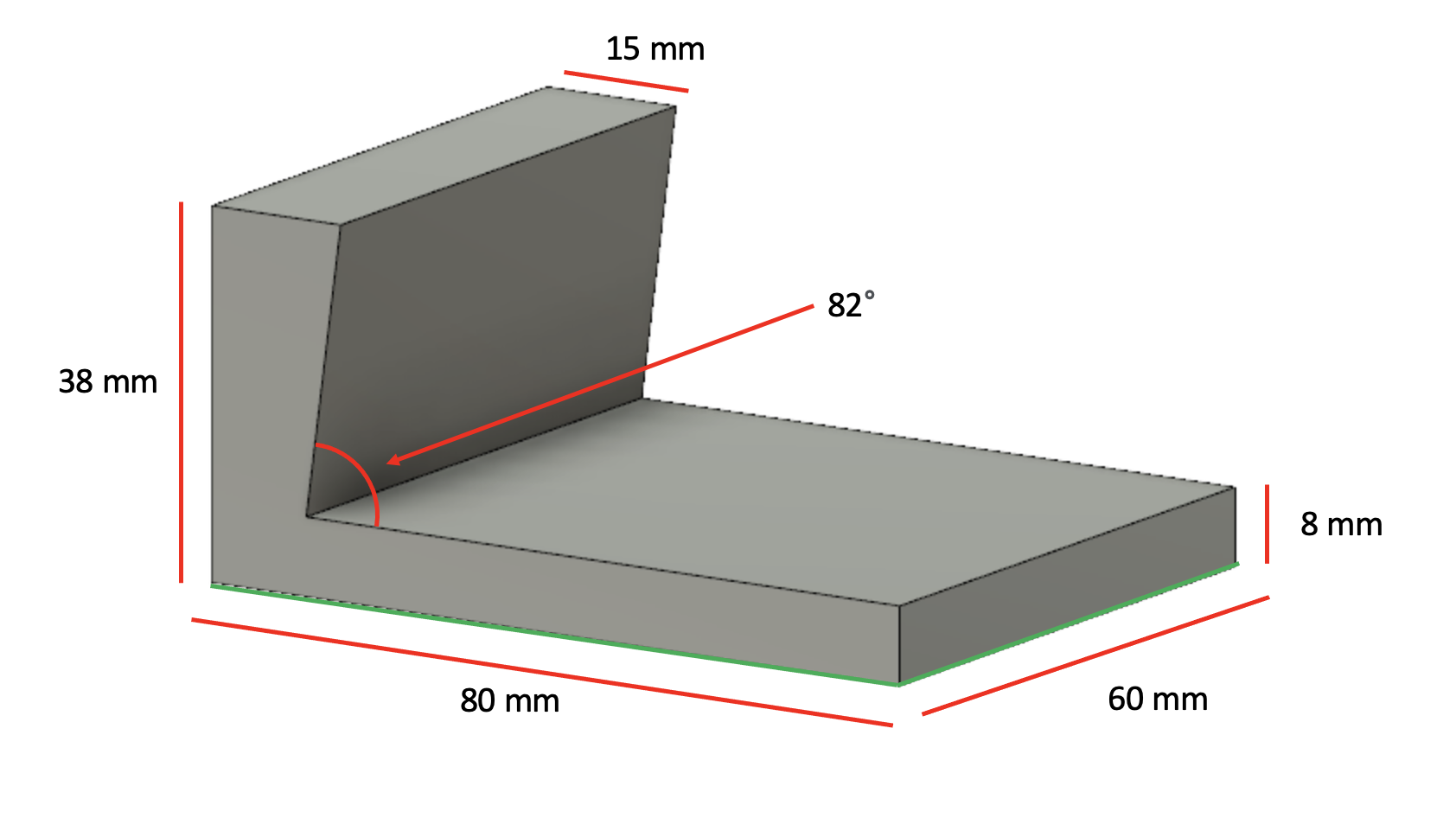} 
\end{minipage}
\tcblower

\hl{No.} Violated rules: \hl{3) Support \& overhangs} \hle{1. Horizontal bridges}

\vspace{6pt}
\parbox[][10pt][t]{\linewidth}{\textit{\textbf{When provided with Design M:}}}
\hle{No. Violates rule 3 (support \& overhangs), rule 5 (horizontal bridges), and rule 6 (holes).}

\end{promptbox}
\end{contextbox}%

\caption{Example prompts and answers for the design for additive manufacturing.}
\label{dfm-additive}
\end{figure}

\begin{table*}[h]
    \centering
    \caption{Scores achieved by GPT-4V on the design for additive manufacturing experiments across three trials.}
    \label{tab:additive_table}
    \begin{tabularx}{1\textwidth}{@{}l *{12}{Y}@{}} \toprule
         \textbf{Design \#} & \multicolumn{3}{c}{\textbf{Manufacturable?}} & \multicolumn{3}{c}{\textbf{Correct Rule}} & \multicolumn{3}{c}{\textbf{\# Incorrect Rules}} & \multicolumn{3}{c}{\textbf{Score}} \\
         \cmidrule(lr){2-4} \cmidrule(lr){5-7} \cmidrule(lr){8-10} \cmidrule(l){11-13}
         \textbf{Trial \#} &  1 & 2 & 3 & 1 & 2 & 3 & 1 & 2 & 3 & 1 & 2 & 3\\
         \midrule
         Design A   & 1 & 1 & 1 & 0 & 0 & 0 & -4 & -3 & -5 & -3 & -2 & -4 \\
         Design B   & 1 & 1 & 1 & 1 & 0 & 1 & -1 & -2 & -2 & 1  & -2 & 0  \\
         Design C   & 1 & 1 & 1 & 1 & 1 & 1 & -2 & -1 & -3 & 0  & 1  & -1 \\
         Design D   & 1 & 1 & 1 & 0 & 0 & 0 & -3 & -3 & -3 & -2 & -2 & -2 \\
         Design E   & 1 & 1 & 1 & 1 & 1 & 1 & -1 & -1 & -2 & 1  & 1  & 0  \\
         Design F   & 1 & 1 & 1 & 1 & 1 & 1 & -1 & -1 & -1 & 1  & 1  & 1 \\
         Design G   & 1 & 1 & 1 & 1 & 1 & 1 & -2 & -4 & -2 & 0  & -2 & 0  \\
         Design H   & 1 & 1 & 1 & 0 & 0 & 0 & -2 & -1 & -3 & -1 & 0  & -2 \\
         Design I   & 1 & 1 & 1 & 0 & 0 & 0 & -2 & -3 & -2 & -1 & -2 & -1 \\
         Design J   & 1 & 1 & 1 & 0 & 0 & 0 & -1 & -2 & -2 & 0  & -1 & -1 \\
         \bottomrule
         \multicolumn{13}{@{}p{\textwidth}@{}}{{\footnotesize Scores for Designs K-T are not shown in the above table, since GPT-4V always (across all three trials) incorrectly predicted those designs to be not manufacturable, even though they came from the manufacturable set. As such, the \textbf{Manufacturable?} score for Designs K-T for all trials is 0.}}\\
    \end{tabularx}
    
\end{table*}

\subsubsection{Design for Subtractive Manfuacturing}\label{mfg:subtractive}
Subtractive manufacturing is the most widely used manufacturing technology in the industries for manufacturing complex parts. This design process requires careful attention to the manufacturability of a part and typically this process is iterative. This is particularly challenging for parts with interacting features \citep{gao1998automatic}. Unfortunately, there is a very limited number of datasets for this task in the literature. Recently, deep learning-based approaches have been implemented to identify machining features using synthetic CAD datasets \citep{cao2020graph, zhang2018featurenet}. These datasets are created using a curated set of design principles. To this end, we utilize the MFCAD dataset \citep{cao2020graph} to query GPT-4V for manufacturing feature recognition from the image of a CAD model. 

\begin{figure}
\begin{contextbox}[label=dfm-subtractive1]{Design for Manufacturing: subtractive}
\begin{promptbox}{(Identifying machining features from an image of a CAD design)}[When provided with example 1:]
Here are the machining features, from an image you need to identify which machining features are present in the stock of material in the image. \\ 
----- \\ 
List of machining features: 
Rectangular through slot, Triangular through slot, Rectangular passage, Triangular passage, 6 sided passage, Rectangular through step, 2 sided through step, Slanted through step, Rectangular blind step, Triangular blind step, Rectangular blind slot, Rectangular pocket, Triangular pocket, 6 sided pocket, Chamfer, Stock\\

\begin{minipage}[t]{0.5\linewidth}
    Example 1:

    \hspace{.15\linewidth}\includegraphics[width=.7\linewidth]{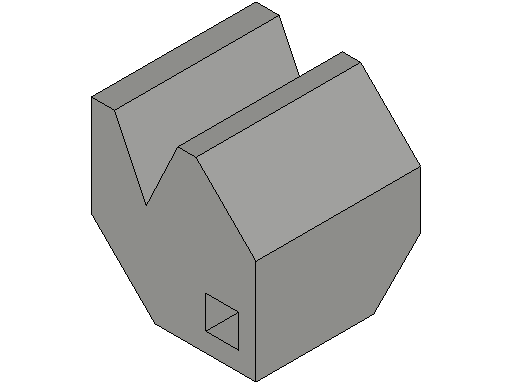}    
\end{minipage}%
\begin{minipage}[t]{0.5\linewidth}
    Example 2:

    \hspace{.15\linewidth}\includegraphics[width=.7\linewidth]{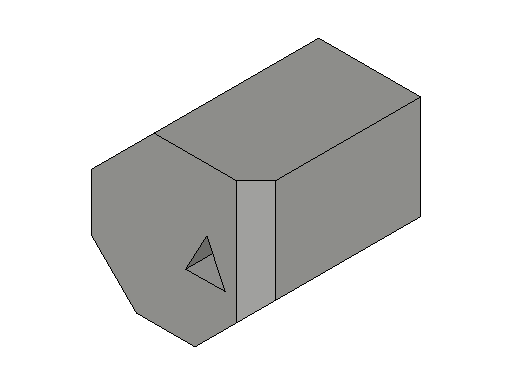}   
\end{minipage}%
\tcblower

The image shows a stock material with these features:

\begin{enumerate}[leftmargin=*,labelindent=0mm]
    \item \hle{Slanted through step: A slanted cut that goes through the width of the stock is present.}
    \item \hl{Rectangular through slot: On the side of the stock, there is a rectangular slot that appears to pass all the way through.}
\end{enumerate}

The combination of these features indicates a detailed machining process, providing both angled surfaces and through cuts. If further analysis is needed or you have more images to consider, please continue to provide them.

\vspace{6pt}
\parbox[][10pt][t]{\linewidth}{\textit{\textbf{When provided with example 2:}}}

Based on the image provided, here are the machining features that can be identified:

\begin{enumerate}[leftmargin=*,labelindent=0mm]
    \item \hl{Triangular through slot: Visible on one side, it's a triangle-shaped slot that goes completely through the material.}
    \item \hle{Rectangular passage: There appears to be a rectangular feature that could be a passage; it's visible on the front face but it's not clear if it goes all the way through.}
    \item \hle{2 sided through step: The bottom portion of the stock has a step that has two sides visible, indicating it is a through feature.}
    \item \hl{Stock: The overall piece of material itself is considered the stock.}
\end{enumerate}

It is not possible to confirm the presence of some features, like pockets or blind steps, without additional views. If there are other angles or cross-sections available, that would help in identifying more features accurately.
\end{promptbox}
\end{contextbox}

\caption{Example prompts and answers for the feature identification in subtractive manufactring.}
\label{dfm-subtractive}
\end{figure}

\paragraph{Methodology} We perform a quantitative study based on multiple queries to GPT-4V. We randomly pick $50$ samples from the MFCAD dataset and create the images of each CAD model. Each of these images corresponds to a ground truth that assigns machining features to each surface of the CAD model. In general, there are $15$ possible machining features that we test in all of the experiments which excludes the stock material block. The list of machining features is the following: rectangular through slot, triangular through slot, rectangular passage, triangular passage, 6 sided passage, rectangular through step, 2 sided through step, slanted through step, rectangular blind step, triangular blind step, rectangular blind slot, rectangular pocket, triangular pocket, 6 sided pocket, chamfer. We query GPT-4V with each of these images and ask for the machining features that are present in the design. First, we provide an initial prompt to focus on design for manufacturing. Next, we query GPT-4V about each of the images, as shown in Fig.~\ref{fig:dfm2}, sequentially. Figure~\ref{dfm-subtractive} shows two example prompts and the corresponding responses from GPT-4V. 

 The dataset used in this experiment is provided with this document as an open-source small-scale evaluation dataset for vision-language LLMs. This design for the subtractive manufacturing dataset is mainly based on the MFCAD dataset~\citep{cao2020graph} and consists of $15,488$ images of CAD models and their corresponding machining features as labels. We only show results for $8$ image-label pairs in this document as the inaccuracy of GPT-4V makes it difficult to quantify the performance. We anticipate that this dataset can be useful in evaluating a more capable vision-language model that can understand 3D geometry and engineering design images.

\begin{itemize}
\item Initial prompt: \textit{I am going to ask you a series of questions about some machining feature recognition from an image of a stock of material.} 
\item Image prompt: \textit{Here are the machining features, from an image you need to identify which machining features are present in the stock of material in the image.
-----
List of machining features: 
Rectangular through slot, Triangular through slot, Rectangular passage, Triangular passage, 6 sided passage, Rectangular through step, 2 sided through step, Slanted through step, Rectangular blind step, Triangular blind step, Rectangular blind slot, Rectangular pocket, Triangular pocket, 6 sided pocket, Chamfer, Stock}
\end{itemize}

We repeat these $50$ queries three times and obtain similar responses. GPT-4V identifies at least one feature in most images but fails to consistently identify features.  

\paragraph{Discussion: Design For Manufacturing}
Overall, we note that GPT-4V never successfully answers all parts of any of our DfM queries. While the model is able to correctly answer pieces of our questions (e.g. that the design breaks an AM rule, that the design contains a certain machining feature), its answers are never fully accurate. In particular, we notice that GPT-4V sometimes struggles to or forgets to follow directions specified in a prompt, and its performance deteriorates as the complexity of designs increases.

In terms of its ability to predict the success of additive manufacturability based on provided design rules, GPT-4V \textit{always} states that the provided design will not be able to be successfully produced using AM. The response that the part would not be able to be 3D-printed successfully was consistent across all 60 queries, for both the problematic and manufacturable design sets. In other words, for all the designs that were manufacturable---and which violated no FDM design rules in the chart---GPT-4V hypothesized that they would break one of the 3D printing design rules. This consistently negative response to printability likely reflects a cautious posture on the part of the model. We also observed from the data in Table~\ref{tab:additive_table} that GPT-4V always maintains that multiple design rules are broken, while all designs in the problematic design set violate just one of the rules listed in the Hub design rule chart. As such, GPT-4V is never fully correct in answering any of our questions about additive manufacturability based on the provided design rules. GPT-4V's listing of many rules in response to our question about rules violated could further reflect its tendency to take on a cautious position. Less than half the time (13/30 queries) is GPT-4V able to correctly identify the violated rule for the problematic designs. We also note that the model sometimes seems to get confused by and/or forgets our ask to name the rules by the numbers we assigned to each one in our prompt (see Fig.~\ref{dfm-additive}). As seen in Fig.~\ref{dfm-additive}, the model lists the first rule by the correct number but then lists the second rule by an incorrect number. 

Overall, in the context of additive manufacturing tasks, the implications of using GPT-4V are nuanced. The model consistently predicts that designs will not be successfully produced using additive manufacturing methods, regardless of their actual manufacturability. This uniform negativity indicates a cautious approach, likely to avoid over-optimistic assessments, but also leads to an overestimation of manufacturing challenges. GPT-4V’s tendency to list multiple broken design rules, even when only one is violated, further reflects its caution. However, this approach can be misleading in real-world scenarios where precise and accurate manufacturability assessments are crucial. The model’s difficulty in following specific prompt instructions, such as correctly identifying rules by assigned numbers, points to a need for further development in its ability to process and respond to detailed additive manufacturing queries. While GPT-4V’s partial answering capability suggests a basic understanding of additive manufacturing principles, its current limitations underscore that it is not yet a reliable tool for comprehensive and accurate manufacturability assessments in additive processes.

\newlength{\myimagewidth}
\setlength{\myimagewidth}{2.5cm} %

\newlength{\myitemsep}
\setlength{\myitemsep}{-0.25em} %

\begin{figure*}
\centering
\begin{subfigure}{1\textwidth}
\centering
\begin{tabular}{@{}p{0.75cm} p{3.5cm} p{3.5cm} p{3.5cm} p{3.5cm}@{}}
 & \includegraphics[width=\myimagewidth]{0-0-0-0-1-23.png} & \includegraphics[width=\myimagewidth]{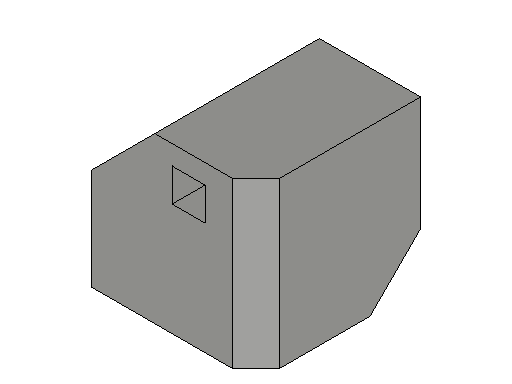} & \includegraphics[width=\myimagewidth]{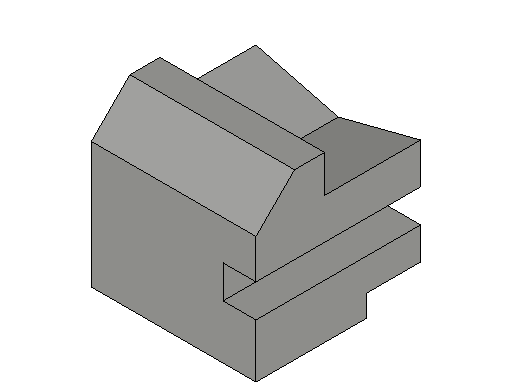} & \includegraphics[width=\myimagewidth]{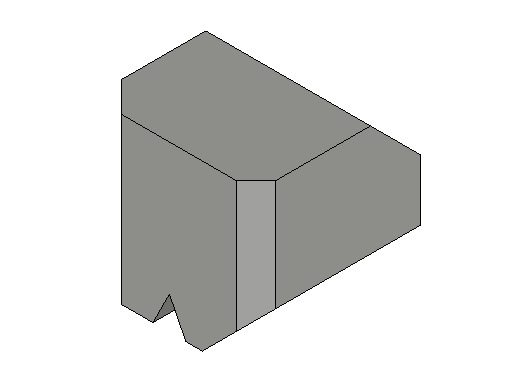}\\
\specialrule{\lightrulewidth}{0.2ex}{0.4ex}
GPT-4V 
& 
\begin{minipage}[t]{0.35\textwidth} \footnotesize{\begin{itemize}[leftmargin=*,itemsep=\myitemsep]%
\item Triangular through slot \item Rectangular passage \item 2 sided through step\end{itemize}} \end{minipage} 
& 
\footnotesize{\begin{minipage}[t]{0.35\textwidth} \begin{itemize}[leftmargin=*,itemsep=\myitemsep]%
\item Triangular through slot \item Rectangular passage
\end{itemize} \end{minipage}} 
& 
\footnotesize{\begin{minipage}[t]{0.35\textwidth} \begin{itemize}[leftmargin=*,itemsep=\myitemsep]%
\item Slanted through step \item 2 sided through step
\end{itemize} \end{minipage}} 
& 
\footnotesize{\begin{minipage}[t]{0.35\textwidth} \begin{itemize}[leftmargin=*,itemsep=\myitemsep]%
\item Rectangular through step \item 2 sided through step
\end{itemize} \end{minipage}} \\

\specialrule{\lightrulewidth}{0.2ex}{0.4ex}
GT
& 
\begin{minipage}[t]{0.35\textwidth} \footnotesize{\begin{itemize}[leftmargin=*,itemsep=0pt] \item Triangular through slot \item Rectangular through slot \end{itemize}} \end{minipage} 
& 
\footnotesize{\begin{minipage}[t]{0.35\textwidth} \begin{itemize}[leftmargin=*] \setlength{\itemsep}{\myitemsep} \item Rectangular through slot \item Rectangular passage \end{itemize} \end{minipage}} 
& 
\footnotesize{\begin{minipage}[t]{0.35\textwidth} \begin{itemize}[leftmargin=*]\setlength{\itemsep}{\myitemsep}   \item Rectangular through slot \item Slanted through step \item Rectangular through step \end{itemize} \end{minipage}} 
& 
\footnotesize{\begin{minipage}[t]{0.35\textwidth} \begin{itemize}[leftmargin=*,itemsep=\myitemsep]  \item Rectangular through slot \item 6 sided passage \end{itemize} \end{minipage}} \\
\specialrule{\lightrulewidth}{0.2ex}{0.4ex}
\end{tabular}
\end{subfigure}

\begin{subfigure}{\textwidth}
\centering
\begin{tabular}{@{}p{0.75cm} p{3.5cm} p{3.5cm} p{3.5cm} p{3.5cm}@{}}

 & \includegraphics[width=\myimagewidth]{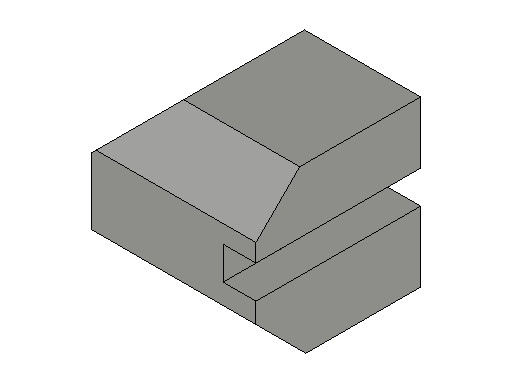} & \includegraphics[width=\myimagewidth]{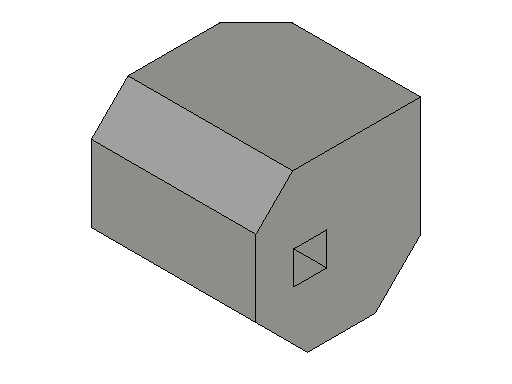} & \includegraphics[width=\myimagewidth]{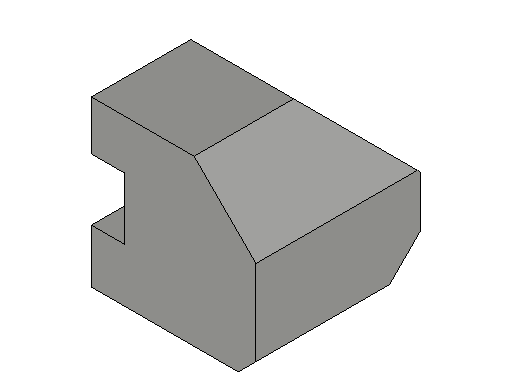} & \includegraphics[width=\myimagewidth]{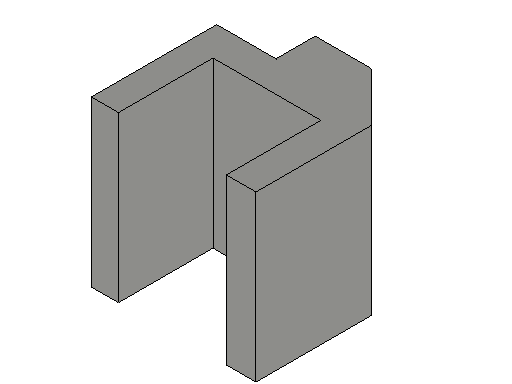}\\
\specialrule{\lightrulewidth}{0.2ex}{0.4ex}
GPT-4V 
& 
\begin{minipage}[t]{0.35\textwidth} \footnotesize{\begin{itemize}[leftmargin=*] \setlength{\itemsep}{\myitemsep}
\item Rectangular through step \item Slanted through step
\end{itemize}} \end{minipage} 
& 
\footnotesize{\begin{minipage}[t]{0.35\textwidth} \begin{itemize}[leftmargin=*] \setlength{\itemsep}{\myitemsep}
\item Slanted through step \item Rectangular pocket
\end{itemize} \end{minipage}} 
& 
\footnotesize{\begin{minipage}[t]{0.35\textwidth} \begin{itemize}[leftmargin=*]\setlength{\itemsep}{\myitemsep}
\item Slanted through step
\item Rectangular through step
\end{itemize}
\end{minipage}} 
& 
\footnotesize{\begin{minipage}[t]{0.35\textwidth} \begin{itemize}[leftmargin=*] \setlength{\itemsep}{\myitemsep}
\item Rectangular through step \item Rectangular passage
\end{itemize} \end{minipage}} \\

\specialrule{\lightrulewidth}{0.2ex}{0.4ex}
GT
& 
\begin{minipage}[t]{0.35\textwidth} \footnotesize{\begin{itemize}[leftmargin=*] \setlength{\itemsep}{\myitemsep} \item Rectangular through slot \item Rectangular through step \end{itemize}} \end{minipage} 
& 
\footnotesize{\begin{minipage}[t]{0.35\textwidth} \begin{itemize}[leftmargin=*] \setlength{\itemsep}{\myitemsep} \item Rectangular blind slot \item Rectangular through slot \end{itemize} \end{minipage}} 
& 
\footnotesize{\begin{minipage}[t]{0.35\textwidth} \begin{itemize}[leftmargin=*]\setlength{\itemsep}{\myitemsep} \item Rectangular through slot \item Triangular pocket \end{itemize} \end{minipage}} 
& 
\footnotesize{\begin{minipage}[t]{0.35\textwidth} \begin{itemize}[leftmargin=*] \setlength{\itemsep}{\myitemsep} \item Rectangular through step \item Slanted through step \item Rectangular through slot \end{itemize} \end{minipage}} \\
\specialrule{\lightrulewidth}{0.2ex}{0.4ex}
\end{tabular}
\end{subfigure}

\caption{Machining feature recognition from CAD images: results are shown for eight selected samples where each ground 
truth (GT) is also shown corresponding to the GPT-4V response.}
\label{fig:dfm2}
\end{figure*}

In subtractive manufacturing tasks, GPT-4V demonstrates an ability to identify at least one machining feature in most images (12/20), but its performance is inconsistent, particularly with more complex designs. For example, GPT-4V often identifies `triangular through slot' instead of `rectangular through slot' and `2-sided through step' instead of `6-sided passage'. The model also misidentifies distinct features, such as confusing a 'triangular through slot' with a 'rectangular through slot', and exhibits challenges in understanding more intricate geometric features. This inconsistency in feature identification can lead to unreliable assessments in scenarios where precision in subtractive manufacturing is essential. While GPT-4V seems to fare better with simpler geometric objects, its difficulty with complex objects suggests that its current use might be more suitable for preliminary assessments or educational purposes, rather than for detailed, technical manufacturing evaluations. The somewhat random nature of its explanations and inability to satisfy detailed engineering design concerns indicates that significant improvements are necessary before GPT-4V can function as a stand-alone tool in subtractive manufacturing tasks. As such, while GPT-4V can provide some support in these tasks, it requires careful human oversight and verification to ensure accuracy and relevance in practical manufacturing scenarios.

Based on the study's insights into GPT-4V's performance in Design for Manufacturing tasks, future work should focus on enhancing the model's precision and depth of understanding in both additive and subtractive manufacturing processes. For additive manufacturing, research should aim to calibrate GPT-4V's cautious approach, enabling it to differentiate between manufacturable and non-manufacturable designs more accurately, and follow specific guidelines more precisely. In the realm of subtractive manufacturing, efforts need to be directed toward improving GPT-4V's ability to consistently and correctly identify complex machining features. This includes training the model to handle a broader range of geometries and intricate design elements, thus reducing its current limitations in assessing detailed and technical aspects of manufacturing designs. Additionally, developing a better way for AI models to understand 3D geometry could enhance GPT-4V's interpretative capabilities, leading to more reliable and practical applications in the manufacturing sector. These advancements would not only make GPT-4V a more robust tool for manufacturing design but also pave the way for its broader application in automated manufacturing processes.
In the next section, we turn to another application of GPT-4V for manufacturability: post-manufacturing inspection.

\subsection{Post-Manufacturing Inspection}\label{mfg:inspection}
Engineering inspection constitutes a whole domain within itself: parts must be inspected after they are fabricated to ensure that they meet certain technical requirements, and for critical components, inspection can continue into the lifetime of the part. Inspection is a key aspect of the engineering design process, as it may help in improving the next iteration of the design.
Oftentimes, inspection necessitates a visual component (e.g., detection of a defect through an image, X-ray, graph of collected data, etc.) alongside extensive engineering knowledge of detailed documents, like engineering standards. As such, we are curious to understand if GPT-4V, with multimodal capabilities, can aid engineers with defect detection in images. 

\begin{figure}[h]
    \centering
    \includegraphics[width=1.0\linewidth]{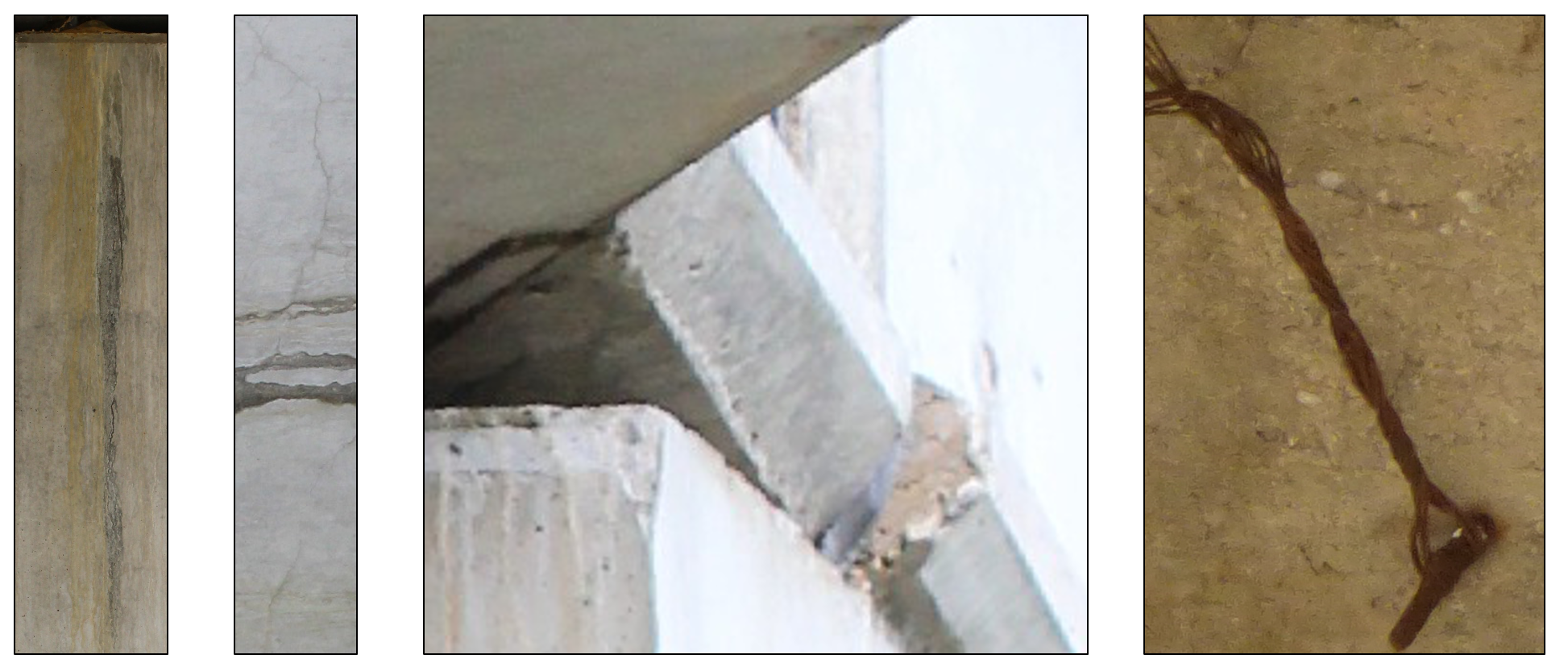}
    \caption{Bridge structural concrete images adapted from the CODEBRIM dataset \citep{mundt2019meta} under its specific license. From left to right, as named in the original dataset: 1) \textit{image\_0000005\_crop\_0000001.png} - contains efflorescence and corrosion stain defects. 2) \textit{image\_0000046\_crop\_0000001.png} - contains crack defects. 3) \textit{image\_0000109\_crop\_0000003.png} - contains spallations and corrosion strain defects. 4) image \textit{image\_0001189\_crop\_0000004.png} - contains exposed bar defects.}
    \label{fig:concrete_examples}
\end{figure}

\begin{figure}
\begin{contextbox}[]{Concrete Defect Classification}

\begin{promptbox}{Classifying structural concrete defects, image\_0000159\_crop\_0000002.png from \citep{mundt2019meta}. The correct answer is B) spallations.}[When provided with example 1:]
This photo was taken of structural concrete in a bridge. You are assessing the bridge’s structural safety, and you are looking for defects in the concrete. Defects that could be in the concrete are A) cracks, B) spallations, C) efflorescence, D) exposed bars, E) corrosion stains. List which defects – by their letters – you see in the photo. The image could contain multiple defects, one defect, or no defects.\\

\begin{minipage}{0.5\linewidth}
Example 1:

\centering
    \includegraphics[height=0.2\textheight]{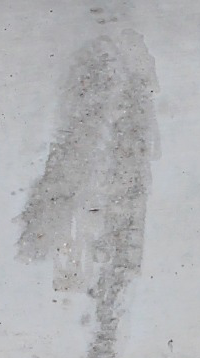}    
\end{minipage}%
\begin{minipage}{0.5\linewidth}
Example 2:

\centering
    \includegraphics[height=0.2\textheight]{image\_0000057\_crop\_0000004.png}    
\end{minipage}

{\tiny Images reproduced from~\citep{mundt2019meta} under their specific license.}
\tcblower

Based on the provided image, I can identify the following defects in the structural concrete:

\begin{itemize}[topsep=1pt]
    \item \hle{A) cracks}
    \item \hle{C) efflorescence}
\end{itemize}

Please consult with a professional structural engineer for a comprehensive assessment of the bridge's structural safety.

\vspace{6pt}
\parbox[][10pt][t]{\linewidth}{\textit{\textbf{When provided with example 2:}}}
The image provided shows a close-up view of a concrete surface. From what is visible in this image, \hl{I cannot clearly identify any of the listed defects such as cracks (A), spallations (B), efflorescence (C), exposed bars (D), or corrosion stains (E)}. The image resolution or the perspective may not be sufficient to detect such defects accurately. It's important to have a clear and properly focused image, ideally with some context for scale and location, to perform an accurate assessment of concrete for potential structural defects.

\end{promptbox}
\end{contextbox}

\caption{Example prompts and answers for the concrete defect classification.}
\label{inspection_concrete}
\end{figure}

\paragraph{Methodology} For our analysis, we use the CODEBRIM (Concrete DEfect BRidge IMage) dataset, released by \citep{mundt2019meta}. It entails images of structural concrete from bridges that contain none or some of the following defects: cracks, spallations, efflorescence, exposed bars, and corrosion strains. Sample images from the dataset can be seen in Fig.~\ref{fig:concrete_examples}. We chose a subset of 23 images from the CODEBRIM dataset for our experiments. The images were chosen such that each of the five defect types was present in at least five images. Five of the 23 images were ``background'' images, containing no defects.   We provided GPT-4V each image in a separate context window\footnote{This experiment was conducted using the Nov. 6, 2023 version through the API interface.} and asked the model to identify any of the five defects it could find. If the model was hesitant to respond---due to image resolution or safety concerns---but still suggested certain defects, we counted that as a response. To understand repeatability, each of the 23 image experiments was repeated three times, for a total of 69 queries. Two queries and responses can be seen in Fig.~\ref{inspection_concrete}. The results for all experiments can be seen in Tables \ref{tab:defect_crack}--\ref{tab:defect_corrosion_stains}.

\begin{table}[ht!]
\centering
\caption{Confusion matrix for crack defects as predicted by GPT-4V. $F_1$ score is 0.45.}
\label{tab:defect_crack}
\begin{tabular}{@{}l  c  c@{}}
\toprule
& True Positive & True Negative \\
\midrule
Predicted Positive & 11 & 24 \\
Predicted Negative & 3 & 19 \\
\bottomrule
\end{tabular}
\end{table}

\begin{table}[ht!]
\centering
\caption{Confusion matrix for spallation defects as predicted by GPT-4V. $F_1$ score is 0.67.}
\label{tab:defect_spallation}
\begin{tabular}{@{}l  c c@{}}
\toprule
& True Positive & True Negative \\
\midrule
Predicted Positive & 11 & 3 \\
Predicted Negative & 8 & 35 \\
\bottomrule
\end{tabular}
\end{table}

\begin{table}[ht!]
\centering
\caption{Confusion matrix for efflorescence defects as predicted by GPT-4V. $F_1$ score is 0.27.}
\label{tab:defect_efflorescence}
\begin{tabular}{@{}l  c c@{}}
\toprule
& True Positive & True Negative \\
\midrule
Predicted Positive & 3 & 4 \\
Predicted Negative & 12 & 38 \\
\bottomrule
\end{tabular}
\end{table}

\begin{table}[ht!]
\centering
\caption{Confusion matrix for exposed bar defects as predicted by GPT-4V. $F_1$ score is 0.62.}
\label{tab:defect_exposed_bar}
\begin{tabular}{@{}l  c c@{}}
\toprule
& True Positive & True Negative \\
\midrule
Predicted Positive & 9 & 5 \\
Predicted Negative & 6 & 37 \\
\bottomrule
\end{tabular}
\end{table}

\begin{table}[ht!]
\centering
\caption{Confusion matrix for corrosion stain defects as predicted by GPT-4V. $F_1$ score is 0.65.}
\label{tab:defect_corrosion_stains}
\begin{tabular}{@{}l  c c@{}}
\toprule
& True Positive & True Negative \\
\midrule
Predicted Positive & 15 & 10 \\
Predicted Negative & 6 & 26 \\
\bottomrule
\end{tabular}
\end{table}

\paragraph{Discussion} 
The confusion matrices (Tables~\ref{tab:defect_crack}--\ref{tab:defect_corrosion_stains}) provided for different defect types in structural concrete offer insights into GPT-4V's defect detection capabilities. For 12 experiments (8 different images), GPT-4V would not provide an answer to our question, citing resolution issues, safety concerns, or just plainly stating that it could not assist with the request. When it did answer, GPT-4V did not perform particularly well in predicting types of defects, as seen by the relatively low F1 scores for each class of defect. We also note that GPT-4V tends to over-predict the crack defect; this is evidenced by the relatively high recall (true positive rate) score of 0.79 and a relatively low specificity (true negative rate) score of 0.44 when compared with the other classes. Perhaps most familiar with the crack defect, GPT-4V may over-predict cracks out of an abundance of caution and the safety implications of missing a true positive defect in concrete images of a bridge. For 14 out of the 69 queries, GPT-4V had a perfect defect prediction of all defect classes in the image. Intriguingly, half of these perfect scores were for images without any defects, suggesting that GPT-4V might be more adept at discerning the absence of defects rather than accurately classifying the type of defect present.

The findings of this study have significant implications for the application of AI in engineering inspection tasks. While GPT-4V demonstrates potential in identifying defects in structural concrete, its moderate performance underlines the need for further model refinement and continued reliance on human expertise or more specialized machine learning tools. The model's ability to discern the absence of defects could be leveraged in preliminary inspections to streamline processes, yet the necessity for human verification remains paramount, especially in safety-critical assessments. These results suggest avenues for future research focused on improving AI accuracy through diverse training datasets and approaches of expert feedback. 

\subsection{Summary}
To assess the manufacturing-related knowledge of GPT-4V, we performed three types of experiments. A concise summary of our findings for each are provided below.
\begin{enumerate}[(i)]
    \item \textbf{Design for Additive Manufacturing \ref{mfg:dfam}}
    In the realm of additive manufacturing, does GPT-4V consistently predict the 3D printability of a design based on a set of provided DfAM rules?
    \begin{itemize}
        \item GPT-4V uniformly (in all instances) indicated that designs would not be suitable for 3D printing. This conclusion was drawn irrespective of whether the designs actually conformed to the specified additive manufacturing rules.
    \end{itemize}

    \item \textbf{Design for Subtractive Manufacturing \ref{mfg:subtractive}}
    Is GPT-4V capable of identifying manufacturing features in subtractive manufacturing designs?
    \begin{itemize}
        \item GPT-4V exhibited a basic grasp of feature geometries but lacked consistency in its responses. The model struggled to differentiate between similar features and frequently resorted to making arbitrary guesses.
    \end{itemize}

    \item \textbf{ Post-Manufacturing Inspection \ref{mfg:inspection}} 
    To what extent can GPT-4V accurately classify different types of defects in images, specifically in the context of identifying concrete defects in manufacturing?

    \begin{itemize}
        \item Based on our experiments with concrete defect classification, we find that GPT-4V may have the potential to distinguish between images that have defects and images that do not. However, it was unable to consistently and accurately classify different types of concrete defects.
    \end{itemize}

\end{enumerate}

\section{Engineering Textbook Problems}\label{sec:education}
\paragraph{Overview and Motivation}
In this last section, we take a step back from the product development process and investigate GPT-4V's abilities to solve problems that are present in engineering curricula. During their curriculum, students are regularly asked to solve engineering design problems that require them to interpret sketches, graphs, tables, and images to answer a related question. As such, students need to integrate their natural language processing and visual information understanding skills with their domain knowledge to solve this type of problem. 
The underlying idea is that these are tasks and assignments used to evaluate humans' readiness to be engineers. Consequently, they may enable us to draw some comparison with GPT-4V's readiness to support engineering tasks. Textbook problems, exam questions, and standardized tests have been quite popular ways to evaluate LLMs~\citep{katz2023gpt,wang2023scibench}. These problems are often well-defined, self-contained, and mostly closed-form type questions~\citep{Taraban2011}, supporting replicability~\citep{ZONG2023100506}. For example, for text input only, SciBench~\citep{wang2023scibench} features 695 collegiate-level textbook problems drawn from physics, chemistry, and mathematics. Using this benchmark, SciBench~\citep{wang2023scibench} aimed at evaluating the reasoning and problem-solving skills of LLMs.

Following a similar approach, we propose to use engineering textbook problems requiring visual information to evaluate GPT-4V's understanding and problem-solving capabilities through the pairing of different visual and textual information.

\paragraph{Methodology}
We gathered questions from two undergraduate engineering design classes publicly available under CC-BY-NC-SA on MIT OpenCourseWare~\citep{OCW_2007,OCW_2008}. The class materials include problem sets and exams. All class materials come with model solutions, which we use as ground truth. To ensure that we are evaluating GPT-4V's multimodal capabilities, we select questions that reference one or more pictures in the question prompt. We ignore questions that require students to annotate an image as GPT-4V cannot generate images, but questions asking for sketches are included. Indeed, sketches can be parameterized and drawn using coding languages.

To ensure independence, we reset GPT-4V's context window for each question except for multi-part questions, where we prompt GPT-4V sequentially. For multi-part questions that have multiple images, we supplement each sub-question with only the images required to solve that particular question to avoid confusing GPT-4V with superfluous information. For example, consider a multi-part question has two images $X$ and $Y$, and part (a) only requires $X$ to solve, part (b) requires $Y$ to solve and part (c) requires both $X$ and $Y$ to solve, we would supplement question (a) with $X$, question (b) with $Y$ and question (c) with both $X$ and $Y$.

We evaluate GPT-4V's correctness based on a binary scale: one point is only given for fully correct answers, and none otherwise. Being ``fully correct'' means outputting an answer that is semantically similar to the ground truth for free-text questions. For questions involving calculations, the correct numerical answer must be provided and the intermediate steps should reasonably lead to the correct solution. For multi-part questions, we award a point for each correct part. We group the errors into three categories:

\begin{itemize}
    \item \textbf{Reasoning:} Incorrect explanation or calculation.
    \item \textbf{Inference:} Incorrect information extraction from the image.
    \item \textbf{Imprecise:} Vague answer or explanation without execution.
\end{itemize}

The questions were repeated three times to account for the variability of the model and, overall, a question was considered correctly answered if at least two repeats were correct.

\textit{Can GPT-4V Solve Engineering Textbook Problems?}
We extracted 21 questions from two classes, resulting in a total of 44 questions when counting each sub-part individually. We observe that GPT-4V can answer 16 of these 44 questions correctly, giving an average 36\% accuracy. An overview of all the repeats and questions is provided in Table~\ref{tab:edu_overall}. Relatively to the type of image, GPT-4V answered correctly most questions involving 3D models and tables (63\% and 67\% respectively) but had a lower success rate for photographs (33\%), diagrams (29\%) and graphs (0\%), see Table~\ref{tab:edu_sum_image}. In terms of question format, GPT-4V performed slightly better on free-text questions (44\%), than on any other format, see Table~\ref{tab:edu_sum_type}. Overall, we observe that GPT-4V makes mostly reasoning errors (20), followed by imprecise answers (5) and inference errors (3).
Thus, it seems to be more helpful for questions that require explanations and for problems that ask questions about tables or 3D models.

\begin{table}[ht]
    \centering
    \caption{Summary of GPT-4V's score on textbook problems grouped by type of image.}
    \label{tab:edu_sum_image}
    \begin{tabular}{@{}ccccccc@{}} 
    \toprule
                    & Photo & Diagram     & Graph & 3D & Table      & Overall  \\
      \midrule
      Correct & 1 & 8 & 0 & 5 & 2 & 16 \\
      Total         & 3          & 28          & 2     & 8        & 3          & 44         \\
     \midrule
     Avg. & 33\%     & 29\% & 0\%   &  63\%    & 67\% & 36\%\\ \bottomrule
    \end{tabular}
\end{table}

\begin{table}[ht]
    \centering
    \caption{Summary of GPT-4V's score on textbook problems grouped by question format.}
    \label{tab:edu_sum_type}
    \begin{tabular}{@{}ccccccc@{}} 
    \toprule
           & Free text & MCQ & Numerical  & Draw       & Overall     \\
          \midrule
          Correct  & 7           & 0   & 8          & 1 & 15        \\
          Total    & 16          & 1   & 24         & 3          & 44        \\
     \midrule
         Accuracy & 44\%  & 0\%   & 33\% & 33\% & 36\%\\ \bottomrule
    \end{tabular}
\end{table}

\begin{table}[t]
\caption{Detailed list of questions including image type and question format and answers by GPT-4V for each trial, along with the type of error.\label{tab:edu_overall}} 

\tiny
\begin{tabular}{@{}lcclllc@{}}
\toprule
&  Image Type&  Format&  \#1 / Error &  \#2 / Error&  \#3 / Error
 & Overall\\ \midrule
Q1-a&  Diagram&  Free text&  Imprecise&  Imprecise & Imprecise
 &\texttimes
\\
             Q1-b&  Diagram&  Free text& Imprecise& Imprecise&  Imprecise
 &\texttimes
\\
             Q1-c&  Diagram&  Free text&   Imprecise&   Imprecise&  Imprecise
 &\texttimes
\\
             Q2&  Photograph&  MCQ&   Reasoning&   Reasoning&  Reasoning
 &\texttimes
\\
             Q3-a&  3D-model&  Free text&  \checkmark  &  \checkmark  &  \checkmark 
 &\checkmark
\\
             Q3-b&  3D-model&  Free text&  \checkmark  &  \checkmark  &  \checkmark 
 &\checkmark
\\
             Q4-a&  Diagram &  Numerical&  \checkmark  &   Reasoning&  Reasoning
 &\texttimes
\\
             Q4-b&  Diagram&  Numerical&   Inference&   Inference&  Inference
 &\texttimes
\\
             Q5&  Diagram&  Draw&   Imprecise&   Imprecise&  Imprecise
 &\texttimes
\\
             Q6&  Photograph&  Numerical&  \checkmark  &  \checkmark  &  \checkmark 
 &\checkmark
\\
             Q7&  Photograph&  Numerical&   Reasoning&   Reasoning&  Reasoning
 &\texttimes
\\
             Q8-a&  3D-model&  Free text&  \checkmark  &   Reasoning&  Reasoning
 &\texttimes
\\
             Q8-b&  3D-model&  Free text&   Reasoning&   Reasoning&  \checkmark 
 &\texttimes
\\
             Q8-c&  3D-model&  Draw&   Imprecise&   Imprecise&  Imprecise
 &\texttimes
\\
             Q9-a&  Table&  Numerical&  \checkmark  &   Inference&  \checkmark 
 &\checkmark
\\
             Q9-b&  Table&  Numerical&   Reasoning&  \checkmark  &  \checkmark 
 &\checkmark
\\
             Q9-c&  Table&  Numerical&   Reasoning&   Reasoning&  \checkmark 
 &\texttimes
\\
             Q10&  3D-Model&  Free text&  \checkmark  &  \checkmark  &  \checkmark 
 &\checkmark
\\
             Q11&  Diagram&  Numerical&   Reasoning&   Reasoning&  Reasoning
 &\texttimes
\\
             Q12-a&  Diagram&  Numerical&   Reasoning&   Reasoning&  Reasoning
 &\texttimes
\\
             Q12-b&  Diagram&  Numerical&   Reasoning&   Reasoning&  Reasoning
 &\texttimes
\\
             Q12-c&  Diagram&  Numerical&   Reasoning&   Reasoning&  Reasoning
 &\texttimes
\\
             Q12-d&  Diagram&  Free text&  \checkmark  &  \checkmark  &  \checkmark 
 &\checkmark
\\
             Q13&  Diagram&  Numerical&   Reasoning&   Reasoning&  Reasoning
 &\texttimes
\\
             Q14-a&  Diagram&  Numerical&   Reasoning&   Reasoning&  Reasoning
 &\texttimes
\\
             Q14-b&  Diagram&  Free text&   Reasoning&   Reasoning&  Reasoning
 &\texttimes
\\
             Q14-c&  Diagram&  Free text&  \checkmark  &  \checkmark  &  \checkmark 
 &\checkmark
\\
             Q15-a&  Diagram&  Numerical&   Reasoning&   Reasoning&  Reasoning
 &\texttimes
\\
             Q15-b&  Diagram&  Numerical&   Reasoning&   Reasoning&  Reasoning
 &\texttimes
\\
             Q16-a&  Diagram&  Draw&  \checkmark  &  \checkmark  &  \checkmark 
 &\checkmark
\\
             Q16-b&  Diagram&  Free text&  \checkmark  &  \checkmark  &  \checkmark 
 &\checkmark
\\
             Q16-c&  Graph&  Free text&  \checkmark  &   Inference&  Inference
 &\texttimes
\\
             Q16-d&  Diagram&  Numerical&   Reasoning&  \checkmark  &  \checkmark 
 &\checkmark
\\
             Q16-e&  Diagram&  Numerical&   Reasoning&   Reasoning&  Reasoning
 &\texttimes
\\
             Q17-a&  Diagram&  Numerical&  \checkmark  &  \checkmark  &  \checkmark 
 &\checkmark
\\
             Q17-b&  Diagram&  Numerical&   Reasoning&   Reasoning&  Reasoning
 &\texttimes
\\
             Q17-c&  Diagram&  Numerical&  \checkmark  &  \checkmark  &  \checkmark 
 &\checkmark
\\
             Q18-a&  Diagram&  Numerical&   Reasoning&  \checkmark  &  Reasoning
 &\texttimes
\\
             Q18-b&  Diagram&  Numerical&   Reasoning&   Reasoning&  Reasoning
 &\texttimes
\\
             Q19-a&  Diagram&  Free text&   Reasoning&   Reasoning&  Reasoning
 &\texttimes
\\
             Q20-a&  Diagram&  Free text&  \checkmark  &  \checkmark  &  \checkmark 
 &\checkmark
\\
             Q20-b&  Diagram&  Free text&   Inference&   Inference&  Inference
 &\texttimes
\\
             Q21-a&  3D-Model&  Numerical&  \checkmark  &  \checkmark  &  \checkmark 
 &\checkmark
\\
             Q21-b&  3D-Model&  Numerical&  \checkmark  &  \checkmark  &  \checkmark 
 &\checkmark\\ \midrule
           \multicolumn{5}{@{}l}{Overall score} &  & 16 (36\%) \\ \bottomrule
\end{tabular}

\end{table}

\paragraph{Selected Questions and Answers}
In the following, we reproduce selected questions and answers to illustrate the type of questions, as well as the type of errors in the answers.

\begin{figure}
\begin{contextbox}{Q1 - Repeat 1 - Relating proportional changes}
\begin{promptbox}{}
A designer proposes to change the electric motor by increasing the number of windings of wire around the armature significantly (e.g., by doubling it). All other parameters of the motor are preserved as they were (radius of the armature, strength of the magnetic field, and so on). Describe the influence (if any) on: a) stall torque, b) no load speed, and c) maximum power.

\begin{center}
    \includegraphics[scale=0.3]{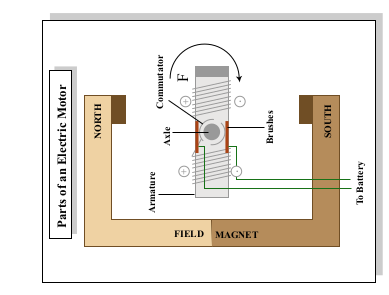}
\end{center}

{\tiny \textit{Image reproduced from~\citep{OCW_2007} under CC BY-NC-SA 4.0.}}
\tcblower

The diagram you provided shows a schematic representation of the main parts of a simple electric motor. When a designer proposes to increase the number of windings around the armature of an electric motor, while keeping all other parameters constant, the following effects can be generally anticipated:

\begin{enumerate}[a), leftmargin=*,labelindent=0mm]
    \item Stall torque: [...] \hl{Stall torque is proportional to the strength of the magnetic interaction}, and \hle{doubling the number of windings will increase the force that the motor can exert when it is not rotating}.
    \item No load speed: The no load speed [...] \hle{might decrease as a result of the increased windings}. [...]
    \item Maximum power: \hl{The maximum power is the product of torque and speed at which the power is maximum.} [...] \hle{the overall effect on maximum power is not straightforward} [...] the motor may operate more efficiently at lower speeds and could potentially have \hle{higher maximum power}. [...]
\end{enumerate}
\end{promptbox}
\end{contextbox}
    \caption{A diagram-based textbook problem about motor parameters. Q1 - Repeat 1.} %
    \label{edu_q1}
\end{figure}

First, we look at an example of imprecise answers. Figure~\ref{edu_q1} shows the question and answer to Q1 (a--c). Although GPT-4V can describe relationships between stall torque, no-load speed, and maximum power, it fails to provide the exact proportions by which the relationships increase or decrease. For Q1, the expected solution is that by doubling the number of windings, the stall torque doubles, the no-load speed is cut in half and the maximum power stays constant. Noteworthy, the provided answer also contains additional explanations that were not asked for.

\begin{figure}
\begin{contextbox}{Q2 - Repeat 1 - Reasoning on an electric circuit based on a photograph}
\begin{promptbox}{}

\begin{center}
    \includegraphics[scale=0.5]{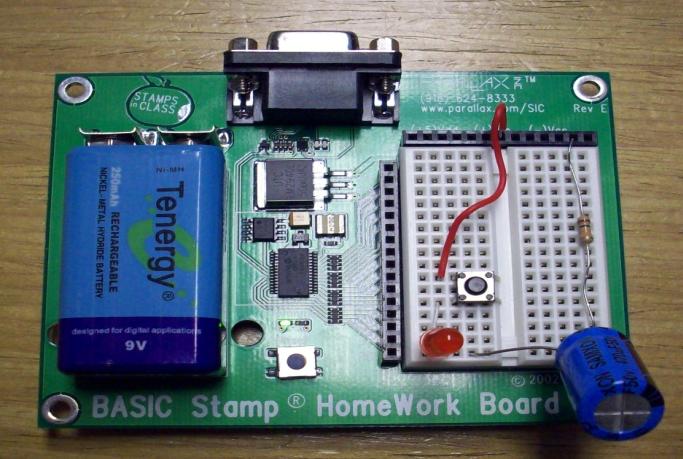}
\end{center}

The components shown here (resistor, capacitor, LED, and normally-open-button) are connected in series. The button is pressed and the LED lights up and then slowly dims over the course of 5 seconds. The button is released and 5 more seconds pass. The button is pressed again. Which statement best describes what happens:

\begin{enumerate}[label={(\alph*)},leftmargin=*,labelindent=0mm]
    \item The LED lights up and then slowly dims over the course of 5 seconds
    \item The LED starts dim then slowly brightens over the course of 5 seconds
    \item The LED lights up steadily
    \item LED does not light
\end{enumerate}

\textbf{Target:} (d)

{\tiny \textit{Image reproduced from~\citep{OCW_2007} under CC BY-NC-SA 4.0.}}
\tcblower

The image you provided shows an electronic setup on a breadboard with a 9V battery connected to a BASIC Stamp Homework Board. The board is labeled and has a microcontroller and other components, but the specific components you mentioned (resistor, capacitor, LED, and normally-open-button) are not directly visible in the picture you provided. [...]

Hence, the correct answer to your question would be:\\
\hle{a) The LED lights up and then slowly dims over the course of 5 seconds.}
\end{promptbox}
\end{contextbox}
\caption{Reasoning on an electric circuit based on a photograph. Q2 - Repeat 1.}
    \label{edu_q2}
\end{figure}

Next, we look at Q2, which is the only multiple-choice question, see Fig.~\ref{edu_q2}. The correct answer to this question is (d) since the capacitor is already charged up to the supply voltage and cannot unload through the LED. GPT-4V seems unable to understand the circuit based on the photograph and thus, bases its responses solely on the provided text. While it understood that the LED and the capacitor were in series, it got some basic physics concepts incorrect.

\begin{figure*}
\begin{contextbox}{Q9 - Repeat 3 - Calculate with information from a table}
\begin{promptbox}[sidebyside]{(Tabular)}
\begin{center}
    \includegraphics[scale=0.33]{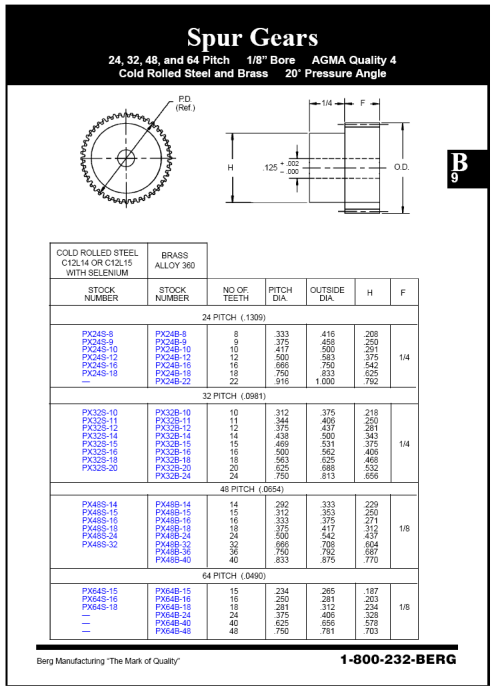}
\end{center}
If gear PX32B-10 and
PX32B-20 are mated together in a
gear train, how far apart should the
centers of their shafts be placed? If gear PX32B-10 and
PX32B-20 are mated together in a
gear train and a torque of 2 ft lbs is
applied to PX32B-10, what is the
torque on PX32B-20?

If gear PX32B-10 and PX32B-20 are mated together in a gear train and a torque of 2 ft lbs
is applied to PX32B-10, what is the direction and magnitude of the reaction force at the shaft where PX32B-20 is mounted?
State any assumptions needed to arrive at your answer.

{\tiny \textit{Image reproduced from~\citep{OCW_2007} under CC BY-NC-SA 4.0.}}
\tcblower

[...] From the chart, the pitch diameter (PD) for PX32B-10 (which has 10 teeth) is \hl{0.312 inches}, and the PD for PX32B-20 (which has 20 teeth) is \hl{0.625 inches}. To find the center distance, we add the two pitch radii:
\begin{align*}
\text{Center distance} &= (\text{PD of PX32B-10} / 2) + (\text{PD of PX32B-20} / 2)\\
       &= (0.312 / 2) + (0.625 / 2) = \text{\hl{0.4685 inches}}
\end{align*}
[...] assuming there is no loss due to friction or other factors, the power transmitted by one gear will be equal to the power transmitted to the other [...]

With PX32B-10 having 10 teeth and PX32B-20 having 20 teeth, PX32B-20 will rotate at half the speed of PX32B-10, due to the gear ratio of 2:1. So if a torque of 2 ft lbs is applied to PX32B-10, the torque on PX32B-20 will be double, given by: [...]
\begin{align*}
\text{Torque on PX32B-20} &= \text{\hl{4 ft lbs}} \\
\end{align*}
As for the direction and magnitude of the reaction force at the shaft where PX32B-20 is mounted, the force can be calculated using the relationship between torque (T), force (F), and radius (r):
\[\text{Torque (T) = Force (F) \texttimes Radius (r)}\]
[...]

Assuming a pure torque application with no other forces at play (like axial or radial loads) and using the pitch radius for PX32B-20, which is half of the pitch diameter (0.625 / 2 = 0.3125 inches):
\[F = 4\text{ ft lbs} / 0.3125\text{ inches} = 4\text{ ft lbs} / 0.02604\text{ feet} \approx \text{\hl{153.61 lbs}}\]
[...] The direction of this force will be radially inward toward the center of the gear, opposite to the force that PX32B-10 is applying to PX32B-20, because for every action there is an equal and opposite reaction, according to Newton's third law of motion.

\end{promptbox}
\end{contextbox}
    \caption{Calculating answers using information from a table. Q9 - Repeat 3.}
    \label{edu_q9}
\end{figure*}

Finally, we look at a question that requires extracting values from a table and performing calculations using them, see Fig.~\ref{edu_q9}. While arguments can be made to input tabular data as text, tables are often inconvenient to input and are more convenient for users to input as images. In this particular answer (repeat 3), GPT-4V is able to correctly extract the values and calculate the center distance, the torque, and the reaction force. It is worth keeping in mind, see Table~\ref{tab:edu_overall} that the two other repeats were not as successful, showing how challenging such a task is.

\subsubsection*{Discussion: Textbook Problems}
As previously mentioned, GPT-4V makes three types of mistakes: reasoning, image misinterpretation, and imprecision. We go into detail below.

\paragraph{Reasoning} GPT-4V can sometimes provide incorrect answers as a result of reasoning errors. This was especially apparent with multi-step reasoning tasks. When asked what happens to a system after a series of actions are performed, such as the one given in Fig.~\ref{edu_q2},  GPT-4V often makes a mistake in reasoning, such as hallucinating a fact, and derails its chain of thought. Furthermore, When provided a question to compute a numerical answer, GPT-4V can sometimes have the correct methodology for arriving at the correct answer but provides an incorrect answer due to an incorrect numerical approximation. For example, when asked to compute $e^{0.3\pi}$, GPT-4V instead used $e^\pi$ as an approximation and arrived at an incorrect answer even though its previous steps were correct. This issue is not difficult to alleviate, however, as previous papers have shown that leveraging tools such as calculators can enable GPT-4V to perform numerical reasoning tasks better.

\paragraph{Image misinterpretation} GPT-4V can have trouble understanding and inferring information in images. In Fig.~\ref{edu_q2}, GPT-4V fails to interpret the circuit shown as a photograph. As a result of this misinterpretation, GPT-4V cannot leverage the information present in it and only answers based on the provided text.

\paragraph{Imprecision} While GPT-4V has reasonable success in providing qualitative answers, it sometimes fails to do so as its answers are too vague or do not capture the main idea of a question. In a question that asked about potential errors in an injection molding experiment, GPT-4V provided a long list of potential issues, but not issues that were specific to the question. GPT-4V also had issues with relating exact numerical relationships between different variables as seen from Fig.~\ref{edu_q1}.

\subsection{Summary} Here we provide a summary of our findings regarding GPT-4V's ability to solve engineering textbook problems.

\begin{enumerate}[(i)]
    \item \textbf{High Level Reasoning} 
    Is GPT-4V able to reason about domain-specific at a high level?
    \begin{itemize}
    \item We evaluated GPT-4V on 16 free-text questions, which ask about high-level reasoning knowledge on Mechanical Engineering topics. Of the 5 types of questions, GPT-4V performs relatively better in this area compared to other tasks. Free-text questions tend to be more open-ended and require less precise reasoning than other types of questions. As a result, GPT-4V can provide answers that are considered correct even though they may not fully correspond to the model solution.
    \end{itemize}
    \item \textbf{Numerical Reasoning} 
    How strong is GPT-4V at numerical reasoning tasks?
    \begin{itemize}
        \item We evaluated GPT-4V on 24 different numerical reasoning tasks and it answered 8 of them correctly. We note that GPT-4V's inability to answer numerical questions comes from two sources. First, it is unable to precisely compute answers, leading future steps in the computation to deviate from the answer. Second, it is unable to logically incorporate domain knowledge during its reasoning process, causing it to use incorrect formulas.
    \end{itemize}
    \item  \textbf{Failure Modes}
    When does GPT-4V fail to provide satisfactory answers?
    \begin{itemize}
        \item We classified the failure modes of GPT-4V into three cases: reasoning, inference, and imprecision. Reasoning errors happen either due to failing complex, precise logical steps or incorporating domain knowledge. This was especially apparent for multi-step reasoning tasks. Inference errors happen when GPT-4V fails to incorporate image information into its answers. Imprecision errors happen when GPT-4V is unable to calculate correct answers as alluded to in the above section.
    \end{itemize}

\end{enumerate}

\section{Spatial Reasoning Abilities}\label{sec:spatial}
\subsection{Mental Rotation and Packing Tests}
Spatial reasoning is the ability of humans to perform mental spatial operations: rotations, translation, projection, and orientation. Spatial reasoning is at play when humans read maps, navigate their homes at night without light, or solve pretty much any problem in the fields of science, technology, engineering, and mathematics (STEM)~\citep{maeda_meta-analysis_2013}.  Spatial reasoning skills are considered essential skills for understanding graphs, diagrams, plots, 3D objects, and representations. Indeed, multiple studies have found that spatial abilities are a good predictor of academic success~\citep{Shea2001,berkowitz_which_2018}. Consequently, spatial reasoning skills have been well studied in humans, and many standardized tests exist, e.g., The \textit{Revised Purdue Spatial Visualization Test: Visualization of Rotations (PSVT:R)}~\citep{yoon_psychometric_2011}, the \textit{Mental Cutting Test ``Schnitte''}~\citep{QuaiserPohl2003}, or the \textit{Novel Spatial Ability Tests}~\citep{berkowitz_spatial_2021}.

Following some of our observations on the apparent struggles of GPT-4V regarding spatial understanding, we specifically tested its spatial abilities in order to provide additional insights. Spatial reasoning tests are also good candidates to evaluate vision language models since they focus on inherently visual tasks and are often not publicly available, to maintain their validity. They are thus unlikely to be part of the training data.

\paragraph{Methodology}
We assessed GPT-4V's spatial reasoning skills using the openly accessible  \textit{packing test} (part of the \textit{Novel Spatial Ability Tests}~\citep{berkowitz_spatial_2021}), and the \textit{MechE Rotation Test}~\citep{picard_meche_2023}. While the first one is openly accessible, the latter is released publicly for the first time in parallel to this work.

\begin{figure}[ht!]
    \centering
    \includegraphics[width=0.75\linewidth]{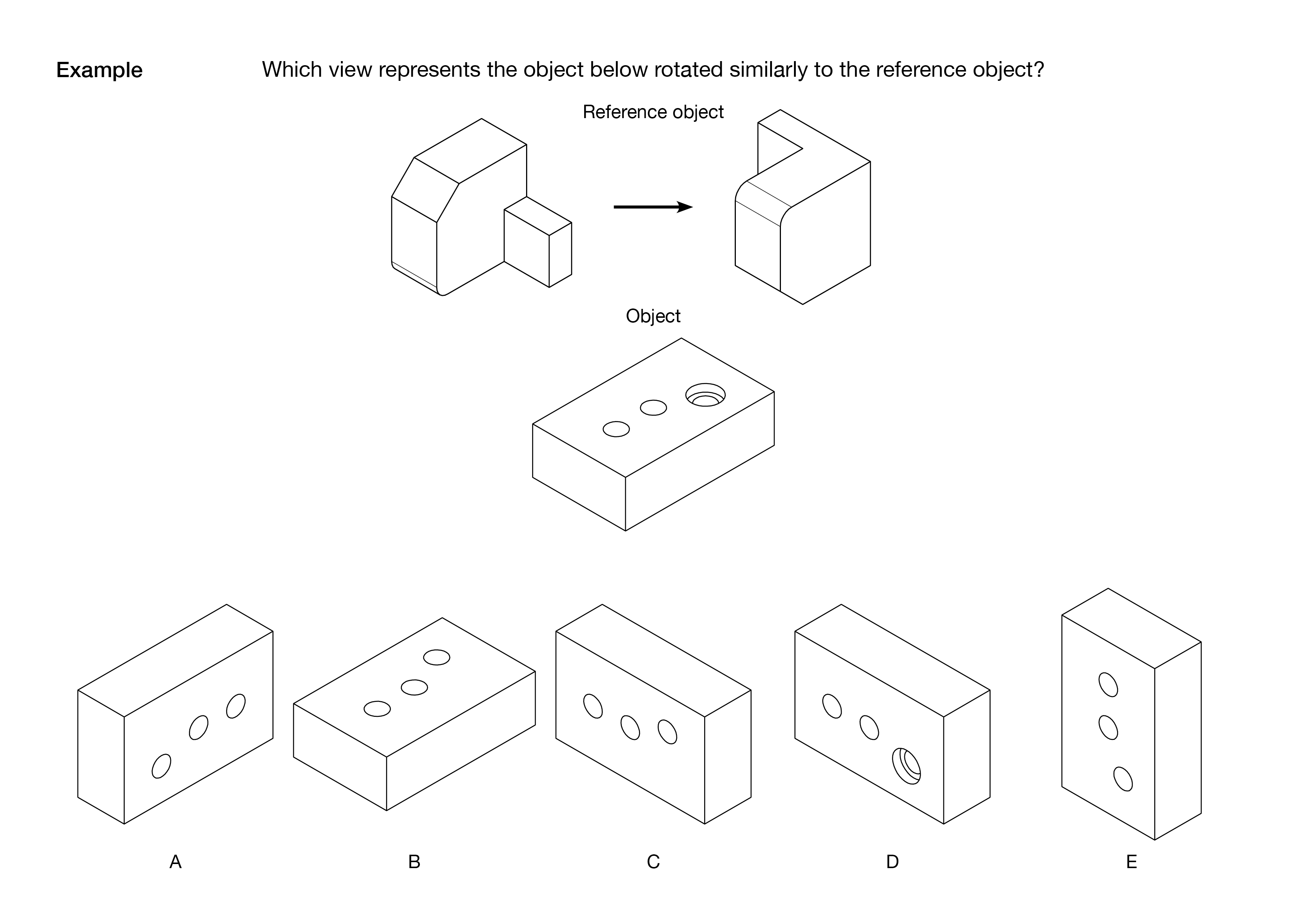}\\
    \includegraphics[width=0.49\linewidth]{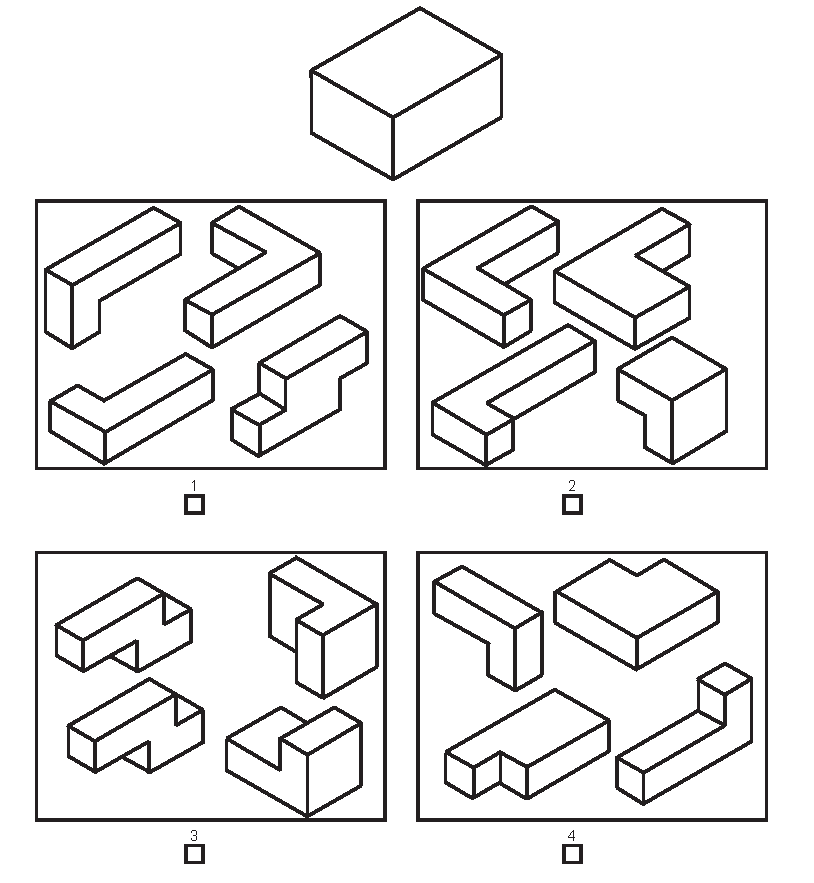}%
    \includegraphics[width=0.49\linewidth]{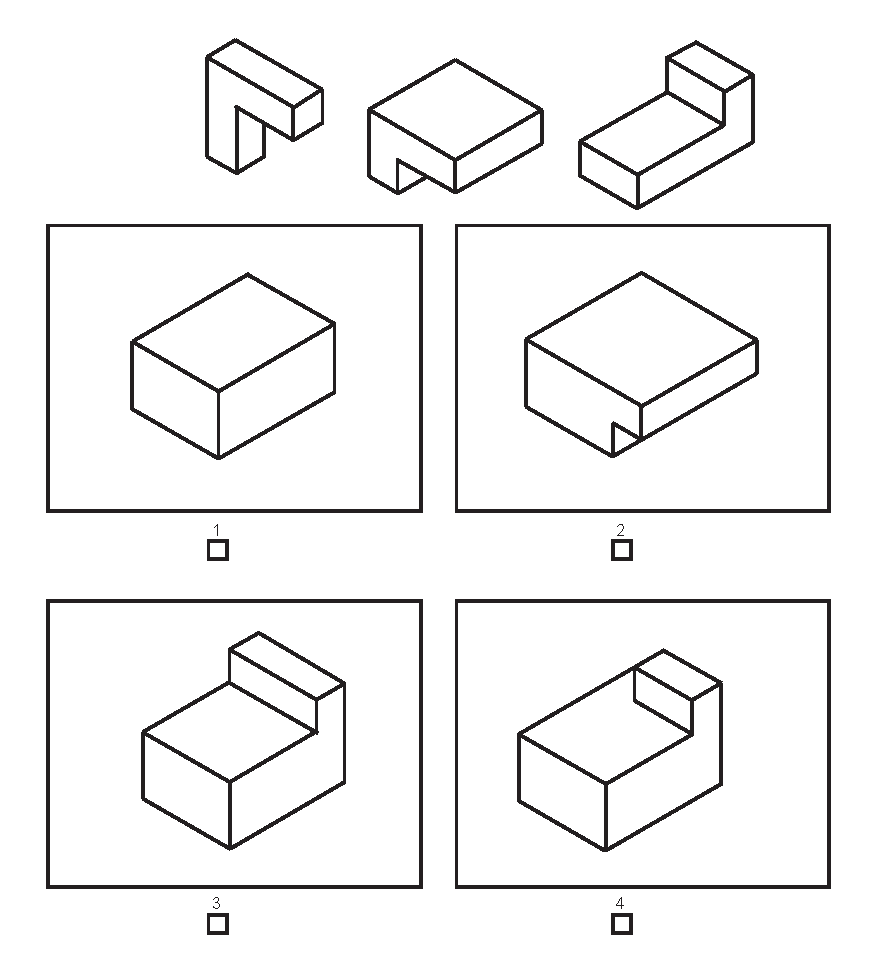}
    \caption{Example questions from the considered spatial reasoning tests.}
    \label{fig:sr-examples}
\end{figure}

The \textit{MechE Rotation Test} follows the general principles of the PSVT:R, but uses objects with features typically seen on mechanical parts. It measures the ability of participants to visualize one or two rotations applied to a reference object and apply them to another object. For each question, five possible configurations of the object are shown and the participants select the correct one. The test is composed of an example---for which the correct answer is given to the participant---followed by ten questions of increasing difficulty. The \textit{packing test} requires participants to evaluate if shapes can be composed of or decomposed into smaller sub-shapes. The \textit{packing test} is split into two parts: packing and unpacking. In the first part, participants have to choose among four options which set of sub-shapes can be packed together to form a larger shape. In the second part, participants do the opposite and select among four large shapes, which can be decomposed into the provided smaller shapes. The example questions for these tests are shown in Fig.~\ref{fig:sr-examples}. 

In this work, we have GPT-4V take the tests as they are given to human participants: We type the exact textual instructions from these standardized tests and include the corresponding images.
Each questionnaire is passed in a single context, sequentially going through the examples and the questions, and providing the instructions and images. To account for some stochasticity, each questionnaire is repeated five times. In addition, and inspired by~\citep{Yang2023}, we evaluate if adding visual marks---reference coordinates and coloring faces---improved the performance of the model on the \textit{MechE rotation} test. The runs using the original test are referred to as \textit{Run H}, while the ones with the marks are called \textit{Run P}.

The full set of prompts is made available as a benchmark for any future vision language model.

\begin{figure}
    \centering
    \includegraphics[width=0.75\linewidth]{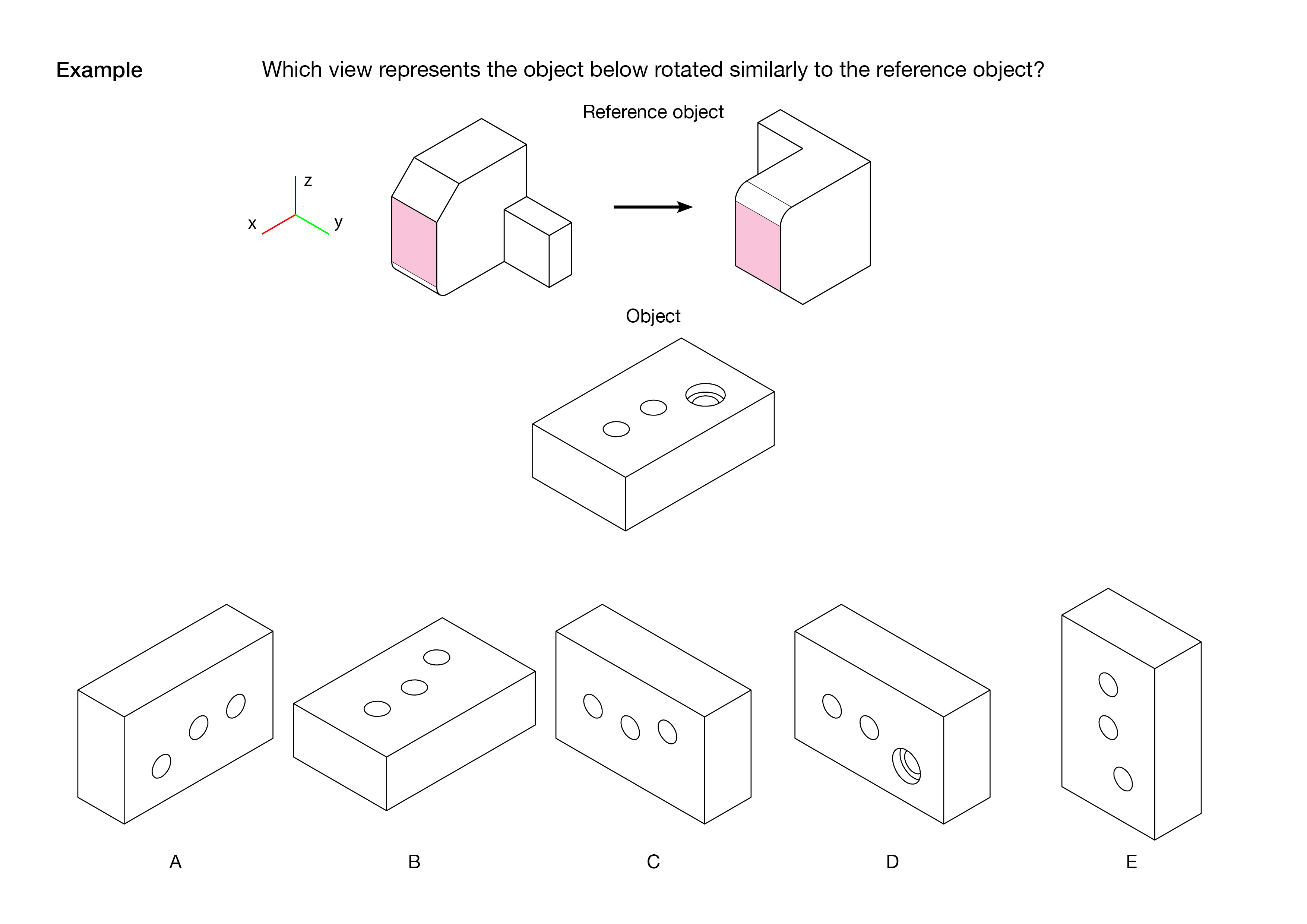}
    \caption{Example of the MechE rotation test with the additional visual prompts to support the model.}
    \label{fig:sr-example-with-visual}
\end{figure}

\paragraph{Scores on The Spatial Reasoning Tests}

The answers of GPT-4V for the \textit{packing test} and the \textit{MechE rotation test} are provided in Table~\ref{tab:packing_scores} and~\ref{tab:mech_rot_scores}, respectively. Starting with the \textit{packing test}, GPT-4V obtains an average score across five runs of 36\%, slightly higher than the expected average score if answering at random (25\%). Interestingly, all five questions that have been answered correctly have been done so by at least two runs, further suggesting that GPT-4V is not answering at random. In comparison to humans, however, it remains significantly lower than the average score of undergraduate (66\%) and graduate (73\%) students reported by~\citep{berkowitz_spatial_2021}. For the \textit{MechE rotation test}, the average scores (16\% and 20\%) are lower and closer to the expected score for random answering (20\%). While slightly higher, it is unclear if the visual prompting supports GPT-4V. While no human results have been published for this test, average scores between 60\% and 70\% are expected based on internal tests and by comparison to the revised PSVT:R test.

\begin{table}
    \centering
    \caption{Answers and scores for the Packing Test~\citep{berkowitz_spatial_2021}. Correct answers are in bold. Each run was conducted within the same context.}
    \label{tab:packing_scores}
    \begin{tabular}{@{}cccccc@{}} \toprule
         (Correct) &  Run 1 & Run 2 & Run 3 & Run 4 & Run 5  \\ \midrule
         Part 1 Q1 (4) & 3 & 2 & 3 & 3 & 3 \\
         Part 1 Q2 (3) & 4 & 1 & 4 & 4 & 4 \\
         Part 1 Q3 (3) & \textbf{3} & \textbf{3} & \textbf{3} & \textbf{3} & \textbf{3} \\
         Part 1 Q4 (2) & \textbf{2} & 1 & \textbf{2} & \textbf{2} & \textbf{2} \\
         Part 1 Q5 (1) & 3 & 4 & 4 & 4 & 3 \\
    \midrule
         Part 2 Q1 (3) & \textbf{3} & \textbf{3} & 2 & \textbf{3} & \textbf{3} \\
         Part 2 Q2 (3) & 4 & 1 & 1 & 4 & 1 \\
         Part 2 Q3 (2) & \textbf{2} & 4 & 3 & \textbf{2} & \textbf{2} \\
         Part 2 Q4 (2) & 1 & 3 & 1 & 3 & 1 \\
         Part 2 Q5 (3) & \textbf{3} & 1 & 4 & 4 & \textbf{3} \\ \midrule
         Scores& 50\% & 20\% & 20\% & 40\% & 50\% \\
 Average & & & & & 36\%\\ \bottomrule
    \end{tabular}
\end{table}

\begin{table*}
    \centering
    \caption{Answers and scores to the MechE Rotation Test~\citep{picard_meche_2023}. Correct answers are in bold. Each run was conducted within the same context.}
    \label{tab:mech_rot_scores}
    \begin{tabular}{@{}c ccccc|ccccc@{}}\toprule
         (Correct) & H1 & H2 & H3& H4&H5 & P1& P2& P3& P4&P5
\\ \midrule
         Q1 (D) & C&C & C& \textbf{D}&\textbf{D}
 & \textbf{D}& \textbf{D}& C& C&B
\\
         Q2 (A) & D&B & \textbf{A}& D&C
 & D& B& D& B&D
\\
         Q3 (C) & \textbf{C}&\textbf{C} & \textbf{C}& E&B
 & \textbf{C}& D& B& \textbf{C}&\textbf{C}
\\
         Q4 (E) & C&D
 & A& A&C
 & C& C& D& B&D
\\
         Q5 (C) & \textbf{C}&B
 & A& B&B
 & D& B& B& \textbf{C}&\textbf{C}
\\
         Q6 (B)& E&D
 & C& D&C
 & D& E& E& D&E
\\
         Q7 (C) & E&E
 & D& \textbf{C}&D
 & E& \textbf{C}& \textbf{C}& B&D
\\
         Q8 (A) & B&C
 & C& B&E
 & C& D& B& C&C
\\
         Q9 (E)& D
 &B
 & A& C&B
 & C& A& C& B&\textbf{E}
\\
 Q10 (A) &C&C & C& B&D
 & B& E& B& C&B
\\ \midrule
 Scores& 20\%&10\% & 20\%& 20\%&10\% & 20\%& 20\%& 10\%& 20\%&30\%
\\
 Average & & & & & 16\% & & & & &20\%\\ \bottomrule
    \end{tabular}
\end{table*}

\begin{figure}
\begin{contextbox}{}
\begin{promptbox}{}
    You will be shown a set of questions evaluating your spatial reasoning abilities, specifically your ability to mentally rotate 3D objects. Each question will demonstrate a set of one or two rotations applied to a reference object (at the top of the image, original position on the left, rotated configuration on the right). Study how the reference object has been rotated and then apply the same rotations to the object in the middle and select from among the five options given at the bottom which one represents the rotated configuration.

    Here is an example question. What is the correct configuration for this example?

    \begin{center}
        \includegraphics[width=0.7\linewidth]{RotationTest-2024-Ex-P.png}
    \end{center}
    
    In the example question, the reference object is rotated 180°. Answers A, C, D, and E are wrong since they cannot be obtained by a 180° rotation. Answer B is the correct answer.
    
    You will now see 10 questions and will be asked to identify which configuration is the correct one. Always start by stating how the reference object is rotated and then answer which configuration among the choices A, B, C, D, or E represents the rotated configuration.

    \tcblower

    Alright, please provide the first question, and I'll help you identify the correct configuration based on the rotation of the reference object.
\end{promptbox}

\begin{promptbox}{}
    \begin{center}
        \includegraphics[width=0.7\linewidth]{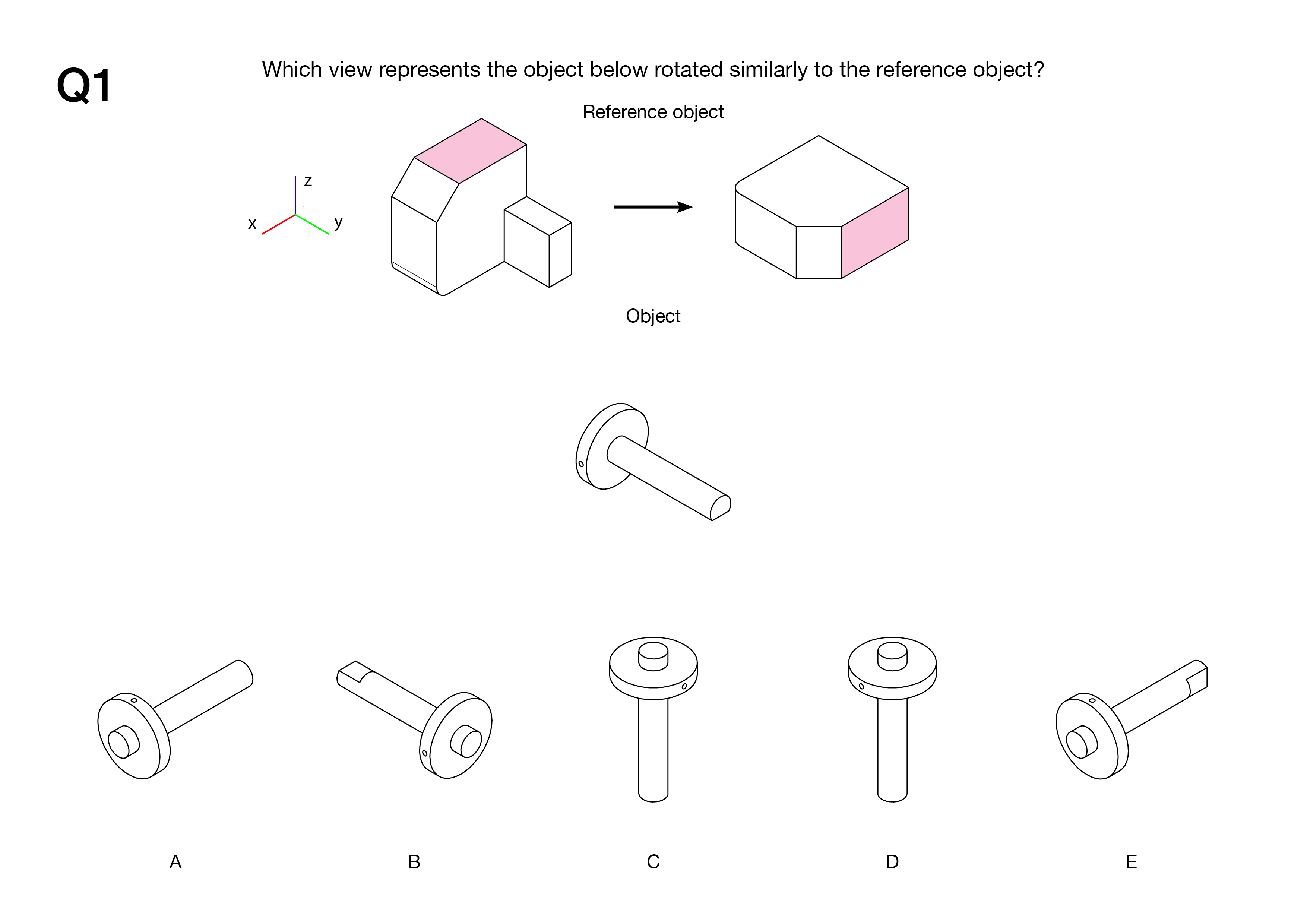}
    \end{center}

    \tcblower

    For Q1: The reference object has been \hl{rotated 90$^\circ$} around the \hle{Z-axis}.

Considering the options:
\begin{itemize}[leftmargin=*,labelindent=0mm,itemsep=-1pt,topsep=2pt]
    \item \hle{A and E are simply showing the original position, so they are incorrect.}
    \item \hle{B has been rotated 90° around the Y-axis, so it's incorrect.}
    \item \hle{C is also incorrect as it has been rotated 90° around the X-axis.}
    \item \hle{D is the correct representation of the object when rotated 90° around the Z-axis.}
\end{itemize}
Answer: \hle{D.}
    
\end{promptbox}
\end{contextbox}
\caption{First two prompts for the run P1 of the MechE rotation test, with the corresponding answers by GPT-4V. Both prompts are executed consecutively in the same context.}
\label{cont-runp1}
\end{figure}

To gain more insights, run P1 is reproduced in Fig.~\ref{cont-runp1}. GPT-4V's answers seem to indicate that the nature of the test and the task is well understood. However, while the answer is correct, the reasoning is incorrect. Based on the provided coordinate system, the reference object is rotated by $90^{\circ}$ around the X-axis, and not the Z-axis as stated by GPT-4V. This type of behavior has already been reported regarding numerical reasoning~\citep{Stechly2023}. As such, it seems like additional visual or textual instructions are needed to properly root the model within a spatial system.

\paragraph{Discussion}
Overall, our evaluation of the spatial abilities of GPT-4V using standardized (human) tests tends to suggest that, compared to humans, GPT-4V has some, although limited, spatial reasoning capabilities. Indeed, while these visualization tasks are hard and constructed to be somewhat deceptive, most untrained undergraduate students in science and technical fields answer at least half of the questions correctly~\citep{yoon_psychometric_2011,berkowitz_spatial_2021}. Unfortunately, this seeming lack could, in part, explain GPT-4V's limitations in performing engineering design tasks, such as CAD generation, see Section~\ref{sec:cad_and_engineering_drawings}. These results also corroborate findings recently reported in the literature~\citep{wen2023road}.

\subsection{Summary} Our evaluation of GPT-4V's performance on standardized spatial reasoning tests is summarized below.

\begin{enumerate}[(i)]
    \item \textbf{Standardized Spatial Reasoning Tests} Can GPT-4V correctly answer multiple-choice questionnaires created to assess spatial reasoning skills in humans?
    \begin{itemize}
        \item We repeated the MechE rotation and the packing tests five times each for a total of $100$ multiple-choice questions. The model's answers are similar success rates expected for the number of options in these questions.
        \item Further, we investigated a visual prompting approach based on~\citep{Yang2023} on the MechE rotation test aiming to assist the model. Repeated five times, we see no improvement in model performance over the unmodified test.
    \end{itemize}
    
\end{enumerate}

\section{Benchmarking LLaVA}\label{sec:llava}

We further explored how well VLMs perform engineering design tasks by performing the quantitative experiments with LLaVA~1.6~34B. It is an open-source VLM with both a chatbot and API interface, and for our experiments, we utilized the API interface with a ``temperature'' and ``top-k'' of 1.0. The results of the quantitative experiments using both GPT-4V and LLaVA~1.6~34B are shown in Table~\ref{tab:gpt_llava}.

Across all 12 relevant quantitative experiments, GPT-4V outperforms LLaVA~1.6~34B. It is important to note that LLaVA~1.6~34B is only able to intake one input image per context, therefore the tasks that required two or more input images cannot be run with this model. We denote this in Table~\ref{tab:gpt_llava} where applicable, such as for the spatial reasoning experiments. One quantitative experiment that we ran with only GPT-4V was the design similarity triplet experiments. The reason being that LLaVA~1.6~34B often output nonsensical answers. For example when presented with three designs labeled A, B, and C, and asked which design is most similar to A, LLaVA~1.6~34B often answered ``Design D,'' or other irrelevant answers like ``The image in the middle.'' Therefore we could not use the triplet similarity results to measure self-consistency or transitive violations. We note, however, that GPT-4V did not have this problem. 

Overall, this section demonstrates how the datasets and experiments published as part of this work can be used to benchmark future VLMs on engineering design tasks. Furthermore, these results reinforce the need for field-specific relevant datasets to evaluate the performance of machine-learning models. Indeed, our results strongly contrast with the performance reported by LLaVA~1.6~34B's authors, showing that the model outperforms GPT-4V and Gemini Pro on several benchmarks~\citep{liu2024llavanext}.

\begin{table}
    \caption{Total score for both GPT-4V and LLaVA~1.6~34B on the quantitative experiments in this work. The maximum possible score is provided as the ``Max." column.}
    \label{tab:gpt_llava}
    \begin{tabular}{@{}llccc@{}}
    \toprule
          &&   Max.&GPT-4V &  LLaVA 1.6 34B \\ \midrule
          \multicolumn{2}{@{}l@{}}{Design description}&   && \\
          &With text description&   30&30&  26\\
          &No text description&   30&16&  14\\
 & No text descr., no N/A&  30&21& 13\\
 \multicolumn{2}{@{}l@{}}{Engg. Drawing Analysis}&   99&86&  29\\
          \multicolumn{2}{@{}l@{}}{CAD Generation (1st try)}&   54&28&  3\\
 \multicolumn{2}{@{}l@{}}{Topology Optimization}& 90& 68 & 43 \\

          \multicolumn{2}{@{}l@{}}{Design for Manufacturing}&   &&  \\
          &Additive&   90& -22 &  N/A\\
 & Machining Features&  60 & 0 & 0 \\
 \multicolumn{2}{@{}l@{}}{Crack Inspections}& 345 & 204 & 172 \\
  \multicolumn{2}{@{}l@{}}{Textbook questions}&  135&51& 26\footnotemark[1]\\ 
 \multicolumn{2}{@{}l@{}}{Spatial Reasoning}&  && \\
 & Rotation& 100 & 18 & N/A\\
 & Packing& 50 & 18& N/A\\ \midrule
 \multicolumn{2}{@{}l@{}}{Total} & 1113 & 540 & 326\\
 \bottomrule
\end{tabular}

\footnotetext[1]{15 of the 135 textbook questions involved two input images, which LLaVA~1.6~34B does not support.}
\end{table}

\section{Discussion}\label{sec:discussion}

In this paper, we aimed to assess the capabilities of VLMs in several engineering design tasks ranging from conceptual design to manufacturing, and develop a reusable benchmark to evaluate future VLMs. Below, we present a detailed discussion of our findings from each section.

\paragraph{Conceptual Design} We examined design similarity analysis, sketch descriptions, and concept selection. We discovered that GPT-4V could evaluate design similarity with high self-consistency and minimal transitive violations. It was also consistent with human-generated idea maps in identifying unique sketches and groups of similar sketches.
Additionally, it effectively matched design sketches to their descriptions when provided with the entire sketch (an average score of 10/10), including a handwritten description, but without the description, it often chose ``None of the above,'' and therefore performed worse (average score of 5.33/10). When ``None of the above'' was not an option, GPT-4V performed better (an average score of 7/10).  This suggests a level of ``caution,'' so when GPT-4V has the chance to not be incorrect, it takes it. GPT-4V could generate useful and accurate text descriptions of designs even for sketches with very low drawing scores. Lastly, the model generated appropriate selection criteria but did not generate Pugh charts when only provided with design sketches. Overall, GPT-4V shows the great potential of VLMs for design sketch analysis and supporting the conceptual design stage beyond what has been identified in previous work~\citep{siddharth_natural_2022,stella_how_2023}.

\paragraph{System-Level And Detailed Design} We investigated GPT-4V's ability to use several Ashby diagrams to suggest appropriate materials, analyze engineering drawings, and generate CAD scripts. We found that GPT-4V could correctly respond where to look for materials in Ashby diagrams, but made errors when asked to be specific. The model faced difficulty in understanding the nuances of a block-with-blind-hole engineering drawing, but it was able to extract most dimensions and assign them appropriate labels. In terms of CAD generation ability, GPT-4V was successful in generating correct CAD scripts on the first attempt only in one out of nine experiments, and our iterations to fix the scripts did not improve the results. LLaVA~1.6~34B scored lower on the engineering drawing analysis and was unable to make a correct CAD script.

We investigated GPT-4V's ability to understand and analyze structures resulting from topology optimization (TO). GPT-4V showed understanding of high-level TO principles like volume fraction, and proposed realistic locations for where boundary conditions may be. However, when provided with an image, it struggled to estimate the volume fraction of a design or identify the presence of floating material (Table~\ref{tab:to-more-quantitative}). When allowed to use a code interpreter, however, GPT-4V was able to estimate volume fraction much more effectively. This suggests that for certain tasks, specifically those using spatial reasoning, engineers may benefit by integrating VLMs with other tools or plug-ins.

\paragraph{Manufacturing and Inspection} In the manufacturing stage, we tested GPT-4V's understanding of design for manufacturing (DfM) for subtractive and additive manufacturing operations. GPT-4V was, as we interpret it, cautious, and suggested that none of the additive manufacturing parts were printable, even when the parts were well within the provided guidelines. On the feature identification task for subtractive manufacturing, GPT-4V was able to identify at least one feature 12 out of 20 times, but never all of them. The provided explanations were most of the time inconsistent and confused different technical terms. Furthermore, we assessed GPT-4V's ability to inspect images to find and identify defects. For the evaluated cases, GPT-4V tended to overly predict the presence of defects and was inconsistent in identifying the type of defect. Even more than in the detailed design stage, design for manufacturing and inspection is all about precision, and GPT-4V overall fails to deliver reliable and consistent performance on the evaluated task.

\paragraph{Engineering Textbook Problems}
We evaluated GPT-4V's ability on textbook problems in engineering. Overall, GPT-4V achieves rather low scores with a 36\% accuracy. It performed the best for textbook problems asking for explanations (free text answers) but struggled for numerical questions, both in the reasoning and in the numerical value extraction from the provided images.

\paragraph{Spatial Reasoning}
Lastly, we evaluated GPT-4V's spatial reasoning abilities through specific tests typically used to evaluate humans. Overall, GPT-4V achieves rather low scores with 36\%, and 18\% accuracy over the packing test, and the MechE rotation test respectively. GPT-4V's scores are indistinguishable from random answer picking and the provided explanations did not match the visual representations. Given these low scores, the spatial reasoning tests could become a competitive benchmark to evaluate future multimodal LLMs. However, the current protocol for these tests, following the testing protocol for humans, requires models to accept multiple images, a capability most models don't currently have.

\paragraph{Looking Forward: Vision-Language Models in Engineering Design}

\begin{itemize}
    \item Future VLMs that are meant to assist with engineering tasks must understand dimensions, scales, and spatial reasoning. The visuals in engineering tasks often provide critical spatial information.  However, current VLMs show weak performance when interpreting images for their precise spatial information. For instance, in Section~\ref{sec:spatial}, which explores spatial reasoning tasks, GPT-4V performed with about the accuracy of random guessing. Furthermore, in Section~\ref{sec:cad_and_engineering_drawings}, GPT-4V was only successful in distinguishing that a design had a blind hole rather than a through hole from an engineering drawing 11\% of the time. As shown in Section~\ref{sec:to}, both models in this study performed poorly when estimating the volume fraction of an image. Understanding relative dimensions, scales and orientations within a visual will greatly increase a VLM's utility for engineering tasks and should be an area of future development.

    \item To best serve engineering tasks, future models should be designed to process multiple images and text simultaneously. Many engineering tasks require viewing visuals in context with each other. This allows engineers to cross-reference charts, compare designs, etc. The most useful VLMs will also have this functionality so that they can more seamlessly integrate into an existing engineering workflow.
    
    \item A useful model must understand and follow instructions. Especially if an engineer is using a VLM for the benefit of automation at large scales, this benefit is undermined if the VLM does not properly follow instructions. For example, if instructed to select from options A, B, and C, the model should not generate a fourth option D. This is a problem that arose in Section~\ref{sec:concepts} in both matching design descriptions and answering the triplet similarity questions. %
    
    \item Engineers should be aware that current models may require users to iteratively adjust their prompts and images to obtain desired results. For certain tasks, this iteration and correction of the VLM may render its benefits moot. For other tasks, iterative feedback could be useful for exploring different solution strategies, or for ensuring that the user has mastery of a subject, and can therefore correct a VLM. The need for iteration is demonstrated in~\citep{makatura_how_2023}, which shows that LLMs have great potential to generate valid CAD scripts, however these results require users to iteratively guide the model to a desired output.

    \item If a VLM is unsure of an answer, it has two primary options for its response: indicate that it is unsure and perhaps request more information, or make up a response that sounds correct. For engineers, we propose that a more useful VLM will indicate that it is unsure and ask for additional information. This occurred in Section~\ref{sec:concepts} when trying to make a Pugh chart for concept selection. In this case the model requests more information about each design. In one trial, GPT-4V stated ``Since I cannot physically interact with the concepts to evaluate them against the criteria you've provided, such as "Ease of handling" or "Durability," I am unable to make informed decisions about the scores. These criteria often require subjective assessments or empirical testing that cannot be performed without direct interaction with the designs. However, I can guide you on how you might consider each criterion for scoring:\ldots'' This indicates that practitioners should be prepared to engage in an iterative information exchange with AI tools.

\end{itemize}

While this work investigates GPT-4V in-depth, we proposed a set of benchmark cases that could be used to assess other current and future models and that could inform about their performance for engineering design. In particular, we provide our datasets, prompts, and specific questions, where applicable. To demonstrate this work's utility for benchmarking other VLMs, we perform comparative experiments with LLaVA~1.6~34B, and show the results in Table~\ref{tab:gpt_llava}. Other open-source vision language models have been released recently, such as Fuyu-8B~\citep{fuyu-8b}. Unfortunately, most of these models have not yet been fine-tuned and aligned to the level of GPT-4V, making them less practical out-of-the-box. As we have shown in Section~\ref{sec:llava}, LLaVA~1.6~34B obtains slightly worse results across all of the quantitative experiments in this work. These open-source models, however, can be particularly useful as base models to develop custom fine-tuned models, for example, to specifically target the needs of engineers.

\paragraph{Limitations of This Study}
Evaluating the performance of GPT-4V faces the same challenges and concerns raised in previous studies of LLMs~\citep{mao2023gpteval}. Below, we highlight a few of the limitations.

Specificity of Engineering Problems: While we attempted to cover a wide range of engineering tasks, the study still focuses on a subset of engineering design problems. This could limit the applicability of the findings to other challenges encountered in the broad spectrum of the field.

Dependency on Prompt Engineering: The results might be highly sensitive to how prompts are engineered. Subtle variations in prompt structure or wording could lead to markedly different responses from the model, affecting the reliability of the evaluations.

Dataset Representativeness: 
The selected benchmark datasets, their quality, diversity, and representativeness can significantly impact the model's performance and our results. 
While we created a large set of evaluation problems for vision-language models, we recognize that these datasets might not fully capture the diversity and complexity of real-world engineering scenarios. This could affect the generalizability of the results to practical engineering applications. 

Black-box and evolving models: 
Model changes, including data leaks, and the lack of control when using the chat interface, mean that we cannot fully define the experimental environment and some of the results may be different if reevaluated.
However, for the assessments, we strived to create larger benchmarks within the limitation of the chat interface and repeated the experiments to obtain a better sample of the model's performance. 
Further, since AI models are frequently updated, our results may not hold for long. Yet, while new vision-language models will enable new capabilities, we believe that this study provides a lot of value by demonstrating the tasks that future models should be evaluated on as well as providing these tasks in our datasets. We release all our quantitative datasets to measure how much future models improve for different engineering problems.

Human-AI Interaction: An important part of the engineering design process involves how humans interact with design. In this study, we did not test the capabilities of how human designers may interact with Vision-Language Models and how this interaction influences the problem-solving process is crucial, as human biases, trust, and interpretations can affect the outcomes~\citep{Zhang2023ai}.

In conclusion, while this study offers valuable insights into the capabilities of VLMs in addressing engineering design problems, it is essential to recognize these limitations as an integral part of our findings. They highlight the areas needing further exploration and remind us of the cautious approach required when generalizing AI capabilities to broader real-world applications. Our research is a step in an ongoing journey, contributing to the evolving dialogue on the role and effectiveness of AI in complex, multifaceted fields like engineering design.

\section{Conclusion and Future Work}

The first avenue for future research following this study involves expanding the scope and depth of the engineering problems evaluated. This can be achieved by incorporating a wider variety of engineering challenges, particularly from domains that were less represented in our initial study. We believe that industry can play a crucial role in this step, by providing representative problems for different types of engineering design tasks faced by them.
Such diversity in problem selection will provide a more comprehensive understanding of VLM's capabilities across the engineering design process. Alongside this, there is a need for enhanced dataset curation. Developing more robust datasets that closely mirror complex, real-world engineering scenarios can significantly improve the model's evaluation. These datasets should capture the multifaceted and multidimensional nature of engineering tasks, allowing for a more nuanced assessment of multimodal VLMs' applicability and effectiveness. An effort should also be made to avoid publicly available datasets, to limit challenges with evaluation data leaking into future model training.

Another critical area of future work lies in the realm of human-AI collaboration. It is imperative to study how engineers interact with VLMs in real-world design scenarios. Such studies can shed light on practical utility, user trust, and the integration of AI into engineering workflows. This includes understanding how engineers' biases and decision-making processes interact with AI-generated solutions. Additionally, conducting longitudinal studies to monitor the impact of model evolution over time on its performance in engineering tasks will be highly beneficial. Given the rapid developments in AI, understanding how updates and changes affect its applicability and effectiveness is crucial. This will help in keeping the AI applications in engineering up-to-date and relevant, ensuring that they continue to meet the evolving demands of the field.

\section*{Funding Data}

\begin{itemize}
\item Swiss National Science Foundation (Postdoc.Mobility Fellowship P500PT\_206937).
\item National Science Foundation Graduate Research Fellowship.
\end{itemize}

\bibliographystyle{plainnat}

\end{document}